\documentclass[lettersize,journal]{IEEEtran}
\usepackage{amsmath,amsfonts}
\usepackage{array}
\usepackage[caption=false,font=normalsize,labelfont=sf,textfont=sf]{subfig}
\usepackage{textcomp}
\usepackage{stfloats}
\usepackage{url}
\usepackage{amsfonts}
\usepackage{multirow}
\usepackage{booktabs}
\usepackage{verbatim}
\usepackage{graphicx}

\usepackage{cite}
\usepackage{mathtools}
\usepackage{algorithmicx}
\usepackage{algorithm}
\usepackage{algpseudocode}
\usepackage{soul}
\usepackage{color, xcolor} 

\usepackage{lscape}

\usepackage[figuresright]{rotating} 
\begin{document}

\title{AutoPV: Automatically Design Your Photovoltaic Power Forecasting Model}

\author{Dayin~Chen,
        Xiaodan~Shi{\IEEEauthorrefmark{1}},
        Mingkun~Jiang,
        Haoran~Zhang,~\IEEEmembership{Senior Member,~IEEE,}       
        Dongxiao~Zhang{\IEEEauthorrefmark{1}},
        Yuntian~Chen,~\IEEEmembership{Member,~IEEE,}
        and Jinyue~Yan{\IEEEauthorrefmark{1}}

\thanks{* Xiaodan Shi Dongxiao Zhang and Jinyue Yan are the corresponding authors.}
\thanks{Dayin Chen is with Department of Building Environment and Energy Engineering, The Hong Kong Polytechnic University, Hong Kong SAR, China; International Centre of Urban Energy Nexus, The Hong Kong Polytechnic University,  Hong Kong SAR, China; Research Institute for Smart Energy, The Hong Kong Polytechnic University, Hong Kong SAR, China; Eastern Institute for Advanced Study, Eastern Institute of Technology, Ningbo, China. E-mail: 23038748r@connect.polyu.hk}
\thanks{Xiaodan Shi is with School of Business, Society and Technology, Mälardalens University, 72123 Västerås, Sweden; Center for Spatial Information Science, the University of Tokyo, 5-1-5 Kashiwanoha, Kashiwa-shi, Chiba 277-8568, Japan. E-mail: xiaodan.shi@mdu.se}
\thanks{Mingkun Jiang is with PV Industry Innovation Center, State Power Investment Corporation, 710061 Xi'an, Shaanxi, China. E-mail: jiangmk\_spic@163.com}
\thanks{Haoran Zhang is with School of Urban Planning and Design, Peking University, Shenzhen, China. E-mail: h.zhang@pku.edu.cn}
\thanks{Dongxiao Zhang and Yuntian Chen are with Ningbo Institute of Digital Twin, Eastern Institute of Technology, Ningbo, China. E-mail: dzhang@eitech.edu.cn; ychen@eitech.edu.cn}
\thanks{Jinyue Yan is with Department of Building Environment and Energy Engineering, The Hong Kong Polytechnic University, Hong Kong SAR, China; International Centre of Urban Energy Nexus, The Hong Kong Polytechnic University,  Hong Kong SAR, China; Research Institute for Smart Energy, The Hong Kong Polytechnic University, Hong Kong SAR, China. E-mail: j-jerry.yan@polyu.edu.hk}

}


\maketitle

\begin{abstract}
Photovoltaic power forecasting (PVPF) is a critical area in time series forecasting (TSF), enabling the efficient utilization of solar energy. With advancements in machine learning and deep learning, various models have been applied to PVPF tasks. However, constructing an optimal predictive architecture for specific PVPF tasks remains challenging, as it requires cross-domain knowledge and significant labor costs. To address this challenge, we introduce AutoPV, a novel framework for the automated search and construction of PVPF models based on neural architecture search (NAS) technology. We develop a brand new NAS search space that incorporates various data processing techniques from state-of-the-art (SOTA) TSF models and typical PVPF deep learning models. The effectiveness of AutoPV is evaluated on diverse PVPF tasks using a dataset from the Daqing Photovoltaic Station in China. Experimental results demonstrate that AutoPV can complete the predictive architecture construction process in a relatively short time, and the newly constructed architecture is superior to SOTA predefined models. This work bridges the gap in applying NAS to TSF problems, assisting non-experts and industries in automatically designing effective PVPF models.

\end{abstract}


\section{Introduction}

\IEEEPARstart{P}hotovoltaic (PV) power forecasting (PVPF) involves predicting future PV power generation based on historical power data and corresponding weather features. The variable and non-controllable nature of PV power generation necessitates precise forecasting, which is crucial for maintaining grid balance and offering technical assistance during trading activities \cite{intro_pv}. With the advent of deep learning, 
technologies such as Multi-Layer Perceptron (MLP) \cite{MLP}, Convolutional Neural Networks (CNN) \cite{cnn}, Long Short-Term Memory (LSTM) \cite{LSTM}, and Temporal Convolutional Network (TCN) \cite{tcn} have been extensively applied in the field of PVPF. Even in physics-based PVPF models, deep learning has been leveraged in various applications, such as enhancing numerical weather prediction (NWP) \cite{nwp} and sky image analysis \cite{sky_gpt}. These advancements highlight the growing mainstream adoption of deep learning techniques in the field of PVPF.

As known to us, PVPF is inherently a time series forecasting (TSF) problem \cite{survey2}, which is a prominent area within data mining. The recent surge in transformer architectures \cite{attention} has led to the development of numerous transformer-based models such as informer \cite{informer}, Autoformer \cite{Autoformer}, and FEDformer \cite{fedformer}, which have demonstrated exceptional performance in TSF problems. Concurrently, several non-transformer-based TSF models, including DLinear \cite{DLinear} and TSMixer \cite{tsmixer}, have also been developed and shown strong performance. These typical TSF models have now started to be applied in the context of PVPF tasks as well.

Given the plethora of PVPF models available, a significant challenge lies in selecting an optimal model for a specific task. This issue is particularly challenging for non-experts and industries, as building a PVPF model requires expertise across multiple domains such as Energy Engineering and Computer Science. Factors including data granularity, data scale, and data distribution can significantly impact the final model design, thus making the model design process a non-trivial endeavor. A model that performs excellently on one task may not perform as well on another, which can lead to suboptimal decision-making and introduce various risks. Therefore, an efficient and automatic model search and construction strategy is essential in the current landscape of proliferating PVPF models.

Neural Architecture Search (NAS) \cite{nas1} is an emerging technology aimed at automating the process of model structure discovery and design, which can address the aforementioned issue. It comprises three fundamental elements: the search space, the evaluator, and the search strategy. The search space encompasses a variety of module choices that can be assembled into a cohesive architecture. The search strategy employs a specific algorithm to navigate through and combine these modules, with the goal of identifying an optimal architecture. Meanwhile, the evaluator is tasked with rating each generated architecture.


However, despite the significant advancements in NAS for many years, there has been limited exploration of NAS specifically for TSF tasks. Instead, most current NAS efforts are predominantly focused on visual tasks. This focus is largely due to the relatively straightforward nature of vision model components, such as convolutional and pooling layers, which lack complex interdependencies and can be seamlessly integrated with other standard modules, including linear layers, activation functions, and softmax layers. In contrast, TSF models are often cohesive, indivisible entities with distinct design principles. For instance, some TSF models employ seasonal-trend decomposition \cite{DLinear, Autoformer, fedformer, client}, while others analyze data in the frequency domain \cite{FreTS, etsformer}. These models are designed from multiple and varied perspectives, making it challenging to decompose them into modular components that can be used to construct an effective search space.

In this study, we introduce AutoPV, a novel NAS system for the automated search and design of predictive architectures tailored to specific PVPF tasks. We construct a novel NAS search space that encompasses various typical data processing techniques from state-of-the-art (SOTA) TSF models, as well as the foundational deep learning models commonly used in PVPF tasks. This meticulously designed search space encompasses four different stages with twelve optional parameters, enabling the exploration of approximately $2.5 \times 10^6$ potential architectures.


To evaluate the effectiveness of AutoPV, we designed two distinct PVPF tasks: one utilizing only historical data and the other incorporating future weather information. These tasks further encompass six sub-tasks with different forecasting lengths, ranging from half a day ahead to one month ahead. All the evaluation experiments were conducted using the dataset from the Daqing PV Station in China, covering the period from 2022 to 2023. Experimental results show that AutoPV exhibits outstanding performance across different PVPF tasks compared to other benchmarks. We believe this work could help bridge the gap in applying NAS to PVPF problems and potentially extend to TSF problems.

The principal contributions of this work are outlined as follows:
\begin{enumerate}
  \item{We construct a detailed knowledge graph and NAS search space for the domains of PVPF. This helps address the research gap in the automated design of effective PVPF models.}

  \item{Based on the constructed search space, we develop an automated PVPF model design system called AutoPV, which can assist non-experts in searching for effective models for specific PVPF tasks.}

  \item {We conduct extensive experiments on a PV power dataset from the Daqing Photovoltaic Station in China to demonstrate the effectiveness of AutoPV across various PVPF tasks and provide a comprehensive analysis of its characteristics. Results show the new searched architecture improves performance by 4.52\% compared to the best baseline model. Furthermore, when incorporating future weather information, it could boosts the improvement by 9.88\%.}

\end{enumerate}

\section{Related Work}
\subsection{Photovoltaic Power Forecasting (PVPF)}

PVPF tasks involve using historical PV power generation data along with weather information to model future PV power generation. Existing works involve a wide array of different models and also explore various forecasting lengths. For example, Gao et al. \cite{LSTM} use LSTM for day-ahead PV power prediction based on weather classification. Mellit et al. \cite{cnn} evaluate short-term PVPF (less than 8 time steps ahead) using deep learning models like LSTM, GRU, and CNN. Li et al. \cite{tcn} employ a TCN model for hours-ahead utility-scale PV forecasting. However, no single model has consistently outperformed others across different prediction scales or datasets. Efficiently designing an optimal model for a specific PVPF task remains a challenging problem.


\subsection{Time Series Forecasting (TSF) Models}

The following content introduces two main architectural categories of TSF models.

\textbf{Transformer-based Models.} The core idea of the transformer \cite{attention} is to allow the model to dynamically focus on different parts of the input by computing attention scores. In recent years, to address the computational complexity of the vanilla transformer \cite{attention} and enhance performance, researchers have proposed numerous transformer variants \cite{Autoformer, informer, fedformer, patchTST, itransformer,client}. Among these, some variants such as Autoformer \cite{Autoformer}, Informer \cite{informer}, FEDformer \cite{informer}, iTransformer \cite{itransformer} and PatchTST \cite{patchTST} are specifically designed for TSF problems. These methods have made various modifications to the vanilla transformer from different perspectives, such as incorporating frequency domain information \cite{fedformer}, dividing time tokens into patches \cite{patchTST}, and introducing data decomposition \cite{Autoformer, client}, to better capture the characteristics of time series data.

\textbf{Non-transformer-based Models.} Although transformer-based structures are currently popular, their validity and necessity for TSF problems have been questioned by some researchers \cite{DLinear}. They maintain that non-transformer-based models can achieve equally good performance while having a lower computational burden. Typical non-Transformer-based models include DLinear \cite{DLinear}, FreTS \cite{FreTS}, MICN \cite{MICN}, TSMixer \cite{tsmixer}, and TimesNet \cite{TimesNet}, among others. These models employ various time series data processing methods, such as decomposition, multi-scale decomposition, and frequency domain analysis, to extract high-level information from the data. The extracted information is often projected to the output using a fully connected (FC) layer. These models have demonstrated superior performance on a variety of publicly available datasets.


In recent years, a new approach of applying diffusion models to TSF has also emerged \cite{diffusion}. However, considering that the diffusion model is fundamentally different in principle from the other two types of methods (i.e., transformer-based models and non-transformer-based models), we have decided not to include diffusion models in the construction of the search space for this study.

\subsection{Neural Architecture Search (NAS)}

NAS was first proposed by Google in the work \cite{nas1}, where they used recurrent neural networks (RNNs) and reinforcement learning to automatically search and construct an image classification model on the CIFAR-10 dataset. The search space included parameters such as the length, width, and stride of the convolution operations. In their follow-up work the next year \cite{nas1}, they refined the approach by searching the structure of a single cell and then repeating it multiple times to build the final network. While this method improved the search efficiency, it still required training on 500 GPUs for 4 days, which is a huge burden for most industries. Additionally, tuning the hyperparameters for the reinforcement learning process is also a complex and tedious task. This motivates people to seek more efficient search algorithm and certain specialized tricks to accelerate the search process of NAS.

Weight sharing is a typical improvement strategy which sharing weights among the different architectures \cite{shareweight}. However, the stability and effectiveness of this method have been questioned \cite{no_sharing}. Beyond that, several studies have proposed different search strategies that eliminate the need for tedious hyperparameter tuning, such as evolutionary algorithm \cite{evo}. In the work \cite{bayesian}, the BANANAS algorithm is introduced, which utilizes Bayesian optimization techniques to search for optimal architectures. Furthermore, in the work\cite{mobonanas}, the BANANAS algorithm has been extended for multi-objective optimization. This extension, referred to as MoBananas, has been demonstrated to outperform many other existing NAS search algorithms




One significant challenge in NAS is that most research efforts have focused on searching CV models, as the modules and operations in the CV search space generally do not have complex relationships or restrictions \cite{wang}. However, the need for automated network design extends to many other domains as well. In recent years, researchers have gradually explored the application of NAS to other areas, particularly by constructing domain-specific search spaces. For example, Seng et al. apply NAS to search for the federated architecture and hyperparameters in their work \cite{nas_fed}. Wang et al. \cite{hat} use NAS to search the structure of transformer models for natural language processing (NLP) tasks. Additionally, Wang et al. \cite{wang} constructed a NAS search space based on five selected models for trajectory prediction.

To the best of our knowledge, there is limited prior work on extending NAS to the TSF or PVPF domain. In this study, we aim to address this gap by building a new search space and implementing an NAS system in the specific context of PVPF.

\section{Methodology}

\begin{figure*}[!t]
\centering
\includegraphics[width=7in]{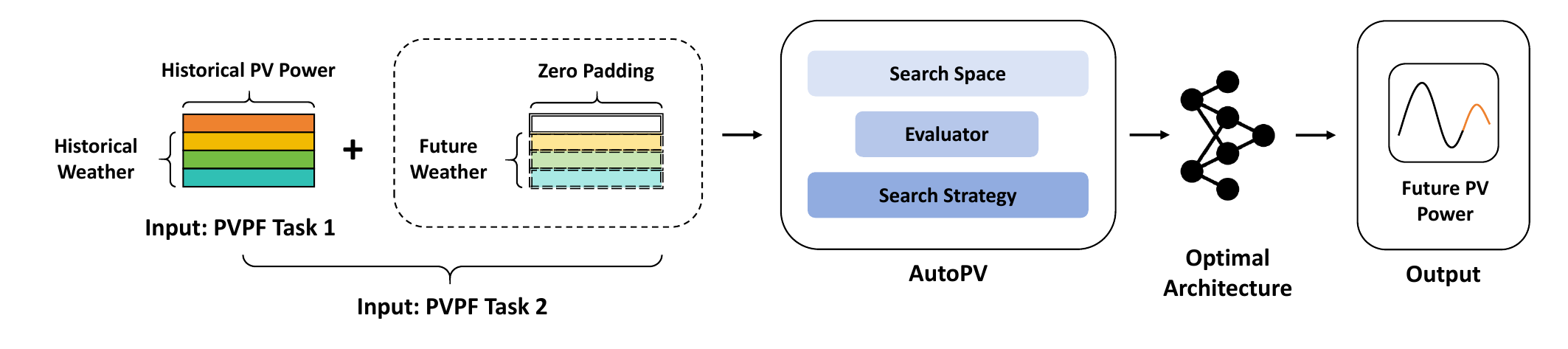}
\caption{The pipeline for leveraging AutoPV in PVPF tasks involves two distinct scenarios: PVPF Task 1, which utilizes only historical data series as input, and PVPF Task 2, which incorporates additional unstable future weather information. Zero padding is used to fill in the missing dimension of the future data. The AutoPV system comprises three main components: a search space, an evaluator, and a search strategy. Given a specific task type and dataset, AutoPV automatically searches for an optimal model architecture. This architecture can then directly perform the forecasting task, as the training process is completed during the architecture search.}
\label{framework}
\end{figure*}

\subsection{Problem Description}
\label{problemdescription}

\textbf{PVPF.} PVPF task can be formally defined as follows:
The input is a time series of historical PV power data, denoted as $\mathbf{X} = \{x_1, x_2, \ldots, x_t\} \in \mathbb{R}^{T_s \times D}$, where $x_t$ represents the data sample at time $t$, $T_s$ is the length of the historical time series, and $D$ is the dimensionality of each data point. A single data point comprises 1 dimension representing PV power generation data and $(D-1)$ dimensions corresponding to weather information features. The searched predictive model, $\mathcal{M}(\mathbf{X})$, takes $\mathbf{X}$ as input and outputs a series of future PV power generation predictions, denoted as $\mathbf{\hat{Y}} = \{\hat{y}_{t+1}, \hat{y}_{t+2}, \ldots, \hat{y}_{t+n}\} \in \mathbb{R}^{T_p \times 1}$. Here, $\hat{y}_{t+n}$ indicates the predicted value at step $n$ in the future, and $T_p$ is the length of the predicted time series. Each data point in $\mathbf{\hat{Y}}$ has only 1 dimension, representing the forecasted PV power generation.

Given the current availability of future weather information, in addition to the standard form of PVPF tasks, we have also designed another PVPF senario that utilizes future weather information, which is named PVPF Task 2. As depicted in Fig. \ref{framework}, the future weather data is appended to the input historical sequence data along the time dimension. To handle the unknown future PV power data, we employ zero padding as a placeholder. In this case, the new input data length becomes $T_s^{\prime}$ which equals $T_s + T_p$. In real-world scenarios, future weather information is typically obtained from weather reports, which can inherently contain biases. To account for these uncertainties, dynamic Gaussian noise is often added to the appended future weather data. The variance of this noise increases exponentially as the forecasting length grows longer.




\textbf{NAS.} NAS requires first building a search space, which we denote as $\mathbb{S}$. By selecting modules from $\mathbb{S}$, we can construct various PVPF architectures $\mathcal{M}$. The overall goal of the NAS framework is to discover and construct an optimal architecture $\mathcal{M}^{*}$ that minimizes the future PV power prediction error. In this work, we uniformly use the L1 loss, also known as mean absolute error (MAE), as the training and evaluation criterion. Let the ground truth of PV power generation from $t+1$ to $t+n$ be $\mathbf{Y} = \{y_{t+1}, y_{t+2}, \ldots, y_{t+n}\} \in \mathbb{R}^{T_p \times 1}$. The objective function is then:


\begin{equation}
\begin{aligned}
\mathcal{M}^{*} &= \arg\min_{\mathcal{M} \in \mathbb{S}} MAE_{\mathbb{D}_{val}}(\mathbf{W}^{*}_\mathcal{M}, \mathcal{M}) \\
\text{s.t.} \quad & \mathbf{W}^{*}_\mathcal{M} = \arg\min_{\mathbf{W}} \mathcal{L}_{\mathbb{D}_{train}}(\mathbf{W}, \mathcal{M})
\end{aligned}
\end{equation}


Where $\mathcal{L}_{\mathbb{D}_{train}}(\mathbf{W}, \mathcal{M})$ represents the training loss of the selected model $\mathcal{M}$ on the training dataset $\mathbb{D}_{train}$ with the model parameters $\mathbf{W}$.

\subsection{Search Space}

\begin{table*}
    \centering
    \setlength{\tabcolsep}{2.5mm}
  \caption{Stages and modules of the AutoPV search space}
  \label{tab:searchspace}
  \begin{tabular}{llll}
    \toprule
    \textbf{Stage}&\textbf{Function Description}&\textbf{Operations or Parameters} & \textbf{Typical Modules Extracted from Existed Works}\\
    \midrule

    

    \multirow{5}{*}{\parbox{2.5cm}{\textbf{Stage 1: Feature Selection Stage}}} & \multirow{5}{*}{\parbox{3cm}{Construct the optimal feature set of the input data.}} & \multirow{3}{*}{\parbox{4cm}{Feature Selection Method (FSM) }} & $\textbf{FSM}\sb{1}$: No Filter \\
    
   & & & $\textbf{FSM}\sb{2}$: mRMR\cite{MRMR1}\\
   & & & $\textbf{FSM}\sb{3}$: Pearson Correlation\cite{Pearson1}\\
    \cmidrule(l){3-4}
   & &\multirow{2}{*}{\parbox{4cm}{Feature Selection Threshold (FST)}} & $\textbf{FST}\sb{1}$: 0.3 \qquad 
   $\textbf{FST}\sb{2}$: 0.4 \\
   & & &  $\textbf{FST}\sb{3}$: 0.5\\



   \midrule

    \multirow{13}{*}{\parbox{2.5cm}{\textbf{Stage 2: Data Processing Stage}}} & \multirow{13}{*}{\parbox{3cm}{Process the data from different perspectives before passing it to the core predictive structure.}} & \multirow{2}{*}{\parbox{4cm}{Data Generalization Method (DGM)}} & $\textbf{DGM}\sb{1}$: None\\
    
    & & & $\textbf{DGM}\sb{2}$: Adding Gaussian Noise\\

    \cmidrule(l){3-4}
    
& & \multirow{3}{*}{\parbox{4cm}{Stationarization Method (SM)} }& $\textbf{SM}\sb{1}$: None \\

 & & & $\textbf{SM}\sb{2}$: RevIN \cite{RevIN}\\
 & & & $\textbf{SM}\sb{3}$: DAIN \cite{DAIN}\\

    \cmidrule(l){3-4}

   & &\multirow{2}{*}{\parbox{4cm}{Feature Adding Method (FAM)}} & $\textbf{FAM}\sb{1}$: None\\
      & & &  $\textbf{FAM}\sb{2}$: Adding Time Features\\

    \cmidrule(l){3-4}
  
       & & \multirow{6}{*}{\parbox{4cm}{Feature Extraction Method (FEM)} }& $\textbf{FEM}\sb{1}$: None\\
    
    & & & $\textbf{FEM}\sb{2}$: Linear Embedding\\
    
   & & & $\textbf{FEM}\sb{3}$: Decomposition \cite{Autoformer, fedformer, DLinear}\\
   & & & $\textbf{FEM}\sb{4}$: Multi-Scale Decomposition \cite{MICN}\\
      & & & $\textbf{FEM}\sb{5}$: Time \& Feature Mixing \cite{tsmixer}\\
   & & & $\textbf{FEM}\sb{6}$: Frequency Domain Mixing \cite{FreTS}\\

   \midrule

    \multirow{5}{*}{\parbox{2.5cm}{\textbf{Stage 3: Model Construction Stage}}} & \multirow{5}{*}{\parbox{3cm}{Build the core predictive structure of the designed architecture.}} & \multirow{2}{*}{\parbox{4cm}{Core Predictive Structure (CPS) }} & $\textbf{CPS}\sb{1}$: LSTM \qquad $\textbf{CPS}\sb{2}$: MLP (\& linear)\\
    
     & & & $\textbf{CPS}\sb{3}$: CNN \qquad $\textbf{CPS}\sb{4}$: TCN\\

    \cmidrule(l){3-4}
   & &\multirow{1}{*}{\parbox{4cm}{Layer Number (LN)}} & $\textbf{LN}\sb{1}$: 1 \qquad $\textbf{LN}\sb{2}$: 2 \qquad $\textbf{LN}\sb{3}$: 3\\

        \cmidrule(l){3-4}
   & &\multirow{2}{*}{\parbox{4cm}{Hidden Size (HS)}} & $\textbf{HS}\sb{1}$: 64 \qquad $\textbf{HS}\sb{2}$: 128  \\
& & & $\textbf{HS}\sb{3}$: 256 \qquad $\textbf{HS}\sb{4}$: 512\\

 \midrule

    \multirow{3}{*}{\parbox{2.5cm}{\textbf{Stage 4: Model Training Stage}}} & \multirow{3}{*}{\parbox{3cm}{Set the detailed training settings of the designed architecture.}} & \multirow{1}{*}{\parbox{4cm}{Learning Rate (LR)}} & $\textbf{LR}\sb{1}$: 0.0005 \qquad $\textbf{LR}\sb{2}$: 0.001\\


        \cmidrule(l){3-4}
   & &\multirow{1}{*}{\parbox{4cm}{Optimization Function (OF)}} & $\textbf{OF}\sb{1}$: Adam \qquad $\textbf{OF}\sb{2}$: SGD\\

           \cmidrule(l){3-4}
   & &\multirow{1}{*}{\parbox{4cm}{Batch Size (BS)}} & $\textbf{BS}\sb{1}$: 32 \qquad $\textbf{BS}\sb{2}$: 64\\

  \bottomrule
\end{tabular}
\end{table*}

\begin{figure*}
  \centering
 \subfloat[Forecasting process in typical TSF models]{\includegraphics[width=3.5in]{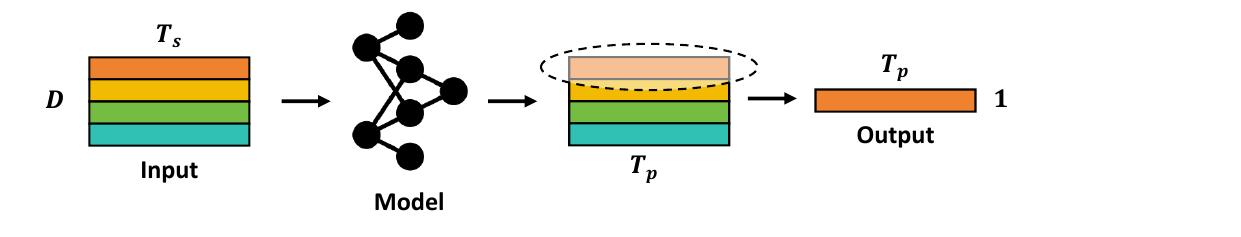}}
  \hfil
  \subfloat[Forecasting process in AutoPV]{\includegraphics[width=3.5in]{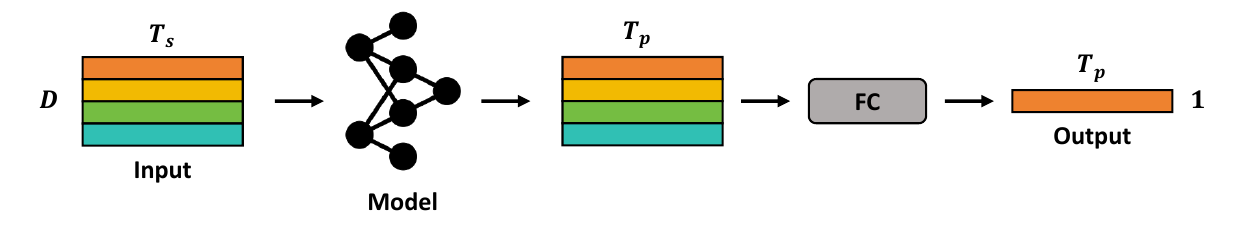}}

  \caption{A minor modification in the core predictive structure of AutoPV. In typical TSF models, the forecasting results for all features are generated first, then the PV power sequence is selected. In contrast, the AutoPV framework adds a fully connected layer to aggregate all forecasting results of different features and generate the final PV power forecast.}
  \label{compare}
\end{figure*}

\begin{figure*}[!t]
\centering
\includegraphics[width=7in]{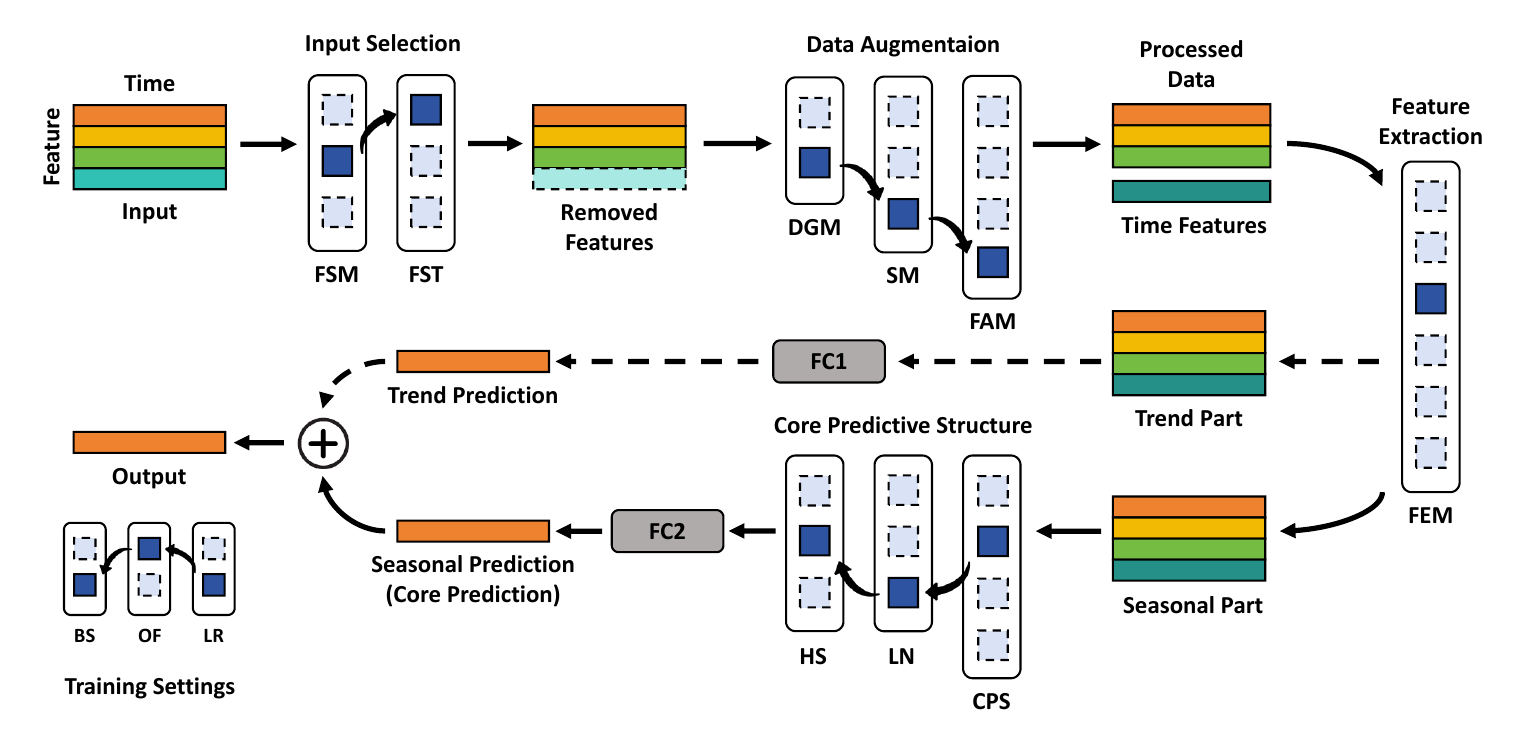}
\caption{The detailed process of neural architecture search and construction. The FFM and FFT parameters determine the final feature set. The DGM, SM, and FAM parameters conduct data augmentation on the original data. Time embedding features can be optionally added at this stage. The FEM parameter extracts different types of information from the processed data. As an example, $\mathbf{FEM}_3$ decomposes the data into trend and seasonal parts. The seasonal part is then input into the core predictive structure, which is decided by the CPS, LN, and HS parameters. The trend part is embedded by a fully connected (FC) layer and added to the output of the core predictive structure. For other $\mathbf{FEM}_x$ methods that output only a single tensor, we process the tensor as the pipeline of the seasonal part. Finally, the loss value is calculated using the prediction output and the ground truth, which serves as the evaluation score. The BS, OP, and LR parameters decide the training settings of the overall architecture.}
\label{pipeline}
\end{figure*}


The search space is the most critical component of the NAS framework. As detailed in Table \ref{tab:searchspace}, we divide the overall AutoPV search space into four stages and employ a total of 12 parameters to describe the entire searched architecture. This encompasses the complete model-building and training pipeline, including feature selection, data processing, model construction, and training process settings. The optional modules in each stage are derived from either typical PVPF works or SOTA TSF models. To ensure that the constructed models remain relatively lightweight, we exclusively extract processing methods from non-transformer-based models to build our search space. Experimental results demonstrate that the models constructed using these simple modules can outperform predefined SOTA transformer-based models.

In the following sections, we illustrate each stage of the AutoPV search space in detail.


\textbf{Stage 1: Feature Selection.} In this initial stage, we refine the data by selecting relevant features, as including irrelevant ones can impair the model's effectiveness. We use two common feature selection methods (FSM) in the PVPF domain: Maximum Relevance and Minimum Redundancy (mRMR) \cite{MRMR1}, and Pearson Correlation \cite{Pearson1}. The Pearson Correlation coefficient measures the linear association between a feature and the target variable, while mRMR evaluates both feature relevance and inter-feature redundancy. We also optimize the feature selection threshold (FST) for these methods, treating it as an additional hyperparameter.

\textbf{Stage 2: Data Processing.} In this phase, a series of preprocessing operations are applied to the input data to prepare it for the core predictive model.

The first parameter, Data Generalization Method (DGM), determines whether Gaussian noise should be added to the original dataset. This technique is known for enhancing the model's training stability and generalization capability \cite{gaussian}.

The second parameter involves two common Stationarization Methods (SM) to address distributional shifts. The first method, RevIN \cite{RevIN}, is an adaptable instance normalization technique that normalizes the input data to a mean-centered distribution before training. After training, RevIN uses a reverse denormalization procedure to restore the output distribution to its original state. The second method, DAIN \cite{DAIN}, employs nonlinear neural networks for dynamic adaptive normalization of the data series.

We also introduce the Feature Adding Method (FAM) parameter to determine whether to include additional timestamp embedding features. Time series data often exhibit seasonal and periodic variations, making it beneficial to derive predictive insights from timestamps. This kind of feature has been widely used in various transformer-based models, such as Transformer \cite{attention}, Informer \cite{informer}, and FEDformer \cite{fedformer}.


The Feature Extraction Methods (FEM) parameter plays a crucial role in enhancing the learning efficacy of TSF models. Within our AutoPV search space, we incorporate a diverse array of feature extraction methods from well-established non-transformer-based TSF models to process the input data, thereby obtaining distinct high-level information. These feature extraction methods have demonstrated their efficiency and have been adapted and evolved within certain transformer-based architectures, leading to the creation of novel structures. The methodologies corresponding to the entries in Table \ref{tab:searchspace} are introduced as follows:

\begin{itemize}
\item{$\mathbf{FEM_1}$}: This method leaves the data in its original state without applying any modifications.
\item{$\mathbf{FEM_2}$}: This method employs a single linear layer to embed the input data. It is a common preprocessing step before introducing data into deep learning architectures such as LSTM, representing one of the simplest strategies for extracting multidimensional information.
\item{$\mathbf{FEM_3}$}: This method performs a seasonal-trend decomposition on the original time series, segregating it into trend and seasonal components that represent the long-term progression and cyclic patterns of the series, respectively. Such a technique is extensively utilized in several prominent models, including Autoformer \cite{Autoformer}, FEDformer \cite{fedformer}, and DLinear \cite{DLinear}. Following their methodologies, we apply an additional FC layer to process the trend component, while the seasonal component is handled by the core predictive structure selected through our search. The outputs of these two components are then aggregated. Fig. \ref{pipeline} shows the complete process.

\item{$\mathbf{FEM_4}$}: Unlike the $FEM_3$ approach, which employs a fixed-size moving average window for data decomposition, $FEM_4$ introduces a multi-scale hybrid decomposition block that employs multiple moving windows with varying kernel sizes. This approach, inspired by MICN \cite{MICN}, aims to capture a broader and more detailed range of patterns within the time series.

\item{$\mathbf{FEM_5}$}: Inspired by the TSMixer model \cite{tsmixer}, $FEM_5$ employs the MLPs on both the time and feature dimensions to integrate information. The similar concept has also been applied in the iTranformer \cite{itransformer}.

\item{$\mathbf{FEM_6}$}: $FEM_6$ is similar to $FEM_5$, with the primary distinction being its application of MLPs in the frequency domain. Specifically, $FEM_6$ utilizes a frequency channel learner to model inter-series dependencies through a frequency-domain MLP applied across channel dimensions. Concurrently, a frequency temporal learner models temporal dependencies by applying a frequency-domain MLP across time dimensions. This method is from the FreTS model \cite{FreTS}.
\end{itemize}

\textbf{Stage 3: Model Construction Stage.} In this stage, we focus on selecting a series of parameters to build the core predictive structure. We incorporate four of the most representative deep learning architectures commonly used in PVPF tasks: LSTM, MLP, CNN, and TCN. The choice of architecture is determined by the Core Predictive Structure (CPS) parameter. Additionally, the detailed structure is defined by the Layer Number (LN) and Hidden Size (HS), which are also essential components of our search space. It is important to note that within this framework, a 1-layer MLP is conceptualized as a linear layer.

Our AutoPV framework introduces a minor modification to the typical TSF architecture. As illustrated in Fig. \ref{compare}, conventional TSF models typically generate the future sequence for all input features, with the sequence corresponding to PV power subsequently extracted as the final forecast result. Instead of directly extracting the PV power sequence, AutoPV incorporates an FC layer at the end to aggregate the forecasting results of all features. This design choice is motivated by the nature of the PVPF task, where all input features contribute to PV power generation. By aggregating information from all features, the AutoPV framework can better leverage the interdependencies between the features to produce a more accurate PV power forecast.

\textbf{Stage 4: Model Training Stage.} This stage involves the calibration of several standard training parameters, including the Learning Rate (LR), Optimization Function (OF), and Batch Size (BS). To improve the search efficiency of the overall architecture, we have constrained the search range for these parameters.

\textbf{Other Hyper-parameters.} In addition to the parameters such as layer number and hidden size discussed earlier, there are other hyper-parameters specific to certain modules in Stage 2 and Stage 3. To maintain the universality of the search space and enhance the efficiency of the search process, we assign standardized fixed values to these hyper-parameters. The details of all fixed hyper-parameters are provided in the Appendix.


\subsection{Architecture Construction Pipeline}

Fig. \ref{pipeline} illustrates the construction of the entire architecture using the 12 parameters defined within the search space. In the initial stage, the parameters FSM and FST execute an input selection operation across the feature dimensions. Subsequently, the DGM, SM, and FAM parameters implement a suite of data augmentation operations, including the addition of noise, data stationarization, and the incorporation of supplementary timestamp embedding features.

The next step involves the FEM parameter. Among the included methods, $\mathbf{FEM}_3$ and $\mathbf{FEM}_4$ decompose the original time series data into two separate streams, whereas $\mathbf{FEM}_5$ and $\mathbf{FEM}_6$ output a single data series with a modified shape. Taking $\mathbf{FEM}_3$ as an example in Fig. \ref{pipeline}, the seasonal component is fed into the core predictive structure which is constructed in the subsequent stage, while the trend component is processed by an FC layer. This processing method is based on prior researches that considers the trend component to exhibit relatively simple patterns that do not necessitate a complex model for processing. The outputs of these two components are then aggregated to reconstruct the original sequence. For other feature extraction methods that yield a single processed output, we treat it similarly to the seasonal component in Fig. \ref{pipeline} and omit the flow path for the trend component.

The core predictive structure is defined by the CPS, LN, and HS parameters. It is a straightforward deep learning model that accepts the seasonal component as input. As mentioned earlier, an additional FC layer is added to aggregate the forecasting results of all input features. The generated seasonal prediction result then is added with the trend prediction as the final output. To enhance clarity, all non-essential processes are depicted with dotted lines in Fig. \ref{pipeline}.

Finally, the BS, OF, and LR parameters are searched and govern the overall training configuration of the architecture.

\subsection{Evaluator}

Within the NAS framework, the evaluator is responsible for determining the efficacy of the architectures identified through the search process. In our AutoPV system, we assess the performance of an architecture based on its prediction error on the validation dataset.

Specifically, after a neural architecture is constructed, it is initially trained on the training dataset. Here we employ an early stopping strategy with a patience of 3 epochs to control the training process. Subsequently, the trained model is used to perform prediction on the validation dataset. The MAE across the entire validation dataset constitutes the final evaluation score, with a lower value indicating a superior architecture. This metric will be referred to as Measured Error in subsequent sections.

During this evaluation process, a hash map is maintained to prevent the redundant assessment of identical structures. This helps improve the efficiency of the overall NAS framework by avoiding unnecessary re-evaluations.

Additionally, given the growing interest in device-dependent NAS \cite{device}, our evaluator also calculates and optimizes the number of parameters within the searched architecture. This aspect will be elaborated upon in the experimental section.

\subsection{Search Strategy}
\label{ss}

\textbf{Architecture Encoding Method.} The majority of existing NAS research treats the search space as a graph, employing an adjacency matrix representation or path encoding \cite{bayesian} to characterize different architectures. This is because these NAS frameworks are primarily tailored to CV tasks, which necessitate the flexible splicing, repetition, and skipping of modules. However, such encoding methods are meaningless to our search space, as the modules for TSF tasks cannot be arbitrarily combined as in visual tasks. Moreover, the option to omit operations (denoted as 'None' in the table) is already incorporated as a selectable module in specific stages, obviating the need to accommodate architectures of dynamic lengths. 

Based on these considerations, we adopt a one-hot encoding sequence with a fixed length to represent each architecture. This encoding method enhances the efficiency of the search process and facilitates the inclusion of non-architectural hyperparameters such as learning rate and batch size \cite{mobonanas}.

\textbf{MoBananas Search.} To evaluate the effectiveness of the constructed search space, we employ the MoBananas algorithm \cite{mobonanas} in this work to search for the optimal architecture while simultaneously maintaining a relatively small model scale. It is a multi-objective optimization variant of the Bananas algorithm \cite{bayesian}, which utilizes Bayesian optimization to conduct an architecture search.

The MoBananas algorithm is characterized by three key innovative components. First, throughout the process of searching and evaluating different architectures, MoBananas trains a Deep Neural Network (DNN) predictor $\hat{f}$, which is capable of forecasting the performance of new architectures. Second, MoBananas selects the top $k_p$ parent architectures for mutation by employing non-dominated sorting \cite{non_dominated}. The third innovation is the application of Thompson Sampling \cite{thompson_sampling} by MoBananas to choose the subsequent generation from a pool of new candidate individuals. The comprehensive procedure of MoBananas Search is detailed in Algorithm \ref{algo2}. To save space, we do not repeat the whole process in the main text.

\begin{algorithm}[ht!]  
	\renewcommand{\algorithmicrequire}{\textbf{Input:}}
	\renewcommand{\algorithmicensure}{\textbf{Output:}}
	\caption{MoBananas Search}  
	\label{algo2}
	\begin{algorithmic}[1] 
		\Require $k_{ini}$, $k_{p}$, $p_m$, $k_{m}$, $k_{l}$, $T_{max}$, $\hat{f}$  
        \State Randomly initialize the population with size of $k_{ini}$
        \State Evaluate the initial individuals
        \Repeat
            \State Train the predictor $\hat{f}$
            \State Sort the population using non-dominated sorting
            \State Select the top $k_p$ parents
            \State Mutate for $k_m$ times with the probability of $p_m$
            \State Evaluate new individuals with predictor $\hat{f}$
            \State Select $k_l$ individuals using Thompson Sampling
            \State Evaluate and add the $k_l$ individuals into the population
            
        \Until {reach maximal iteration number $T_{max}$}
		
	\end{algorithmic}
\end{algorithm}

\section{Experiments}

\subsection{Experimental Setup}

\textbf{Datasets.} The dataset used in this paper is collected from a PV array located in Daqing city, China. Each data point in the dataset contains 11 dimensions, including PV power output, solar irradiation, wind speed, humidity, and other relevant weather parameters. The data was recorded from August 1, 2022 to October 28, 2023, with a sampling frequency of 1 minute, resulting in a total of 612,059 data points. However, the original dataset contained some missing values and outliers. To address these issues, we performed the following data cleaning and completion procedures:


\begin{itemize}

\item{\textbf{Data Deletion}}: We first group the data by day. The following two cases of daily recordings are considered abnormal, and the entire day's recordings will be removed: 1) the ratio of missing values and zero values in a given day exceeds 0.8; 2) the same outlier occurs ten or more times consecutively in one day.



\item{\textbf{Data Insertion}}: For the remaining days of recording in the dataset, we first set the recorded values with timestamps between 21:00 and 24:00, as well as those between 0:00 and 3:00, to 0. This is because there is no solar irradiation during these periods throughout the entire year. Next, starting from one side, for each missing data point, we apply a moving average window of size 3 to calculate the average of the surrounding non-missing values. We then use this average value to impute the missing data.

\end{itemize}

Following data cleaning and imputation, we retain a complete dataset encompassing 369 days, totaling 8,856 samples. To make the variations between data points more pronounced and to improve the meaningfulness of short-term predictions, we down-sample the original dataset to a resolution of one sample per hour. Detailed information regarding this dataset is presented in Table \ref{tab:dataset}.

We partition the dataset in a 6:2:2 ratio. Specifically, the initial 60\% of the dataset is allocated for training, the subsequent 20\% for validation, and the final 20\% for testing. Each training sample was segmented using a sliding window approach, with a window size of ($T_s$ + $T_p$) and a step size of 1. We ensured that the ($T_s$ + $T_p$) values within each sample were continuous in time by skipping any discontinuous parts during the sampling process. The first $T_s$ data points within each window serve as the historical sequence for input, while the subsequent $T_p$ data points represent the actual future PV power output, serving as the ground truth.



\begin{table*}[!t]
\caption{The Detailed Information of the original dataset and the dataset after preprocessing.\label{tab:dataset}}
\centering
\begin{tabular}{c|>{\centering\arraybackslash}p{4cm}>{\centering\arraybackslash}p{4cm}}
\hline
& \textbf{Original Dataset} & \textbf{Preprocessed Dataset}\\
\hline
\textbf{Data Granularity} & minute & hour\\
\textbf{Sample Size} & 612059 & 8856\\
\textbf{Number of Days Included} & 454 & 369\\
\hline
\textbf{Specification} & \multicolumn{2}{c}{210mm PERC Bifacial Module (Power: 590W, Conversion Efficiency: 20.8\%)}\\
\textbf{Feature Information} & \multicolumn{2}{c}{PV power generation and the corresponding weather information}\\
\textbf{Feature Number} & \multicolumn{2}{c}{11}\\
\textbf{Time Span} & \multicolumn{2}{c}{2022-08-01 to 2023-10-28}\\
\textbf{Location} & \multicolumn{2}{c}{Daqing, Heilongjiang, China}\\
\hline
\end{tabular}
\end{table*}

\subsection{Evaluation Metrics}


Two evaluation metrics are employed in the experiments: MAE as shown in Eq. \ref{eq:2} and weighted mean absolute percentage error (WMAPE) as shown in Eq. \ref{eq:4}. The reason why we do not use the standard mean absolute percentage error (MAPE) metric is that the PV power dataset contained numerous small values, which would lead a to a significant impact on the evaluation result. In these two equations, $\mathbf{\hat{Y}}$ represents the prediction results, and $\mathbf{Y}$ denotes the ground truth.

\begin{equation}
\label{eq:2}
\begin{aligned}
MAE(\mathbf{\hat{Y}}, \mathbf{Y}) = \frac{1}{N} \sum_{n=1}^{N} \left| \hat{y}_n - y_n \right|
\end{aligned}
\end{equation}


\begin{equation}
\label{eq:4}
\begin{aligned}
WMAPE(\mathbf{\hat{Y}}, \mathbf{Y}) &= \frac{\sum_{n=1}^{N} w_n\left| \hat{y}_n - y_n \right|}{\sum_{n=1}^{N} w_n \left| y_n \right|} \\
w_n &= \left| y_n \right|
\end{aligned}
\end{equation}

\subsection{Performance Evaluation}

The effectiveness of the AutoPV framework is evaluated on both PVPF Task 1 and PVPF Task 2, as described in Section \ref{problemdescription} and illustrated in Fig. \ref{framework}. For each task, six different sub-tasks with varying forecasting lengths are defined: 12-step (half-day) ahead forecasting, 24-step (one-day) ahead forecasting, 48-step (two-day) ahead forecasting, 72-step (three-day) ahead forecasting, 168-step (one-week) ahead forecasting, and 336-step (one-month) ahead forecasting. To facilitate the discussion, we will refer to these sub-tasks as DQH-12, DQH-24, DQH-48, DQH-72, DQH-168, and DQH-336, respectively. The historical data sequence length for all sub-tasks is fixed at 96.

\begin{figure}
  \centering
 \subfloat[PVPF Task 1]{\includegraphics[width=1.6in]{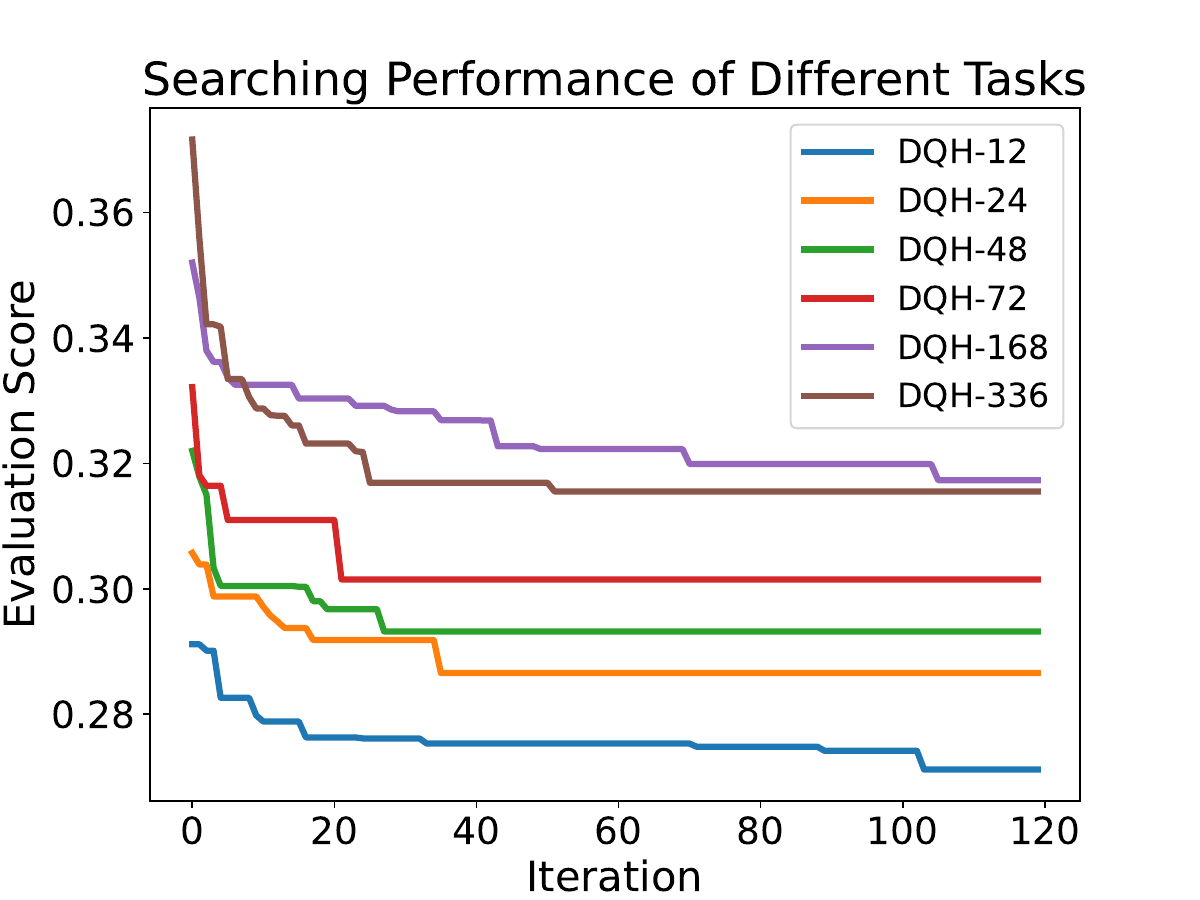}}
  \hfil
  \subfloat[PVPF Task 2]{\includegraphics[width=1.6in]{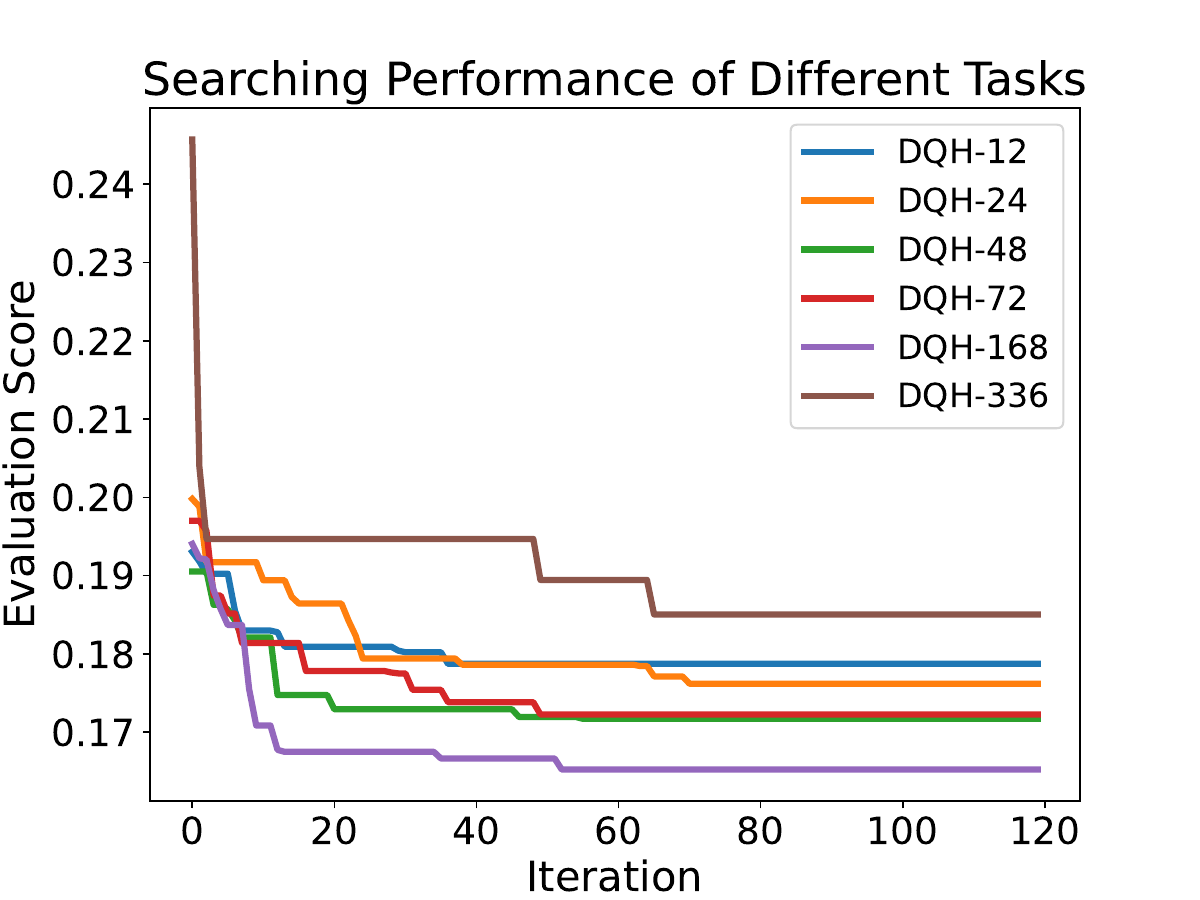}}

  \caption{Search performance of different PVPF tasks.}
  \label{compare_nas}
\end{figure}

\textbf{Search Efficiency Visualization.} In our search experiments, all hyperparameters for the MoBananas algorithm are set as follows: $k_{ini} = 10$, $k_p = 10$, $p_m = 0.2$, $k_m = 5$, $k_l = 10$, $T_{max} = 120$. Each search task is executed on an NVIDIA RTX 3090 GPU. Fig. \ref{compare_nas} illustrates the entire architecture search process, where the ordinate represents the measured error of the optimal model at each iteration. The results indicate that most search tasks could converge at around 70 iterations. By comparing the search results of PVPF Task 1 and PVPF Task 2, we observe that incorporating future weather information significantly reduces the measured error of the predictive architecture.

\begin{figure}
  \centering
 \subfloat[PVPF Task 1]{\includegraphics[width=1.6in]{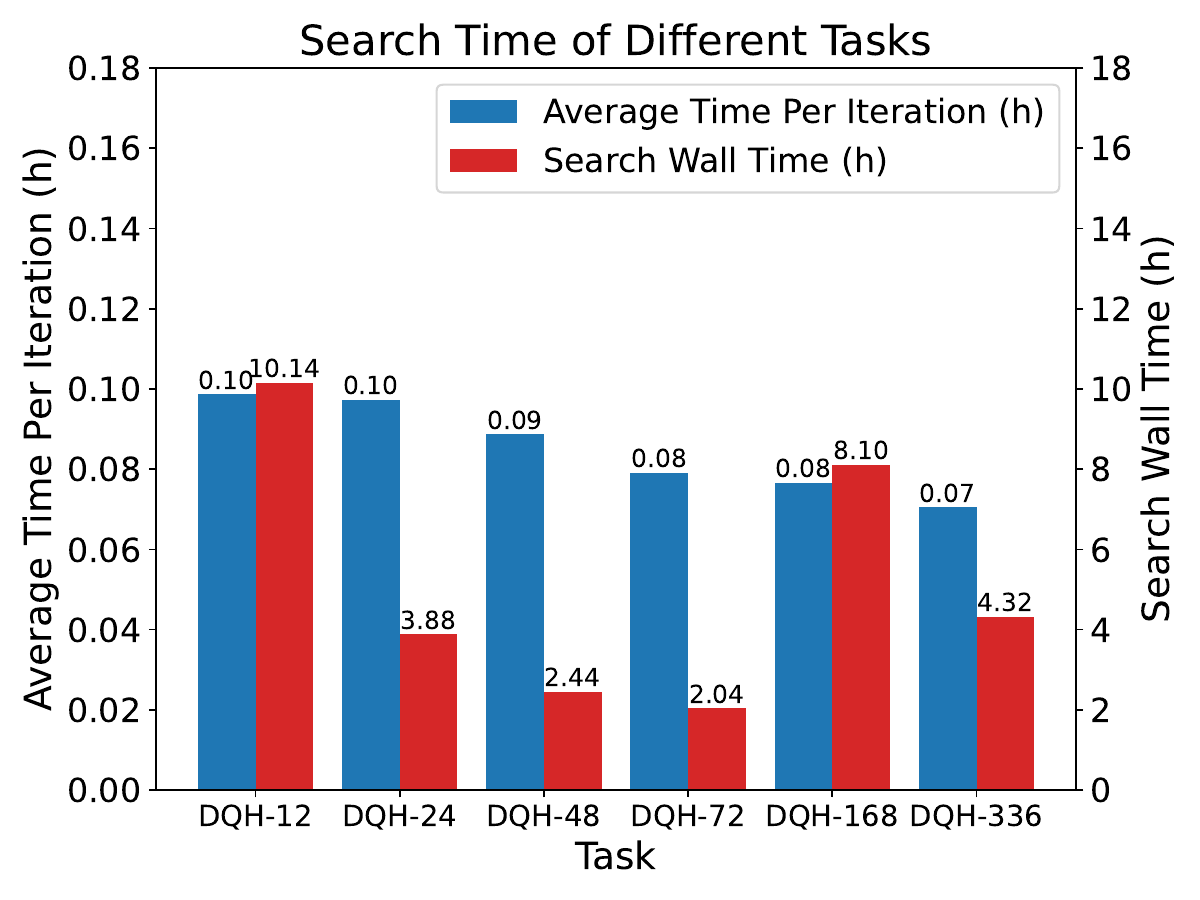}}
  \hfil
  \subfloat[PVPF Task 2]{\includegraphics[width=1.6in]{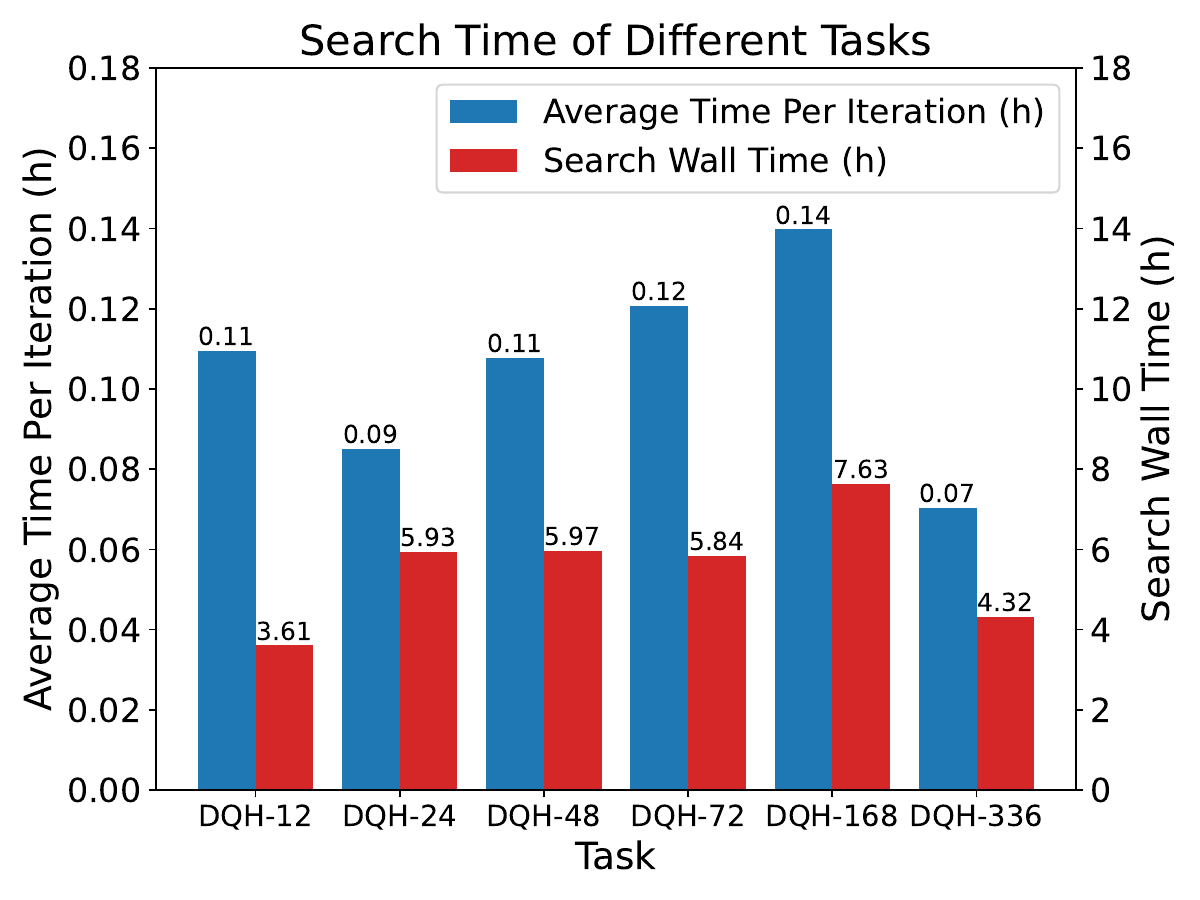}}

  \caption{The average search time of each iteration and the search wall time for finding the optimal architecture across different PVPF tasks.}
  \label{compare_time}
\end{figure}

Fig. \ref{compare_time} presents the average search time per iteration and the total wall time required to find the optimal architecture across different PVPF tasks. Due to the increased input size from integrating future weather data, PVPF Task 2 requires more time for each iteration of search and evaluation. On average, the AutoPV system takes 6.42 hours to construct an optimal forecasting architecture for a specific task. This time can be further reduced in practice by increasing computational resources and reducing the dataset size.

In the Appendix, we also demonstrate the multi-objective optimization process for different search tasks. In each iteration, the system constructs an optimal architecture set, known as the Pareto Frontier, which includes architectures that either exhibit good forecasting performance or have a small model scale. The results show that as the search progresses, the optimal architectures consistently move towards smaller scales and higher performance. In practical applications, a suitable architecture can be selected from the final Pareto Frontier based on the constraints imposed by the computing hardware.

\textbf{Baselines Comparison.}  
In this experiment, we select the architecture with the highest performance score (i.e., the lowest measured error) from the final Pareto frontier of each task as the result architecture of AutoPV for evaluation. This selected architecture is then compared with 13 well-recognized baseline forecasting models, which include: (1) Basic PVPF deep learning models: LSTM, MLP, CNN, and TCN; (2) Transformer-based models: Transformer, Autoformer, Informer, FEDformer, iTransformer and PatchTST; (3) Non-transformer-based models: DLinear, TimesNet and TSMixer. The implementation of the models refers to the Time Series Library \cite{tsl}.

For all evaluated benchmarks, the learning rate is set to 0.001, the optimizer used is Adam, and the batch size is 64. Each of the four basic deep learning models has 3 layers with a hidden size of 512. All transformer-based models also have 3 layers, consisting of 2 encoder layers and 1 decoder layer, with a model dimension fixed at 512. It is important to note that some TSF models, particularly the transformer-based models, require setting a portion of the historical sequence as the labeled sequence. For these models, the length of the labeled sequence is set to 48.

\begin{table*}[!t]
\tiny
\caption{Evaluation results of AutoPV and all selected benchmarks for PVPF task 1 (using historical information only). The input sequence length is fixed as 96 and the prediction lengths are set as different values. Each experiment is conducted 5 times. The best result is highlighted in \textbf{bold}, and the second is highlighted in \textcolor{blue}{blue}.\label{tab:baseline}}
\resizebox{\textwidth}{!}{

\begin{tabular}{c|cc|cc|cc|cc|cc|cc}
\hline
Task & \multicolumn{2}{c|}{DQH-12 (Half Day)} & \multicolumn{2}{c|}{DQH-24 (One Day)}  &\multicolumn{2}{c|}{DQH-48 (Two Days)}  & \multicolumn{2}{c}{DQH-72 (Three Days)}& \multicolumn{2}{c}{DQH-168 (One Week)}& \multicolumn{2}{c}{DQH-336 (One Month)}\\
Model & MAE& WMAPE & MAE& WMAPE & MAE& WMAPE & MAE& WMAPE & MAE& WMAPE & MAE& WMAPE\\
\hline
LSTM & 105.887$\pm$2.735 & 0.209$\pm$0.013 & 109.887$\pm$4.065 & 0.221$\pm$0.006 & 111.591$\pm$3.512& 0.228$\pm$0.007& 116.816$\pm$3.431 & 0.240$\pm$0.018 & 112.305$\pm$1.715 & 0.212$\pm$0.006& 110.623$\pm$7.149 & 0.220$\pm$0.011\\
MLP & 100.150$\pm$1.107 & 0.213$\pm$0.009 & 105.954$\pm$1.470 & 0.209$\pm$0.007 & 110.332$\pm$0.851 & 0.220$\pm$0.010& 112.153$\pm$0.808  & 0.212$\pm$0.007& 107.169$\pm$1.238 & 0.192$\pm$0.003& 133.864$\pm$1.798 & 0.291$\pm$0.006\\

CNN & 102.533$\pm$3.384 & 0.204$\pm$0.005 & 109.660$\pm$1.681 & 0.213$\pm$0.013 & 110.795$\pm$1.305 & 0.214$\pm$0.010 & 110.886$\pm$3.883 & 0.218$\pm$0.017 & \textcolor{blue}{101.103}$\pm$0.891 & 0.183$\pm$0.011& 119.530$\pm$17.926 & 0.281$\pm$0.060\\

TCN & 150.795$\pm$2.653 & 0.242$\pm$0.010 & 155.708$\pm$12.393 &0.260$\pm$0.008& 163.646$\pm$3.901 & 0.300$\pm$0.007 & 170.357$\pm$5.770 &  0.313$\pm$0.003 & 194.609$\pm$15.946 & 0.312$\pm$0.018& 313.508$\pm$71.918 & 0.351$\pm$0.042\\

Transformer & 104.367$\pm$4.023 & 0.242$\pm$0.009 & 110.051$\pm$2.011 & 0.229$\pm$0.006 & 114.694$\pm$5.927 & 0.258$\pm$0.017 &  107.583$\pm$2.680 & 0.234$\pm$0.016 & 107.845$\pm$7.811 & 0.229$\pm$0.034& 266.176$\pm$2.116 & 0.959$\pm$0.015\\

Autoformer & 119.560$\pm$1.709 & 0.234$\pm$0.009 & 126.347$\pm$4.861 & 0.248$\pm$0.035 & 121.435$\pm$3.529&0.222$\pm$0.022&  128.120$\pm$8.409 & 0.231$\pm$0.044 & 126.358$\pm$4.698 & 0.219$\pm$0.014& 123.926$\pm$10.835 & 0.236$\pm$0.029\\

Informer &130.250$\pm$14.545 & 0.299$\pm$0.037 & 117.671$\pm$4.778 & 0.258$\pm$0.023 & 123.558$\pm$3.572 & 0.257$\pm$0.012& 126.003$\pm$4.208 & 0.293$\pm$0.021 & 176.166$\pm$23.971 & 0.546$\pm$0.127 & 263.370$\pm$1.636 & 0.949$\pm$0.022\\

FEDformer & 113.977$\pm$3.710 & 0.217$\pm$0.008 & 113.124$\pm$1.663 & \textcolor{blue}{0.202}$\pm$0.006 & 118.267$\pm$2.329 & 0.198$\pm$0.012 & 119.534$\pm$3.887 & \textbf{0.191}$\pm$0.003 & 117.769$\pm$4.496 & \textcolor{blue}{0.182}$\pm$0.007 & 110.853$\pm$13.213 & 0.201$\pm$0.043\\

iTransformer & 109.782$\pm$1.632 & 0.254$\pm$0.012 & 111.457$\pm$1.788 & 0.248$\pm$0.008& 112.982$\pm$1.086 & 0.238$\pm$0.015 & 113.158$\pm$0.727 & 0.225$\pm$0.006 & 109.275$\pm$1.324 & 0.198$\pm$0.003 & \textbf{96.035}$\pm$0.701 & \textbf{0.173}$\pm$0.002\\

PatchTST & 104.774$\pm$3.070 & 0.235$\pm$0.008 & 106.689$\pm$4.426 & 0.243$\pm$0.026 & 110.472$\pm$4.456 & 0.258$\pm$0.012 & 108.260$\pm$2.974 & 0.219$\pm$0.018 & 108.750$\pm$4.896 & 0.220$\pm$0.022 & 102.828$\pm$3.663 & \textcolor{blue}{0.196}$\pm$0.015\\

TimesNet & 106.745$\pm$2.467 & 0.265$\pm$0.003 & 108.891$\pm$3.191 & 0.270$\pm$0.018& 111.237$\pm$2.051&0.259$\pm$0.010 & 113.144$\pm$4.094 & 0.261$\pm$0.028 & 113.493$\pm$3.783& 0.256$\pm$0.029 & 108.738$\pm$11.966 & 0.255$\pm$0.042\\

DLinear & 99.975$\pm$0.454 & 0.260$\pm$0.001 & 100.289$\pm$0.567 & 0.268$\pm$0.001 & 105.746$\pm$0.625 & 0.270$\pm$0.000 & 107.764$\pm$0.206 & 0.264$\pm$0.000 & 102.819$\pm$0.304 & 0.223$\pm$0.000 & 107.077$\pm$2.675 & 0.246$\pm$0.012\\

TSMixer & \textcolor{blue}{95.888}$\pm$1.797 & \textcolor{blue}{0.197}$\pm$0.008 & \textcolor{blue}{98.980}$\pm$1.424 & \textbf{0.196}$\pm$0.004& \textcolor{blue}{104.834}$\pm$0.292&  \textcolor{blue}{0.195}$\pm$0.005 & \textcolor{blue}{106.230}$\pm$2.010  & \textcolor{blue}{0.193}$\pm$0.010 & 107.009$\pm$2.829 & 0.186$\pm$0.010& 136.102$\pm$22.298 & 0.292$\pm$0.065\\

AutoPV & \textbf{89.865}$\pm$1.418 & \textbf{0.178}$\pm$0.003 & \textbf{94.887}$\pm$1.160 & 0.206$\pm$0.014 & \textbf{102.869}$\pm$1.502 & \textbf{0.179}$\pm$0.007& \textbf{105.084}$\pm$1.549 & 0.199$\pm$0.015 & \textbf{100.508}$\pm$1.155 &  \textbf{0.180}$\pm$0.007 & \textcolor{blue}{102.287}$\pm$2.158  & 0.232$\pm$0.009\\

\hline
\end{tabular}}

\end{table*}

\begin{table*}[!t]
\tiny
\caption{Evaluation results of AutoPV and all selected benchmarks for PVPF task 2 (with future weather information). The input sequence length is fixed as 96 and the prediction lengths are set as different values. Each experiment is conducted 5 times. The best result is highlighted in \textbf{bold}, and the second is highlighted in \textcolor{blue}{blue}.\label{tab:baseline_fw}}
\resizebox{\textwidth}{!}{

\begin{tabular}{c|cc|cc|cc|cc|cc|cc}
\hline
Task & \multicolumn{2}{c|}{DQH-12 (Half Day)} & \multicolumn{2}{c|}{DQH-24 (One Day)}  &\multicolumn{2}{c|}{DQH-48 (Two Days)}  & \multicolumn{2}{c}{DQH-72 (Three Days)}& \multicolumn{2}{c}{DQH-168 (One Week)}& \multicolumn{2}{c}{DQH-336 (One Month)}\\
Model & MAE& WMAPE & MAE& WMAPE & MAE& WMAPE & MAE& WMAPE & MAE& WMAPE & MAE& WMAPE\\
\hline
LSTM & \textcolor{blue}{60.047}$\pm$1.152 & \textcolor{blue}{0.149}$\pm$0.008 & \textcolor{blue}{56.689}$\pm$0.687 & \textcolor{blue}{0.139}$\pm$0.003 & \textcolor{blue}{55.758}$\pm$0.435& \textcolor{blue}{0.132}$\pm$0.002& \textcolor{blue}{54.132}$\pm$1.409 & \textcolor{blue}{0.127}$\pm$0.004 & \textcolor{blue}{52.831}$\pm$1.657 & \textbf{0.113}$\pm$0.003& 79.336$\pm$1.589 & \textcolor{blue}{0.129}$\pm$0.003\\

MLP & 66.555$\pm$1.206 & 0.152$\pm$0.003 & 63.700$\pm$0.829 & 0.147$\pm$0.002 & 63.889$\pm$1.585 & 0.142$\pm$0.003& 67.973$\pm$1.753  & 0.146$\pm$0.004& 72.463$\pm$1.462& 0.152$\pm$0.002& 92.405$\pm$6.545 & 0.154$\pm$0.014\\

CNN & 65.839$\pm$1.727 & 0.154$\pm$0.003 & 61.296$\pm$0.820 & 0.148$\pm$0.002 & 58.680$\pm$0.717 & 0.139$\pm$0.002 & 57.582$\pm$0.875 & 0.138$\pm$0.003 & 56.961$\pm$1.226 & 0.126$\pm$0.002& \textcolor{blue}{76.315}$\pm$1.680 & 0.130$\pm$0.003\\

TCN & 104.235$\pm$3.114 & 0.189$\pm$0.005 & 91.892$\pm$2.068 &0.182$\pm$0.003& 85.015$\pm$2.030 & 0.175$\pm$0.002 & 80.460$\pm$0.693 &  0.170$\pm$0.002 & 93.282$\pm$9.748 & 0.163$\pm$0.006& 1758.367$\pm$143.650 & 1.864$\pm$0.179\\

Transformer & 76.503$\pm$5.776 & 0.183$\pm$0.017 & 71.331$\pm$3.263 & 0.180$\pm$0.005 & 65.233$\pm$1.971 & 0.162$\pm$0.010 &  79.034$\pm$17.013 & 0.180$\pm$0.028 & 105.644$\pm$26.242 & 0.246$\pm$0.070& 231.193$\pm$64.070 & 0.795$\pm$0.295\\

Autoformer & 87.423$\pm$6.473 & 0.187$\pm$0.008 & 96.632$\pm$4.223 & 0.248$\pm$0.035 & 106.222$\pm$12.879 &0.201$\pm$0.016&  106.778$\pm$10.874 & 0.212$\pm$0.018 & 105.826$\pm$4.426 & 0.184$\pm$0.012 & 96.595$\pm$13.050 & 0.180$\pm$0.043\\

Informer &84.727$\pm$8.453 & 0.203$\pm$0.025 & 76.643$\pm$2.796 & 0.202$\pm$0.006 & 78.749$\pm$2.848 & 0.205$\pm$0.007& 98.813$\pm$32.636 & 0.276$\pm$0.125 & 209.240$\pm$44.571 & 0.774$\pm$0.247 & 263.331$\pm$2.248 & 0.950$\pm$0.008\\

FEDformer & 82.617$\pm$5.005 & 0.172$\pm$0.008 & 83.332$\pm$2.287 & 0.167$\pm$0.007 & 88.092$\pm$6.028 & 0.170$\pm$0.006 & 92.605$\pm$4.601 & 0.180$\pm$0.008 & 105.320$\pm$2.501 & 0.189$\pm$0.007 & 136.953$\pm$50.086 & 0.347$\pm$0.220\\

iTransformer & 73.102$\pm$2.296 & 0.179$\pm$0.005 & 69.644$\pm$1.371 & 0.170$\pm$0.003 & 65.071$\pm$0.408 & 0.153$\pm$0.002 & 65.751$\pm$1.660 & 0.151$\pm$0.002 & 66.641$\pm$1.035 & 0.145$\pm$0.003 & 113.858$\pm$8.923 & 0.222$\pm$0.034\\

PatchTST & 104.500$\pm$4.728 & 0.233$\pm$0.007 & 102.964$\pm$1.600 & 0.227$\pm$0.009 & 108.040$\pm$3.738 & 0.224$\pm$0.020 & 108.285$\pm$2.685 & 0.229$\pm$0.023 & 110.355$\pm$1.538 & 0.216$\pm$0.009 & 112.741$\pm$11.564 & 0.211$\pm$0.031\\

TimesNet & 71.038$\pm$3.431 & 0.177$\pm$0.010 & 65.703$\pm$6.692 & 0.167$\pm$0.016 & 59.014$\pm$2.341 &0.147$\pm$0.013 & 60.839$\pm$2.073 & 0.145$\pm$0.004 & 58.973$\pm$3.392& 0.141$\pm$0.011 & 108.738$\pm$11.966 & 0.255$\pm$0.042\\

DLinear & 100.279$\pm$0.913 & 0.262$\pm$0.002 & 100.199$\pm$0.390 & 0.268$\pm$0.001 & 105.236$\pm$0.409& 0.271$\pm$0.000 & 107.233$\pm$0.240 & 0.265$\pm$0.001 & 103.352$\pm$0.389 & 0.223$\pm$0.000 & 108.505$\pm$1.777 & 0.252$\pm$0.006\\

TSMixer & 65.670$\pm$2.994 & 0.158$\pm$0.009 & 63.341$\pm$1.809 & 0.145$\pm$0.004& 60.722$\pm$0.284&  0.138$\pm$0.002 & 60.091$\pm$0.992  & 0.136$\pm$0.003 & 63.335$\pm$2.235 & 0.130$\pm$0.004& 84.873$\pm$6.073 & 0.147$\pm$0.016\\

AutoPV & \textbf{57.405}$\pm$1.306 & \textbf{0.140}$\pm$0.004 & \textbf{54.607}$\pm$0.436 & \textbf{0.135}$\pm$0.003 & \textbf{51.582}$\pm$0.719 & \textbf{0.128}$\pm$0.003& \textbf{50.109}$\pm$0.531 & \textbf{0.121}$\pm$0.002 & \textbf{51.118}$\pm$1.221 &  \textcolor{blue}{0.124}$\pm$0.004 & \textbf{58.525}$\pm$3.214  & \textbf{0.117}$\pm$0.004\\

\hline
\end{tabular}}

\end{table*}

For each task on each model, we fix a global random seed and conduct the training-evaluation experiments five times. The evaluation results including 
average values and the standard deviations are presented in Table \ref{tab:baseline} for PVPF Task 1 and Table \ref{tab:baseline_fw} for PVPF Task 2.

\textbf{1) Analysis of PVPF Task 1.} For PVPF Task 1, as shown in Table \ref{tab:baseline}, in the first five sub-tasks, the AutoPV system successfully finds an optimal architecture that outperforms other predefined models. In the remaining task DQH-336, It is only surpassed by iTransformer. This demonstrates that within the current search space, the AutoPV system is able to construct effective architectures for PVPF tasks with varying forecasting lengths. Based on the average MAE across six sub-tasks, we select the top-performing model from each benchmark category: CNN, PatchTST, and DLinear. When compared to these three models, the AutoPV system demonstrates relative performance improvements in MAE of 9.02\%, 7.21\%, and 4.52\%, respectively.

Additionally, it is evident that non-Transformer-based models generally outperform most Transformer-based models, especially in short-term and medium-term prediction tasks. This help to suggest that constructing the PVPF search space based on non-Transformer models is reasonable and feasible. Regarding the long-term forecasting tasks (DQH-168 and DQH-336), the latest Transformer variants, PatchTST and iTransformer, begin to show performance advantages, likely because transformer-based models are better at handling of long sequences. Despite this, the AutoPV system still achieves relatively close performance by combining some simple non-transformer elements.

\textbf{2) Analysis of PVPF Task 2.} Table \ref{tab:baseline_fw} presents the forecasting results in PVPF Task 2 where the future weather information is incorporated. The results demonstrate that the weather data appending method described in Section \ref{problemdescription} significantly enhances the forecasting performance for most models. Notably, for DLinear and PatchTST, this data appending method does not yield performance improvements. We attribute this to the following reasons: For PatchTST, the attention mechanism between each patch (multiple tokens) is disrupted by zero padding, which affects the attention calculation for the future weather data. In the case of DLinear, since it primarily considers the periodic time series trends of each feature, appending zeros to the power feature does not introduce new trend information.

In comparison, AutoPV has the lowest forecasting error in PVPF Task 2 across all six sub-tasks. We also calculated the performance improvements of AutoPV over three top-performing models: LSTM, iTransformer, and TSMixer. The relative improvements in MAE are 9.88\%, 28.79\%, and 18.76\%, respectively. Compared to PVPF Task 1, AutoPV shows an even greater performance advantage in PVPF Task 2.

The standard deviations in Table \ref{tab:baseline_fw} reveal that the forecasting performance becomes highly unstable for long-term tasks in many models, especially the transformer-based models. This instability is due to the future weather data often being noisy and becoming increasingly unreliable as the forecasting length extends. In contrast, the prediction results of AutoPV exhibit significantly greater stability than most baseline models.

\textbf{3) Comprehensive Analysis.} After analyzing the results in Table \ref{tab:baseline} and Table \ref{tab:baseline_fw}, we observe that the predefined models TSMixer and LSTM also exhibit relatively good performance in PVPF Task 1 and PVPF Task 2, respectively. However, there are two reasons that underscore the necessity of AutoPV. Firstly, as shown in Table \ref{tab:baseline}, TSMixer loses its performance edge in Task DQH-168 and even utterly fails in Task DQH-336. Similarly, in Table \ref{tab:baseline_fw}, the performance of LSTM deteriorates significantly. In contrast, AutoPV consistently maintains high-quality performance across different tasks. Additionally, prior to evaluation, it is uncertain which model will perform well. Even though TSMixer is good at PVPF Task 1, it is outperformed by other models at PVPF Task 2. This indicates that for a new specific task, evaluation experiments are still necessary to select the best model. AutoPV not only automates the process of model construction, effectively reducing labor and time costs, but also provides better-performing models, thereby increasing the accuracy of forecasts.

\begin{figure*}[!t]
\centering
\includegraphics[width=7in]{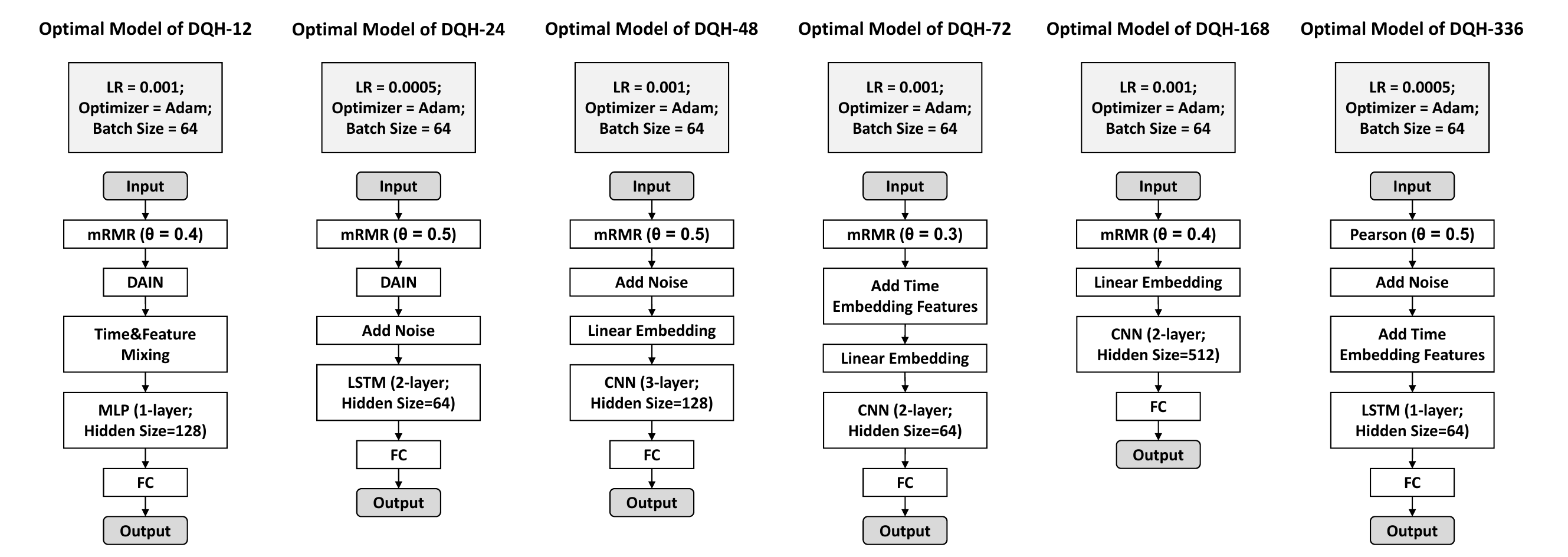}
\caption{The complete architecture searched by AutoPV for the different sub-tasks in PVPF Task 1.}
\label{arc1}
\end{figure*}

\begin{figure*}[!t]
\centering
\includegraphics[width=7in]{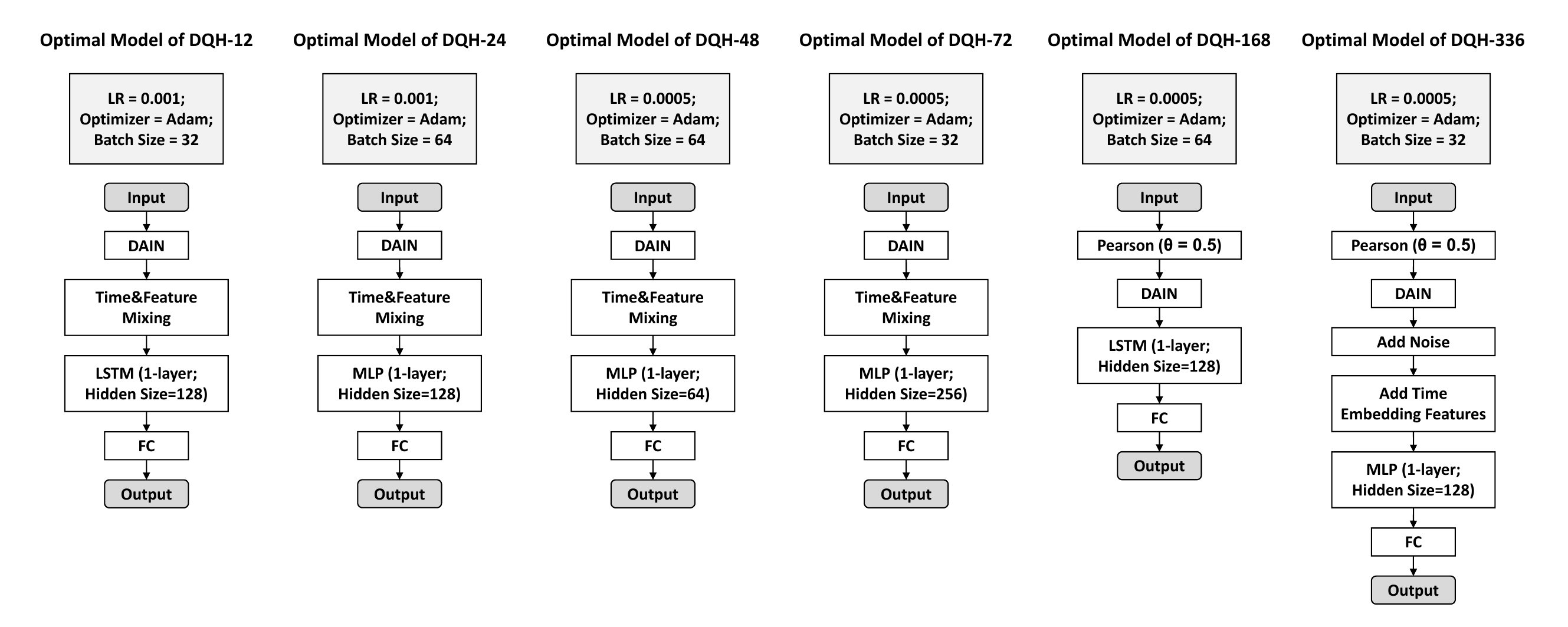}
\caption{The complete architecture searched by AutoPV for the different tasks in PVPF Task 2.}
\label{arc2}
\end{figure*}

\textbf{Architectures Visualization.} Figs. \ref{arc1} and \ref{arc2} illustrate the structures of all the searched architectures. The results indicate that for PVPF Task 1, the distribution of modules within the architectures is scattered. Different sub-tasks employ various training settings, data processing methods, and core predictive models. In contrast, for PVPF Task 2, the $FEM_5$ (Time \& Feature Mixing) module appears with high frequency. Additionally, a 1-layer MLP or 1-layer LSTM structure is predominantly chosen. To demonstrate the effectiveness of module selection in constructing the search space, we present the module proportions of all 12 parameters in Appendix, based on statistics from the optimal architecture set (Pareto frontier) for all tasks. The result shows that, except for the $OF_2$ (SGD optimizer) module, all modules have a non-zero occurrence rate. In future work, we may optimize the search space by removing less frequently used modules and adding more powerful ones.


\begin{figure*}
  \centering
 \subfloat[DQH-12]{\includegraphics[width=2.3in]{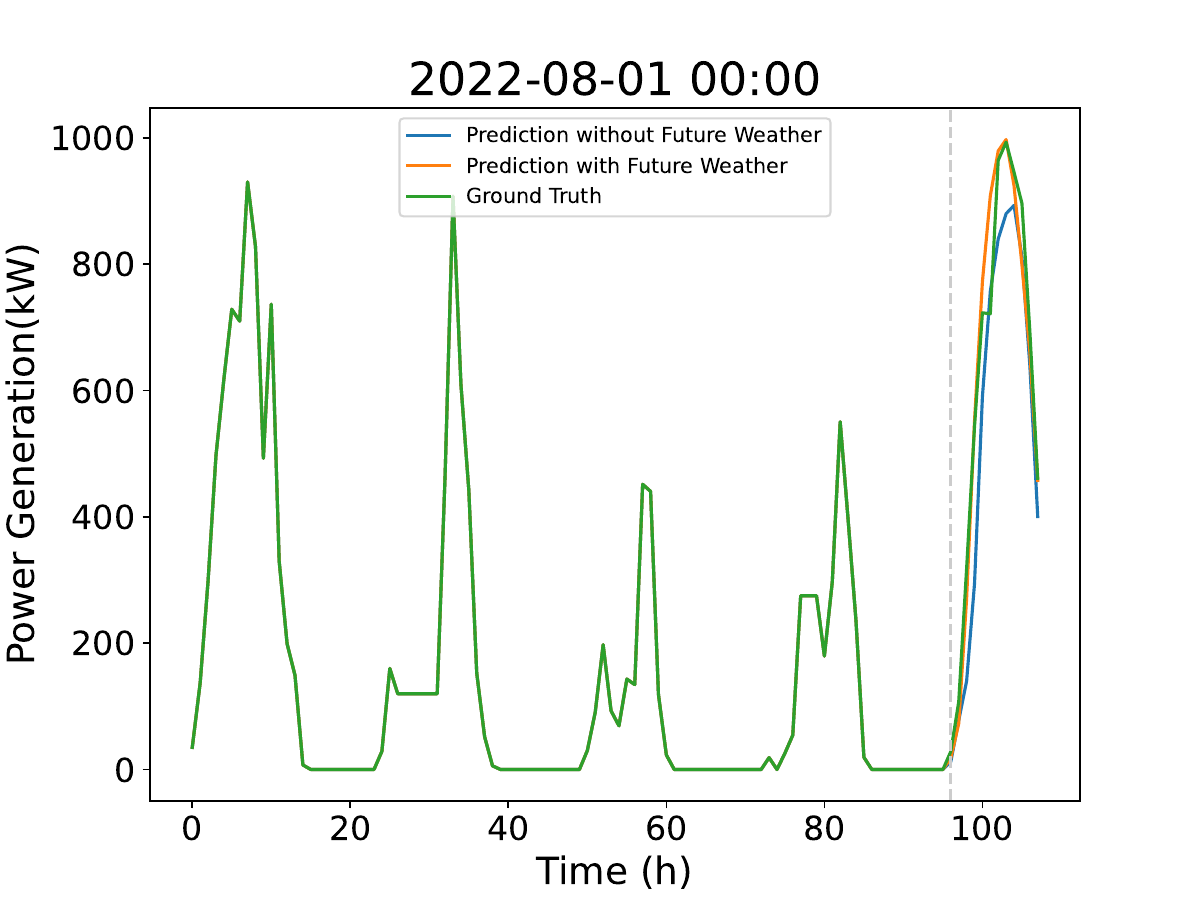}}
  \hfil
  \subfloat[DQH-24]{\includegraphics[width=2.3in]{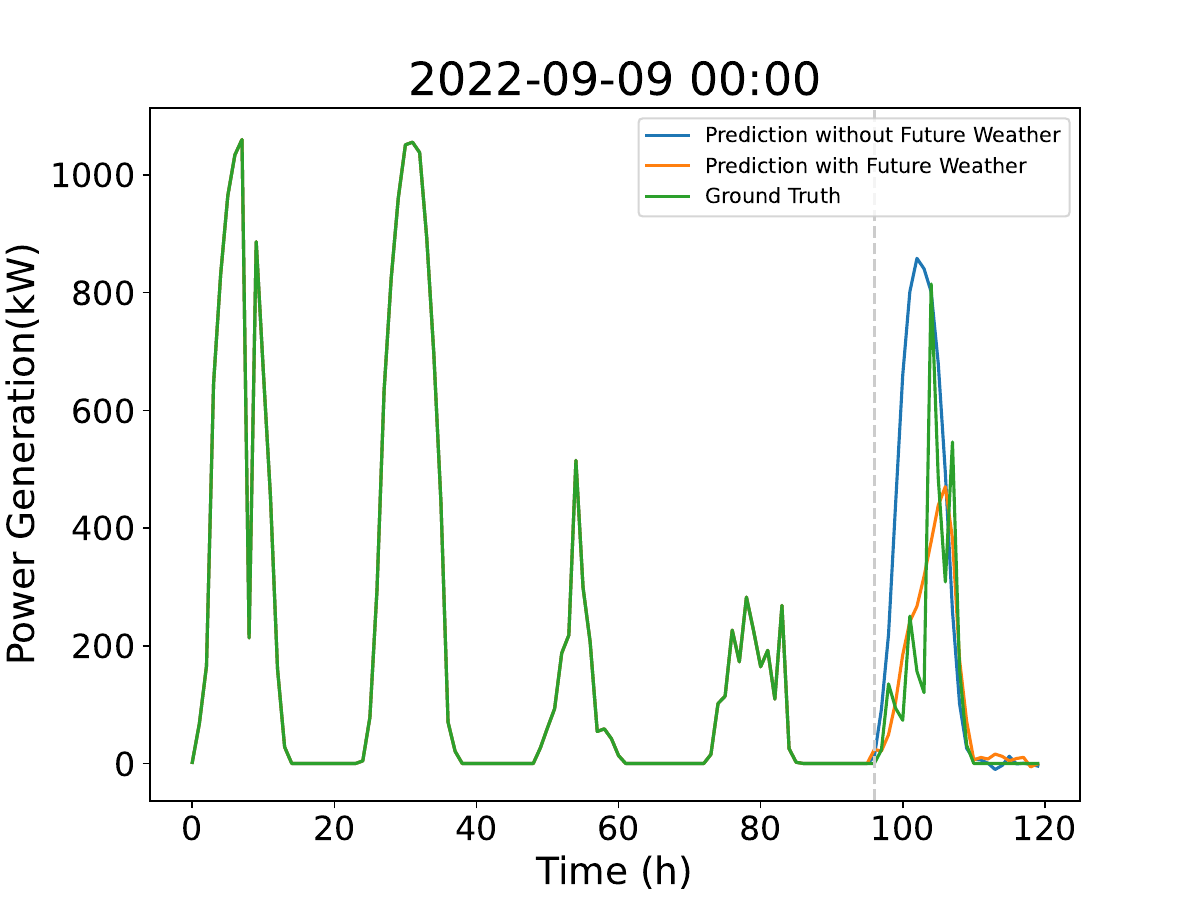}}
    \hfil
  \subfloat[DQH-48]{\includegraphics[width=2.3in]{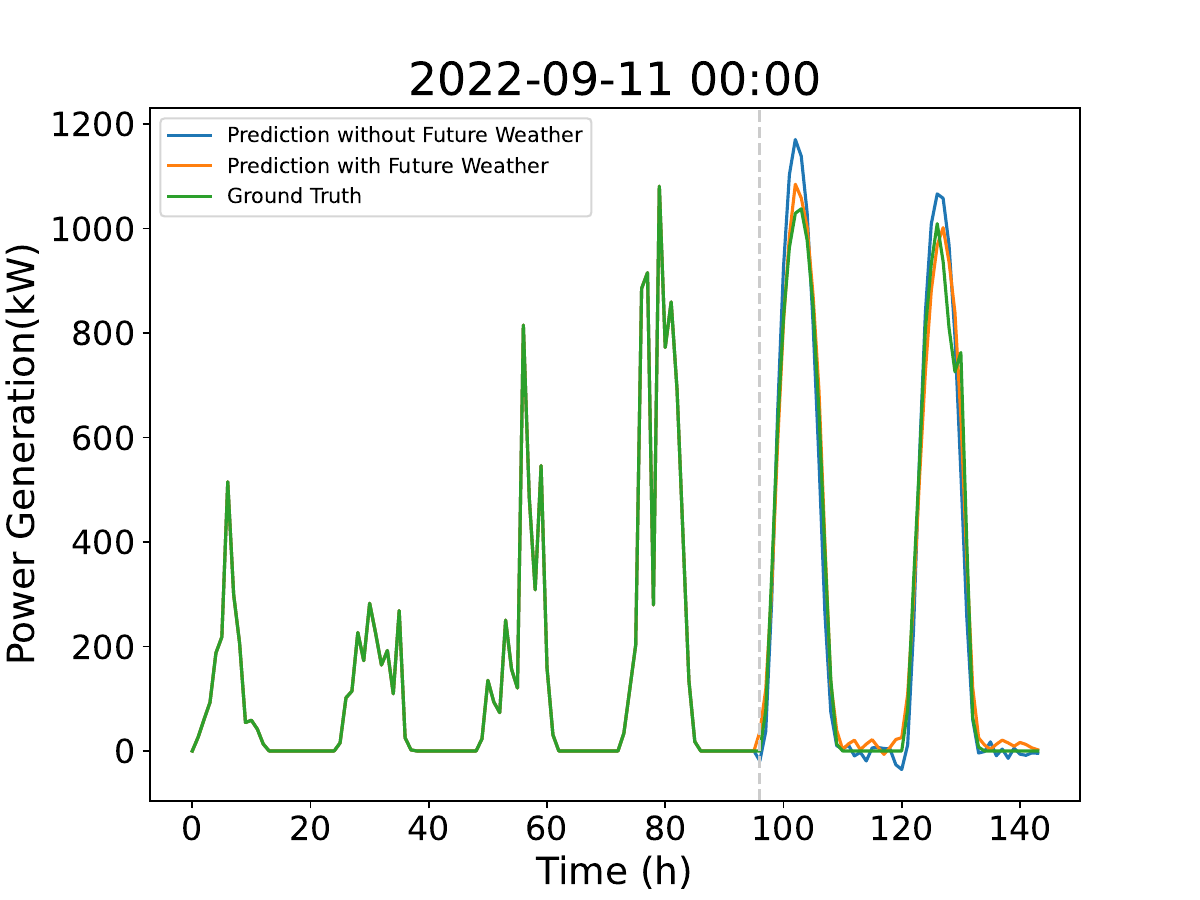}}
    \hfil
  \subfloat[DQH-72]{\includegraphics[width=2.3in]{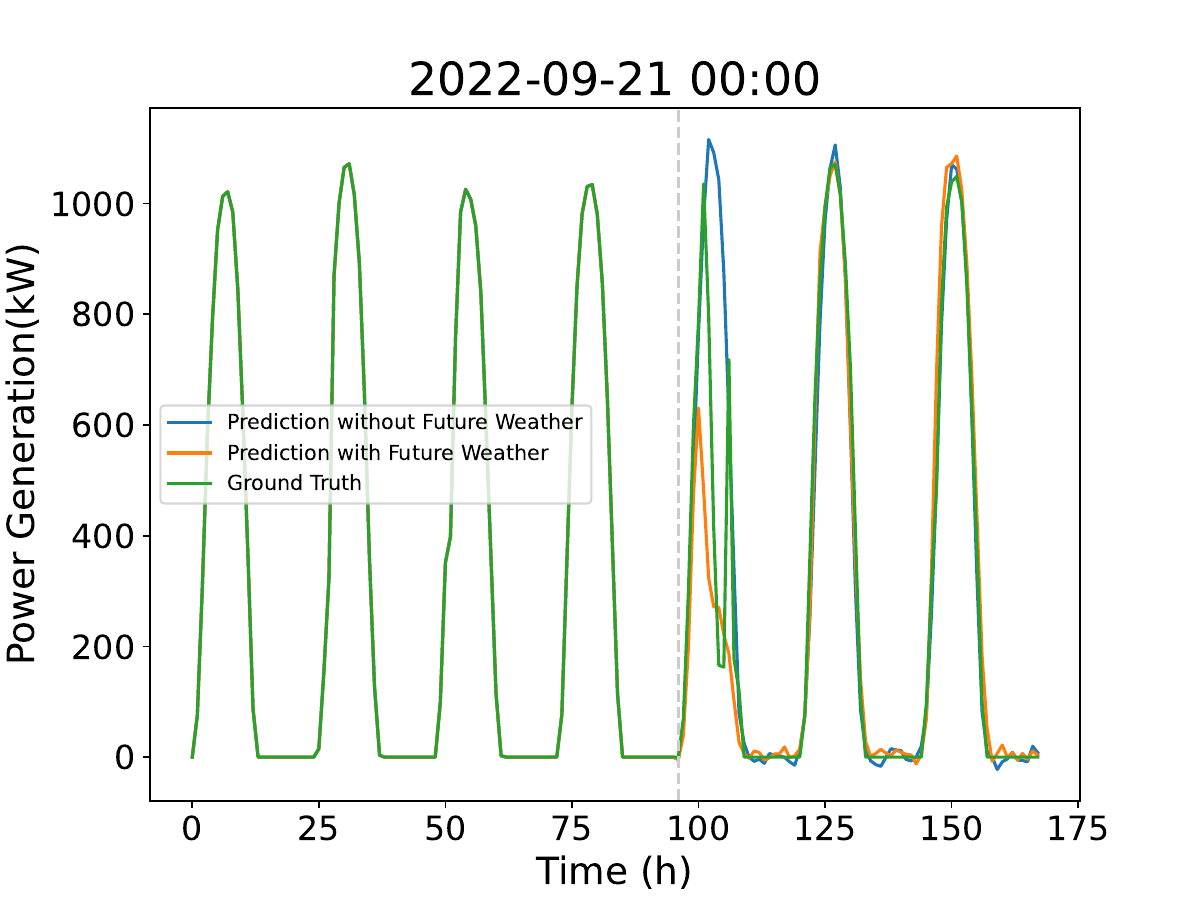}}
    \hfil
  \subfloat[DQH-168]{\includegraphics[width=2.3in]{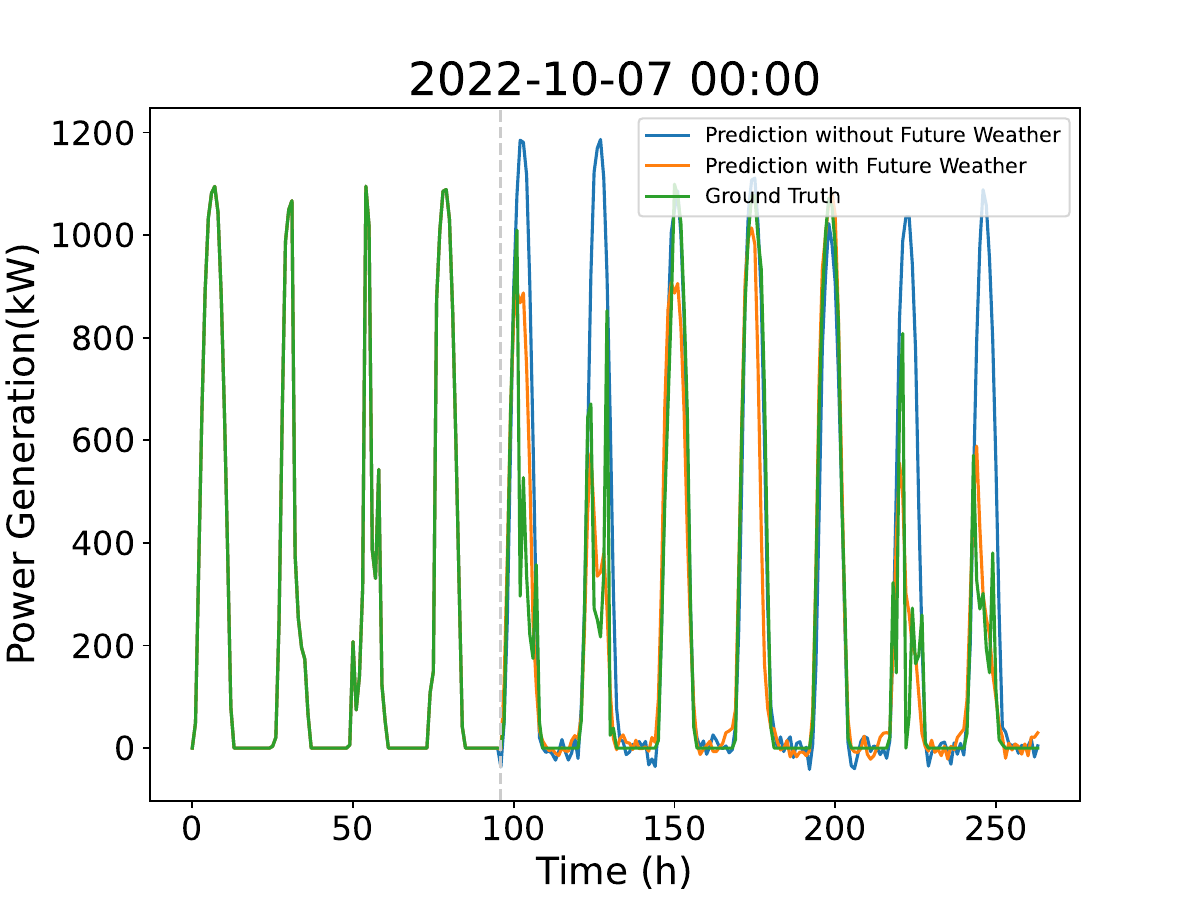}}
    \hfil
  \subfloat[DQH-336]{\includegraphics[width=2.3in]{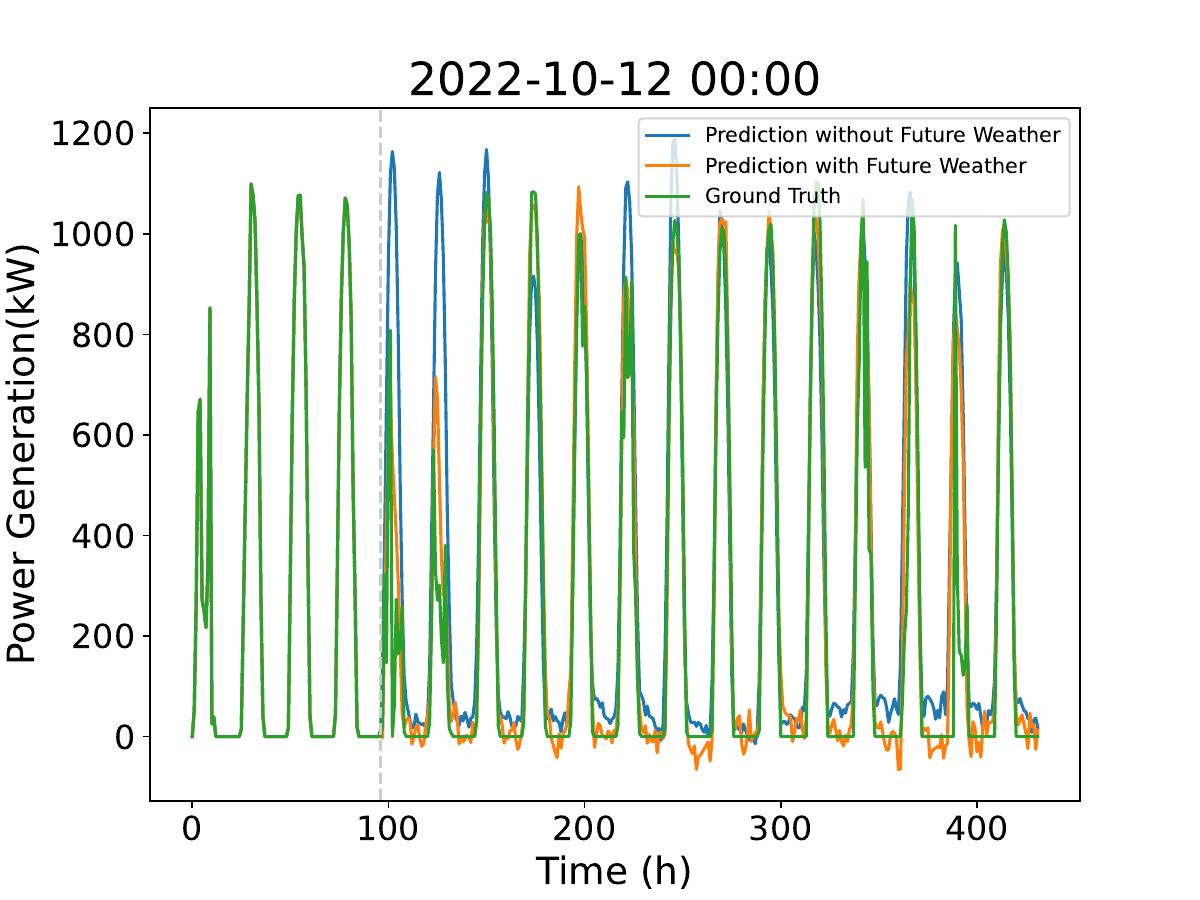}}

  \caption{Visualization of AutoPV forecasting performance: with and without future weather data}
  \label{pred}
\end{figure*}


\textbf{Forecasting Performance Visualization.} In Fig. \ref{pred}, we present the forecasting results of AutoPV across all sub-tasks, both with and without future weather data utilization. The results clearly demonstrate that the periodic pattern of photovoltaic power output is effectively captured. For PV power forecasting on sunny days, the inclusion of future weather data does not significantly impact the results. While on rainy and cloudy days, forecasting PV power generation proves to be more challenging without future weather data. Even when the future weather data is contaminated with noise, it still contributes to a notable improvement in the accuracy of future PV power forecasting.

\section{Conclusion}
In this study, we propose AutoPV, an NAS-based framework designed to automatically search and construct efficient TSF models for specific PVPF tasks. We develop a novel search space that incorporates many state-of-the-art TSF and PVPF models. Utilizing a multi-objective optimization method named MoBananas, we search for the best-performing architecture while also controlling the model scale. To the best of our knowledge, this is the first work that bridges the gap between NAS and TSL tasks. Extensive experiments demonstrate the efficiency and effectiveness of AutoPV across various PVPF tasks. In future work, we plan to expand the search space and include more valuable modules to further enhance the performance of AutoPV.

\section*{Acknowledgments}
This work is supported by the RISUD Projects (No. P0042845 and No. 1-BBWW) and the RISE Project (No. P0051003) of The Hong Kong Polytechnic University, International Centre of Urban Energy Nexus of The Hong Kong Polytechnic University (No. P0047700), and KKS Synergy project, Energy flexibility through synergies of big data, novel technologies \& systems, and innovative markets (20200073). It is also supported by High Performance Computing Center at Eastern Institute of Technology, Ningbo. 



 
%

\bibliographystyle{IEEEtran}
\bibliography{main}

\newpage

\section{Biography Section}
 

\begin{IEEEbiography}[{\includegraphics[width=1in,height=1.25in,clip,keepaspectratio]{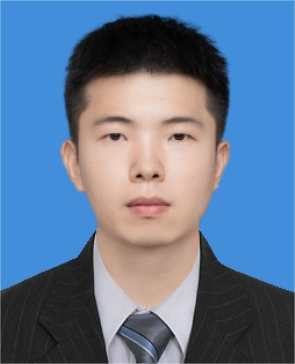}}]{Dayin Chen} is currently pursuing his Ph.D. in Building Energy and Environment Engineering at The Hong Kong Polytechnic University. He holds both a Bachelor's and a Master's degree in Computer Science from the Southern University of Science and Technology. His research interests include mobile computing, crowdsourcing, and neural architecture search. Additionally, Chen is exploring the application of computer science technologies in enhancing building energy efficiency.
\end{IEEEbiography}

\begin{IEEEbiography}[{\includegraphics[width=1in,height=1.25in,clip,keepaspectratio]{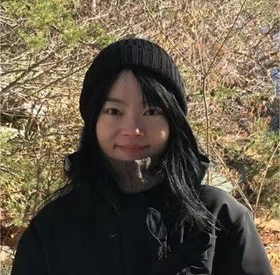}}]{Xiaodan Shi} received the B.E. and M.S. degrees in photogrammetry and remote sensing from Wuhan University, China. She  received the Ph.D. degree with the Center for Spatial Information Science, The University of Tokyo, Kashiwa, Japan. She is working as a researcher with the Center for Spatial Information Science, The University of Tokyo. Her current research interests include computer vision and its applications in pedestrian trajectory prediction, human flow monitoring, multi-objects tracking and data mining in GPS.
\end{IEEEbiography}

\begin{IEEEbiography}[{\includegraphics[width=1in,height=1.25in,clip,keepaspectratio]{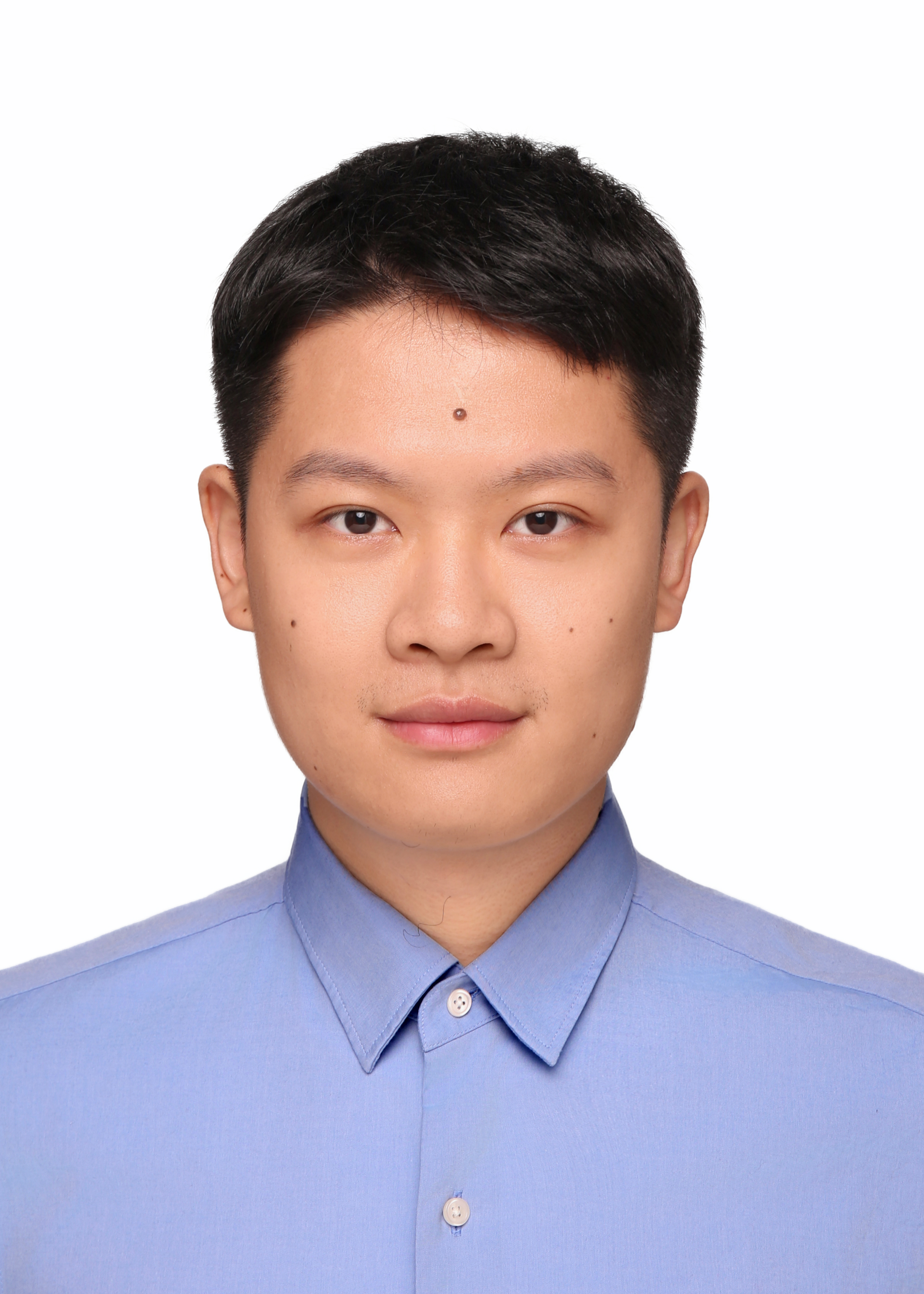}}]{Mingkun Jiang} received the B.E. degree in process equipment and control engineering from Yanshan University, China. He received the M.S. and Ph.D. degrees with the School of Mechanical and Power Engineering, The East China University of Science and Technology, Shanghai, China. He is working as a research engineer with the PV Industry Innovation Center, The State Power Investment Corporation. His current research interests include renewable energy potential evaluation, PV and storage system modelling and its power output prediction.
\end{IEEEbiography}

\begin{IEEEbiography}[{\includegraphics[width=1in,height=1.25in,clip,keepaspectratio]{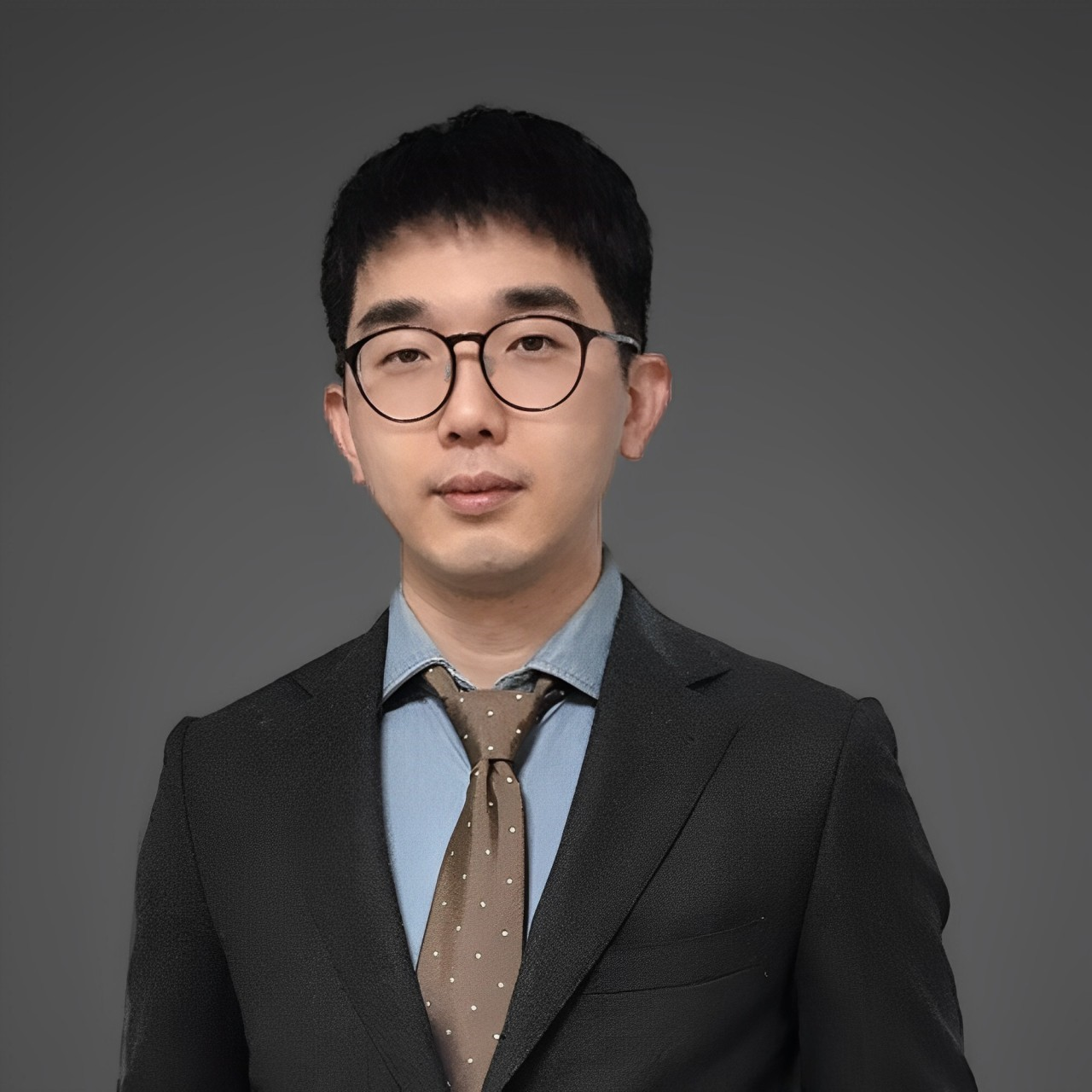}}]{Haoran Zhang} is a researcher with an interdisciplinary background, holding dual Bachelor's degrees in Engineering and Economics, complemented by two Ph.D. degrees in Oil \& Gas Storageand Transportation Engineering ond Environmental science. His research is at the forefront of sustainable development, focusing on clean energy supply chains, green transportation, and smart urban energy systems. His innovative research has garnered prestigious accolades, including the Energy Globe Award, the R\&D 100 Award, the John Tiratsoo Award for Young Achievement, and the Smart 50 Award among others.
\end{IEEEbiography}


\begin{IEEEbiography}[{\includegraphics[width=1in,height=1.25in,clip,keepaspectratio]{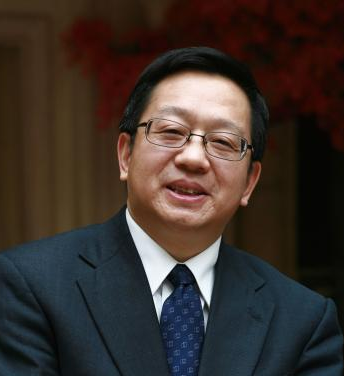}}]{Dongxiao Zhang} received the M.S. and Ph.D. degrees in hydrology from the University of Arizona, Tucson, AZ, USA, in 1992 and 1993, respectively. From 1996 to 2003, he was a Technical Staff Member and the Team Leader with the Los Alamos National Laboratory, Los Alamos, NM, USA. From 2004 to 2007, he was the Miller Chair Professor of petroleum and geological engineering with the University of Oklahoma, Norman, OK, USA. From 2007 to 2010, he was the Marshall Professor of the
Petroleum Engineering Program with the University of Southern California, Los Angeles, CA, USA. From 2010 to 2019, he was the Chair Professor and the Dean of the College of Engineering, Peking University, Beijing, China. He was the Provost and the Chair Professor with the Southern University of Science and Technology, Shenzhen, China. He is currently the Provost and the Chair Professor with the Eastern Institute of Technology, Ningbo, China. His research interests include stochastic uncertainty quantification and inverse modeling, mechanisms for shale–gas and coalbed–methane production, and geological carbon sequestration. Dr. Zhang is a member of the U.S. National Academy of Engineering, a fellow of the Geological Society of America, and an Honorary Member of the Society of Petroleum Engineers.
\end{IEEEbiography}

\begin{IEEEbiography}[{\includegraphics[width=1in,height=1.25in,clip,keepaspectratio]{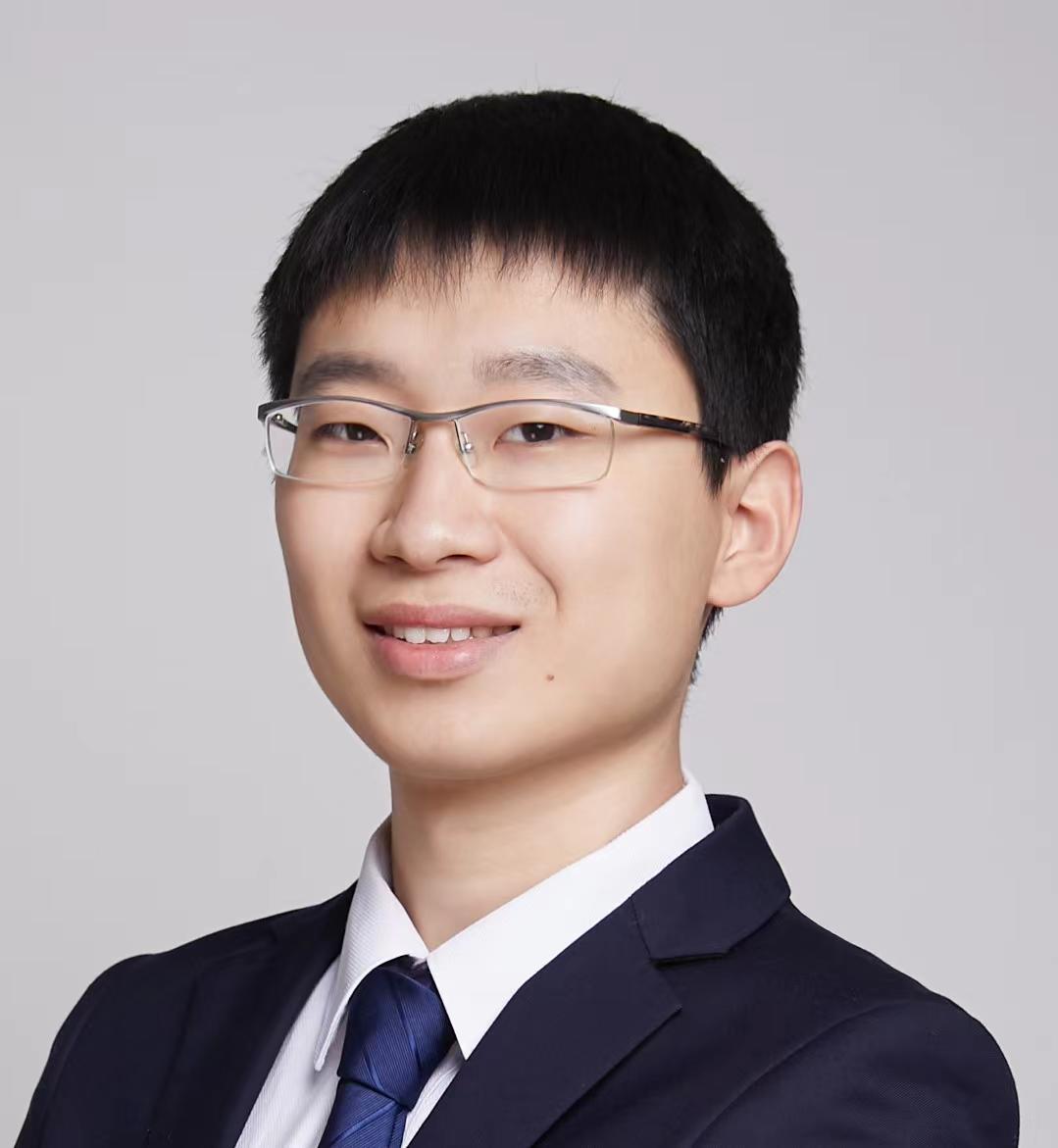}}]{Yuntian Chen} is an assistant professor at Eastern Institute of Technology, Ningbo. He received the B.S. degree from Tsinghua University, Beijing, China, in 2015, the dual B.S. degree from Peking University, Beijing, China, in 2015, and the Ph.D. degree with merit from Peking University, Beijing, China, in 2020. His research field includes scientific machine learning and intelligent energy systems. He
is interested in the integration of domain knowledge and data-driven models.
\end{IEEEbiography}

\begin{IEEEbiography}[{\includegraphics[width=1in,height=1.25in,clip,keepaspectratio]{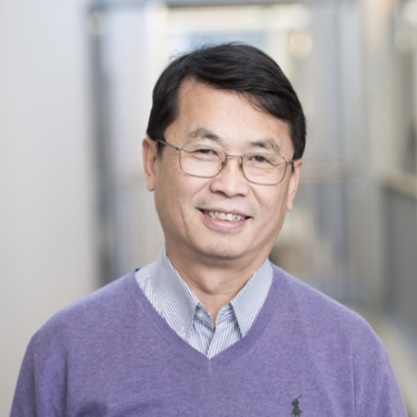}}]{Jerry Yan} is a member of the European Academy of Sciences and Arts and the Fellow of Hong Kong Academy of Engineering Sciences, currently serving as a Chair-Professor at the Hong Kong Polytechnic University. With a PhD from the Royal Institute of Technology (KTH), he has held chair professor positions at Luleå University of Technology, Mälardalen University, and KTH, Sweden. Prof. Yan's research focuses on renewable energy, advanced energy systems, climate change mitigation, and environmental policies. He has a publication record of over 500 papers in renowned journals such as Science, Nature Energy, Nature Climate Change etc and holds more than 10 patents. Having supervised nearly 200 post-doctoral researchers and 50 doctoral candidates, Prof. Yan has secured substantial external grants exceeding 20 million Euros. Prof. Yan's contributions have been acknowledged through prestigious awards, including the Global Human Settlements Award of Green Technology, the EU Energy Islands' Award, Research2Business Top100 and the IAGE Lifetime Achievement Award.
\end{IEEEbiography}

\end{document}


\title{AutoPV: Automatically Design Your PV Power Forecasting Model}

\author{Dayin~Chen,
        Xiaodan~Shi{\IEEEauthorrefmark{1}},
        Mingkun~Jiang,
        Haoran~Zhang,~\IEEEmembership{Senior Member,~IEEE,}       
        Dongxiao~Zhang{\IEEEauthorrefmark{1}},
        Yuntian~Chen,~\IEEEmembership{Member,~IEEE,}
        and Jinyue~Yan{\IEEEauthorrefmark{1}}

}






\begin{table*}[!t]
\caption{Fixed Hyperparameters\label{tab:fixparas}}
\centering
\begin{tabular}{c|c}
\hline
\bf{Module} & \bf{Fixed Parameters}\\
\hline
$\mathbf{DGM}_2$: Gaussian Noise & mean = 0, standard deviation = 0.05 \\
$\mathbf{SM}_3$: DAIN & mean learning rate = 0.00001, gate learning rate = 0.001, scale learning rate = 0.0001 \\

$\mathbf{FEM}_2$: Linear Embedding & in\_features = input size, out\_features = 2 $\times$ input size \\
$\mathbf{FEM}_3$: Decomposition & kernel size = round\_odd(historical length / 3) \\
$\mathbf{FEM}_4$: Multi-Scale Decomposition & kernel sizes = [round\_odd(historical length / $3^1$), round\_odd(historical length / $3^2$), ...] \\
$\mathbf{FEM}_5$: Time\&Feature Mixing & mixer layer number = 2, model dimension = 256, dropout rate = 0.1 \\
$\mathbf{FEM}_6$: Frequency Domain Mixing & embedded size = 128, channel independence = True, sparsity threshold = 0.01, scale = 0.02\\
$\mathbf{CPS}_3$: CNN & convolution kernel size = 3, padding = 1, pool kernel size = 2\\
$\mathbf{CPS}_4$: TCN & kernel size = 2, dropout rate = 0.1, dilation size = 2, 4, 8 (for different layer), stride = 1\\
\hline
\end{tabular}
\end{table*}



\begin{figure*}
  \centering
 \subfloat[PVPF Task 1: DQH-12]{\includegraphics[width=2.3in]{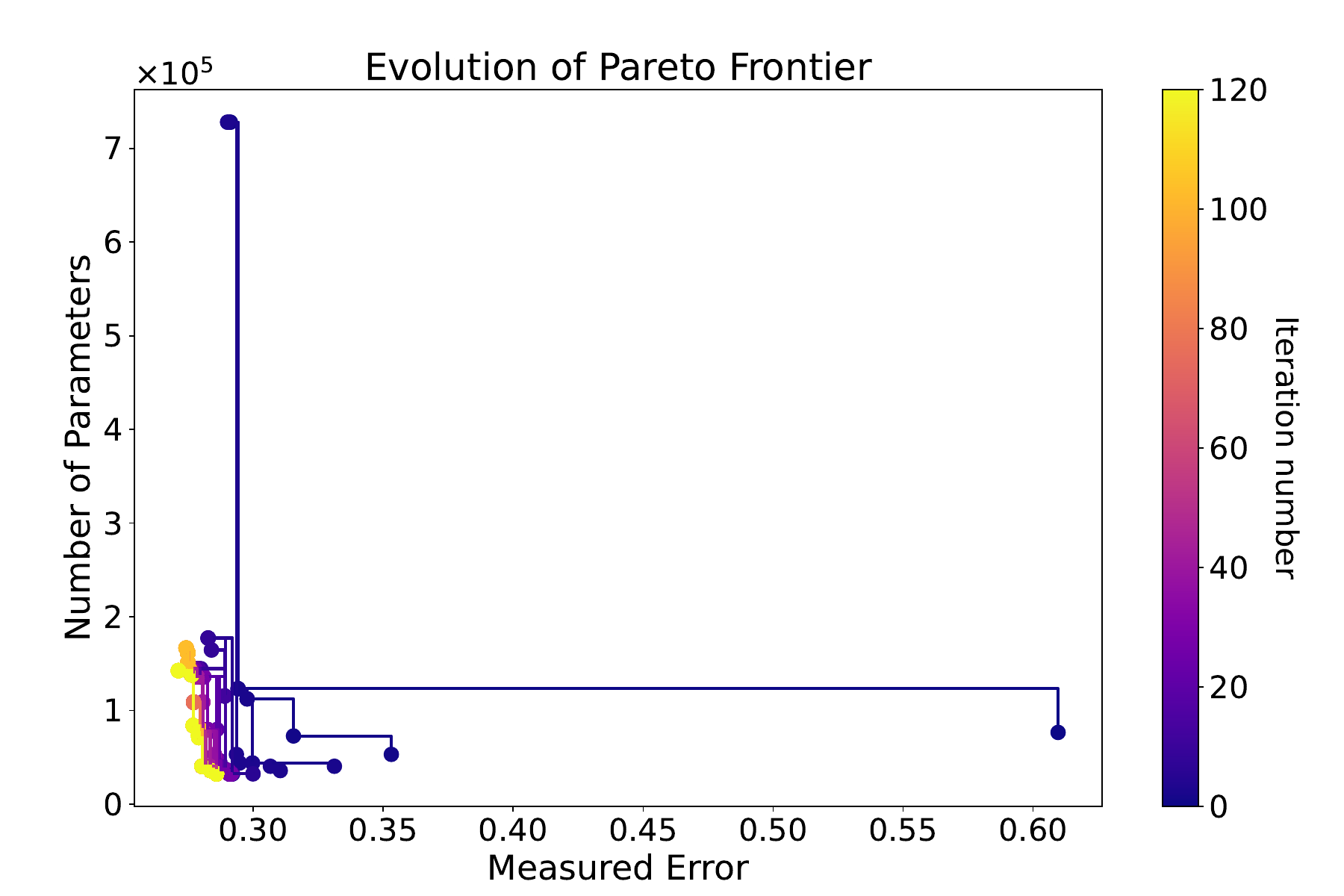}}
  \hfil
  \subfloat[PVPF Task 1: DQH-24]{\includegraphics[width=2.3in]{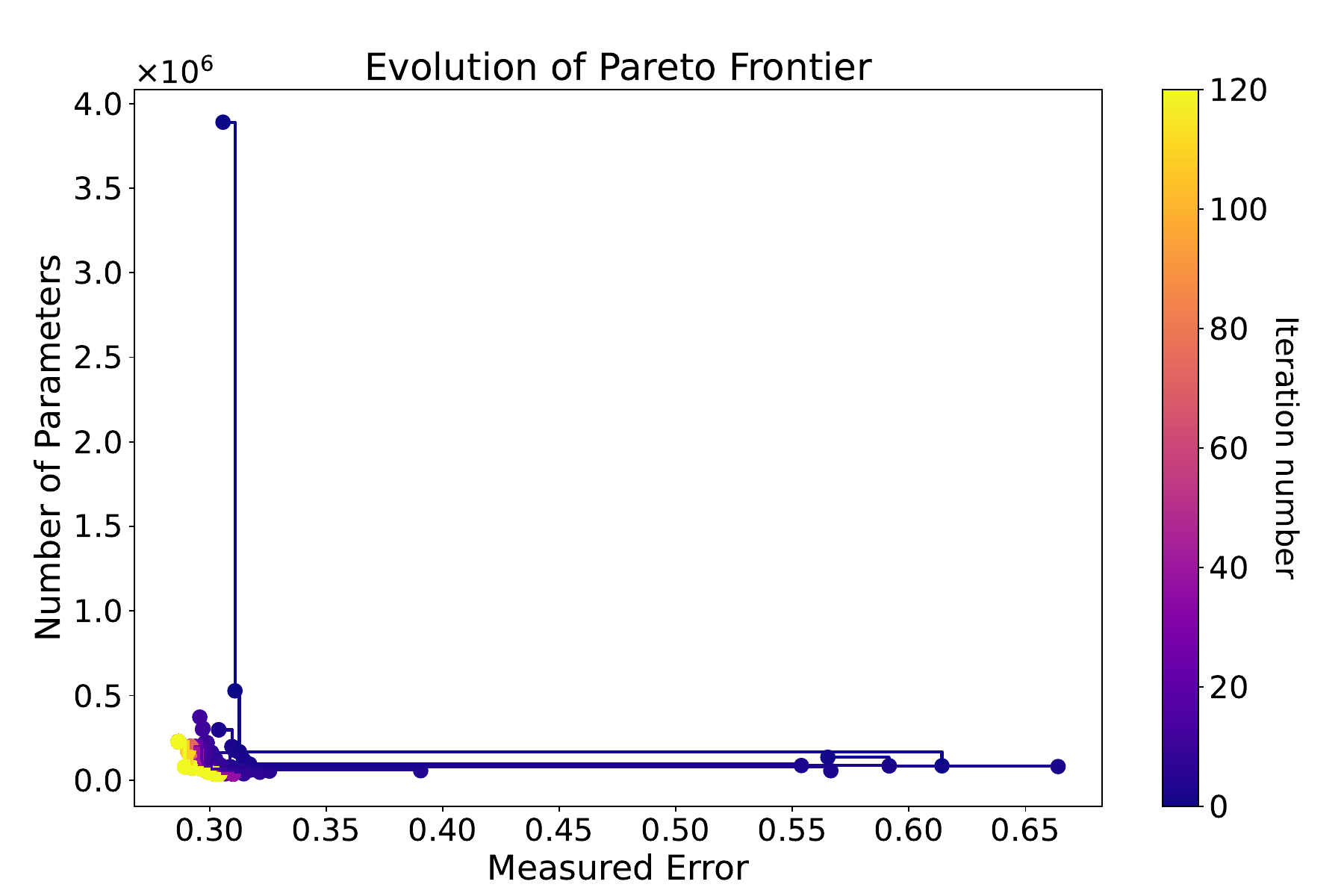}}
    \hfil
  \subfloat[PVPF Task 1: DQH-48]{\includegraphics[width=2.3in]{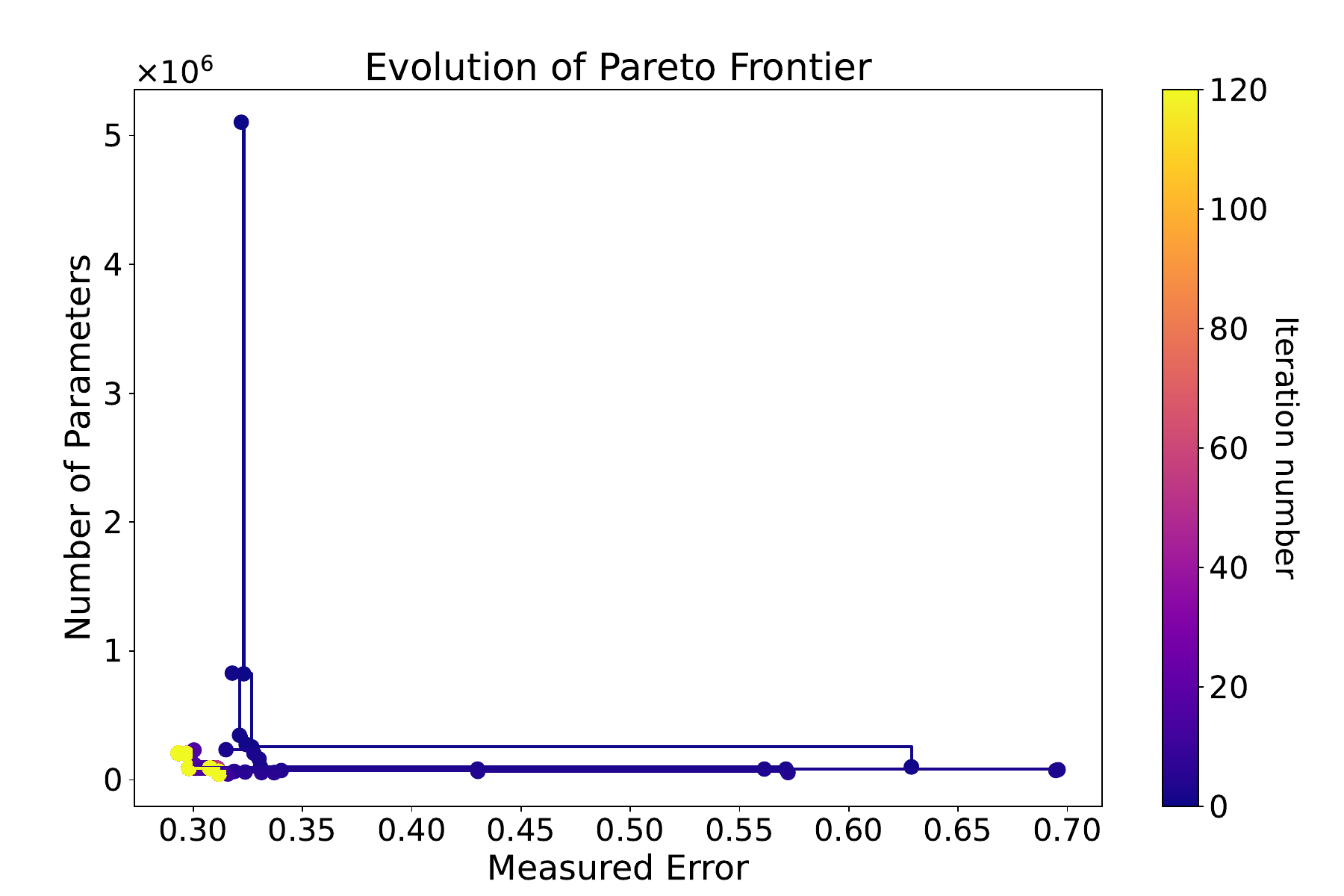}}
    \hfil
  \subfloat[PVPF Task 1: DQH-72]{\includegraphics[width=2.3in]{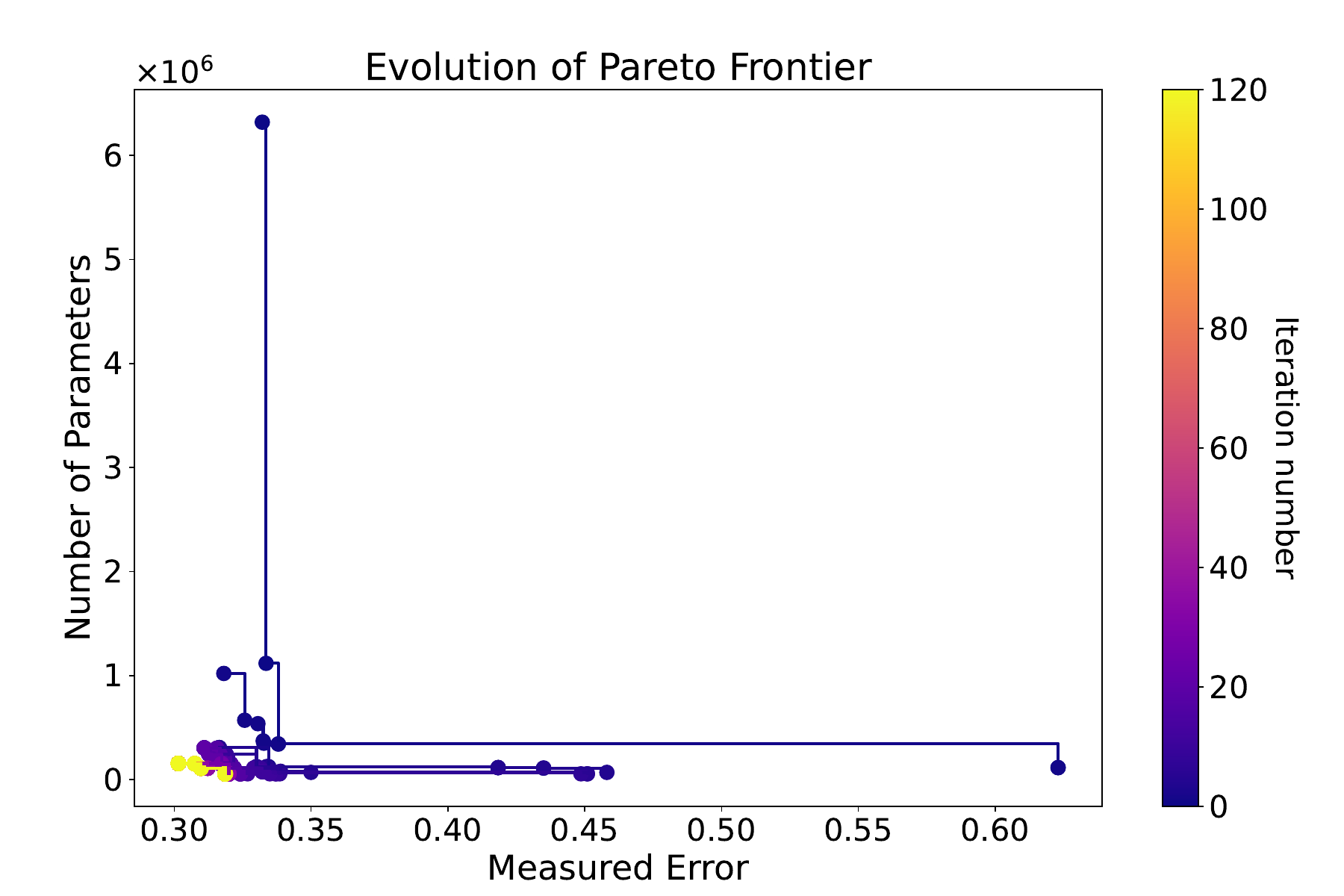}}
    \hfil
  \subfloat[PVPF Task 1: DQH-168]{\includegraphics[width=2.3in]{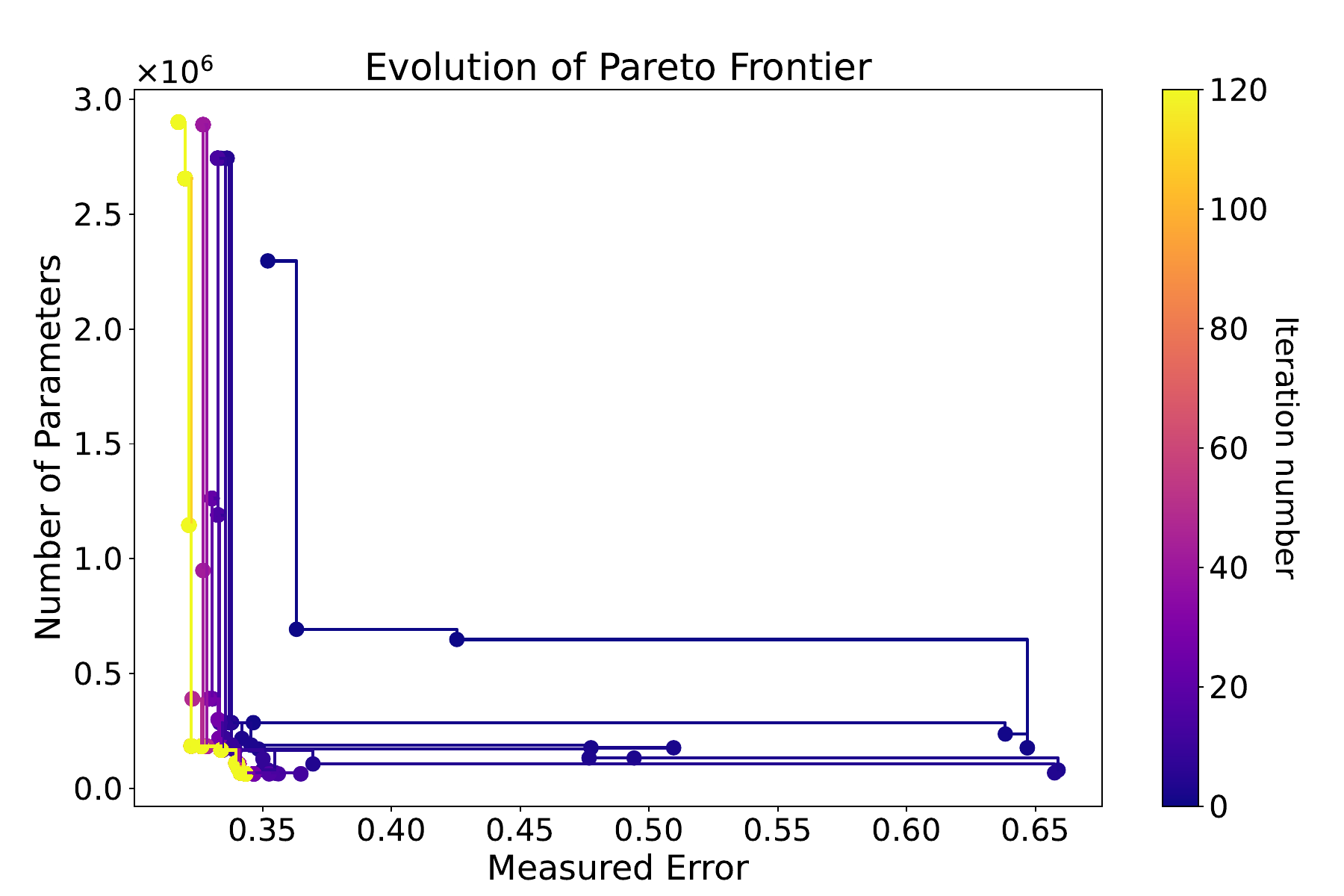}}
    \hfil
  \subfloat[PVPF Task 1: DQH-336]{\includegraphics[width=2.3in]{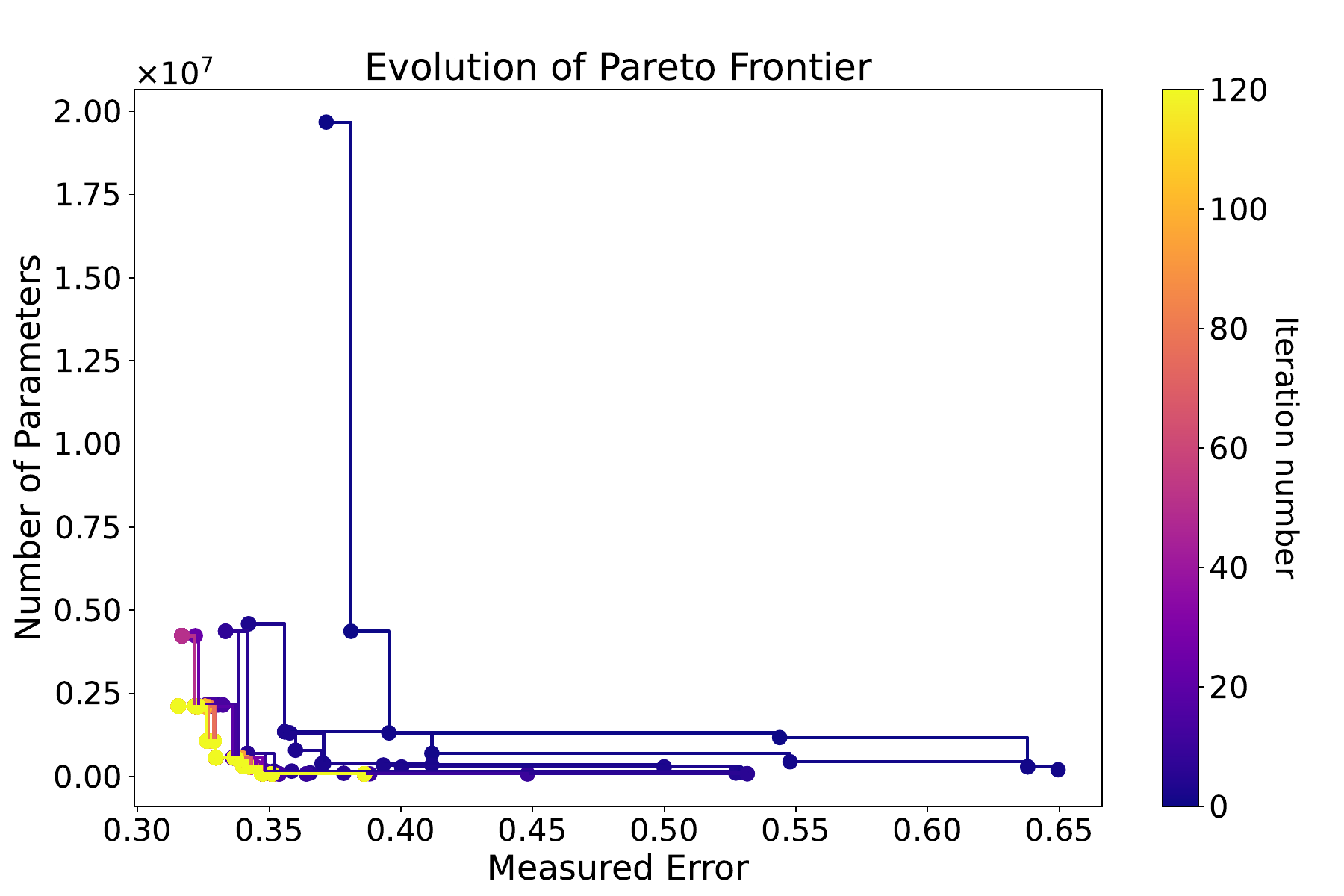}}

  \caption{The optimal architectures found during each iteration for PVPF Task 1. They constitute the Pareto Frontier. The yellow points on the figures indicate the measured error and model scale of architectures in the final Pareto Frontier.}
  \label{pareto1}
\end{figure*}

\begin{figure*}
  \centering
 \subfloat[PVPF Task 2: DQH-12]{\includegraphics[width=2.3in]{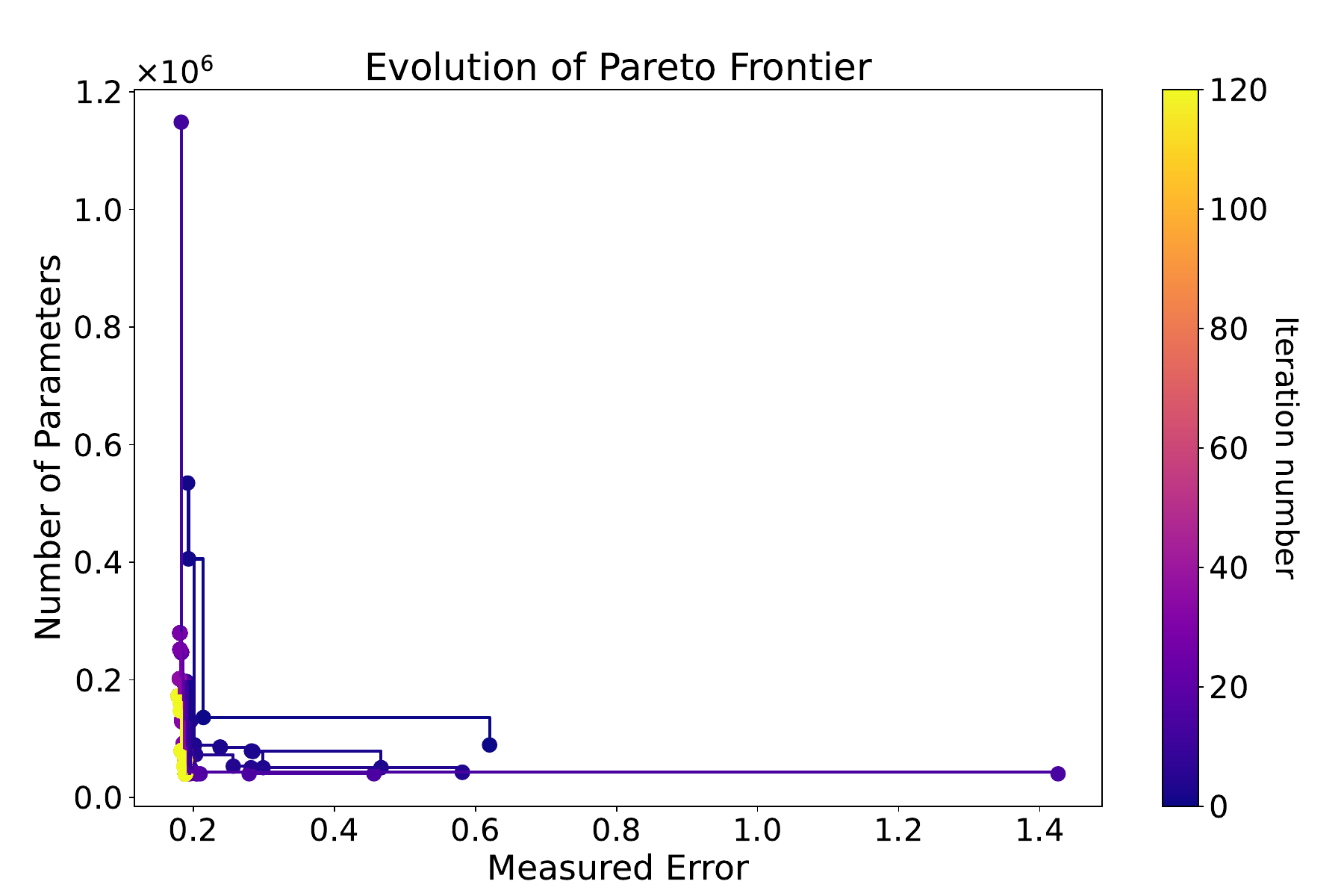}}
  \hfil
  \subfloat[PVPF Task 2: DQH-24]{\includegraphics[width=2.3in]{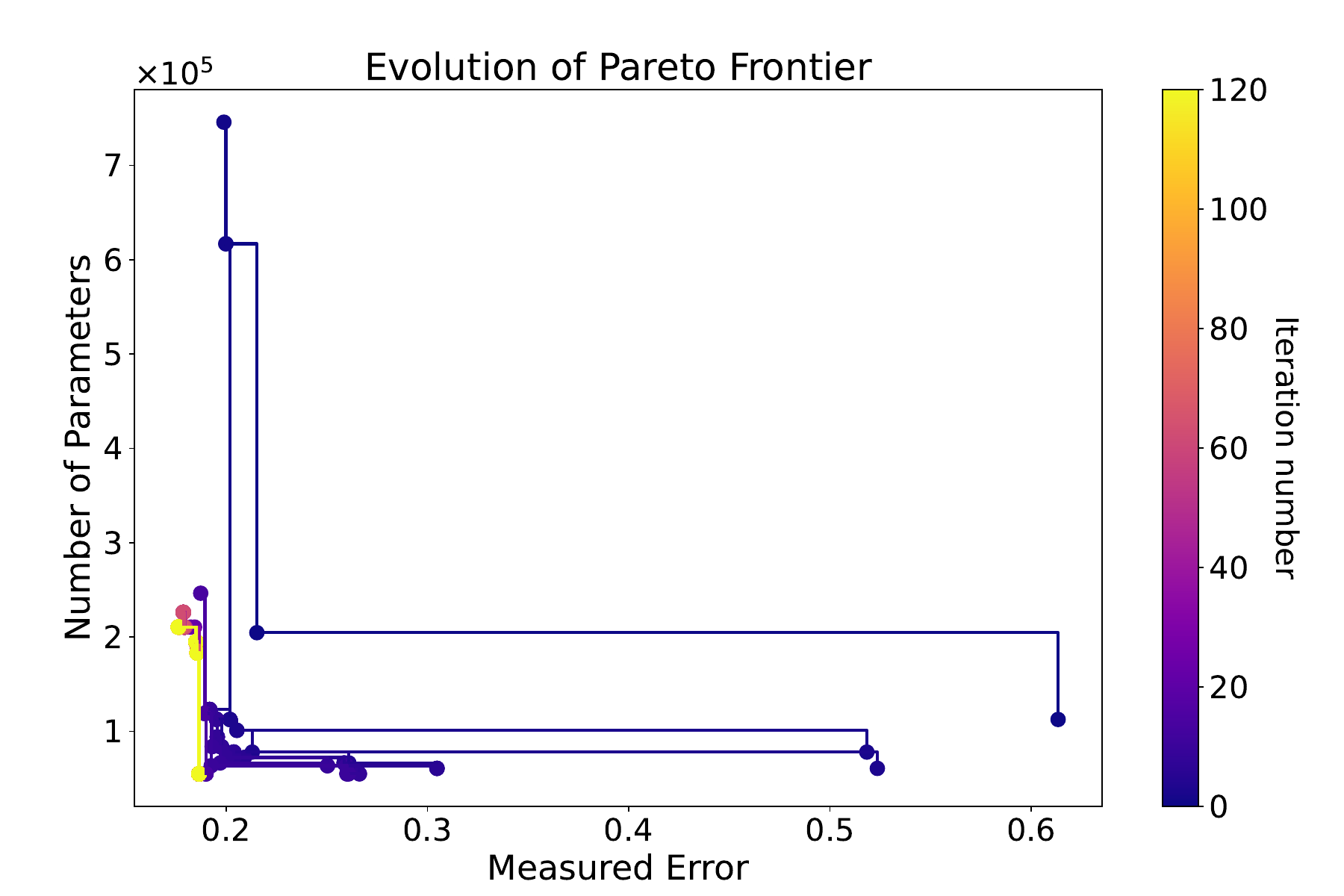}}
    \hfil
  \subfloat[PVPF Task 2: DQH-48]{\includegraphics[width=2.3in]{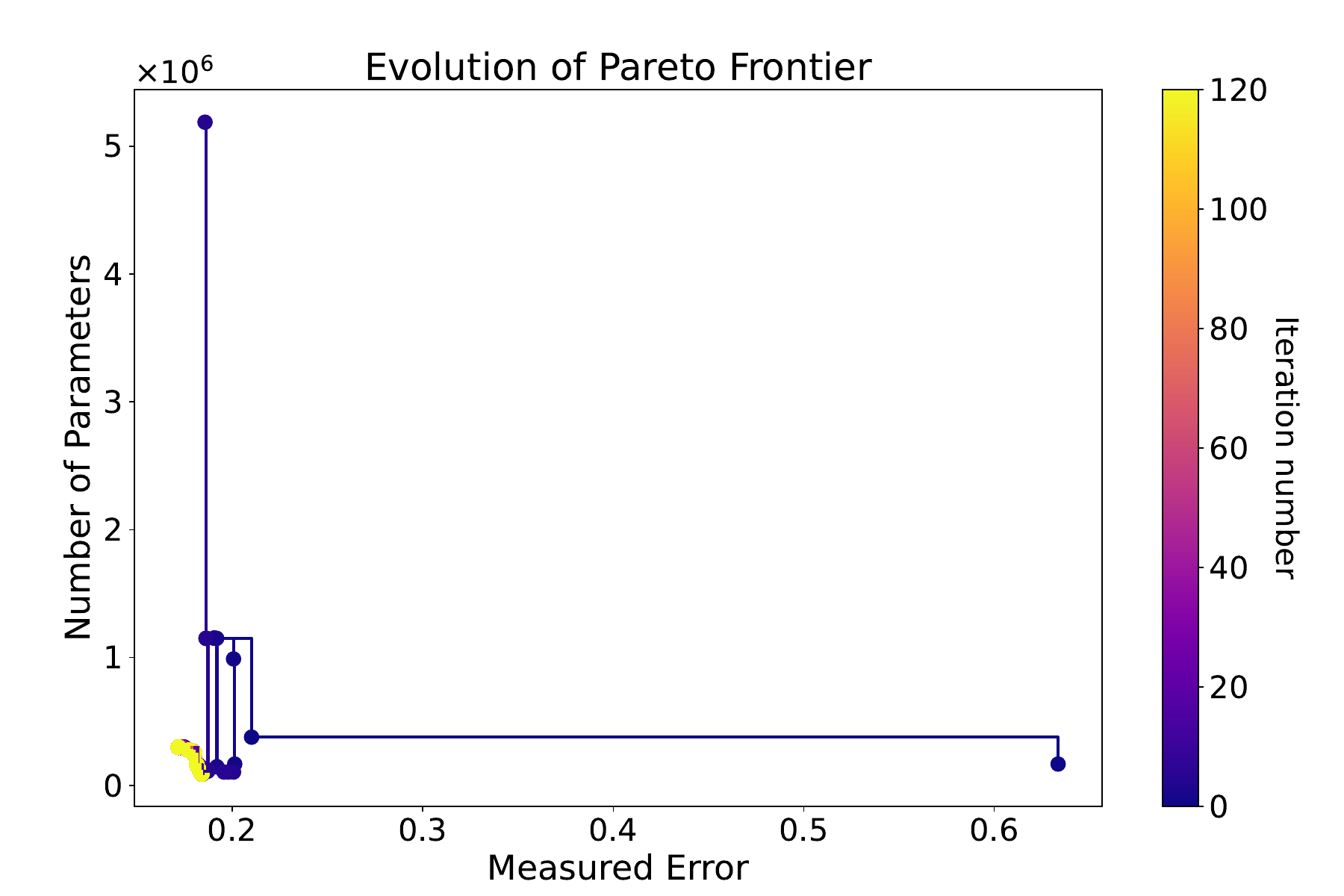}}
    \hfil
  \subfloat[PVPF Task 2: DQH-72]{\includegraphics[width=2.3in]{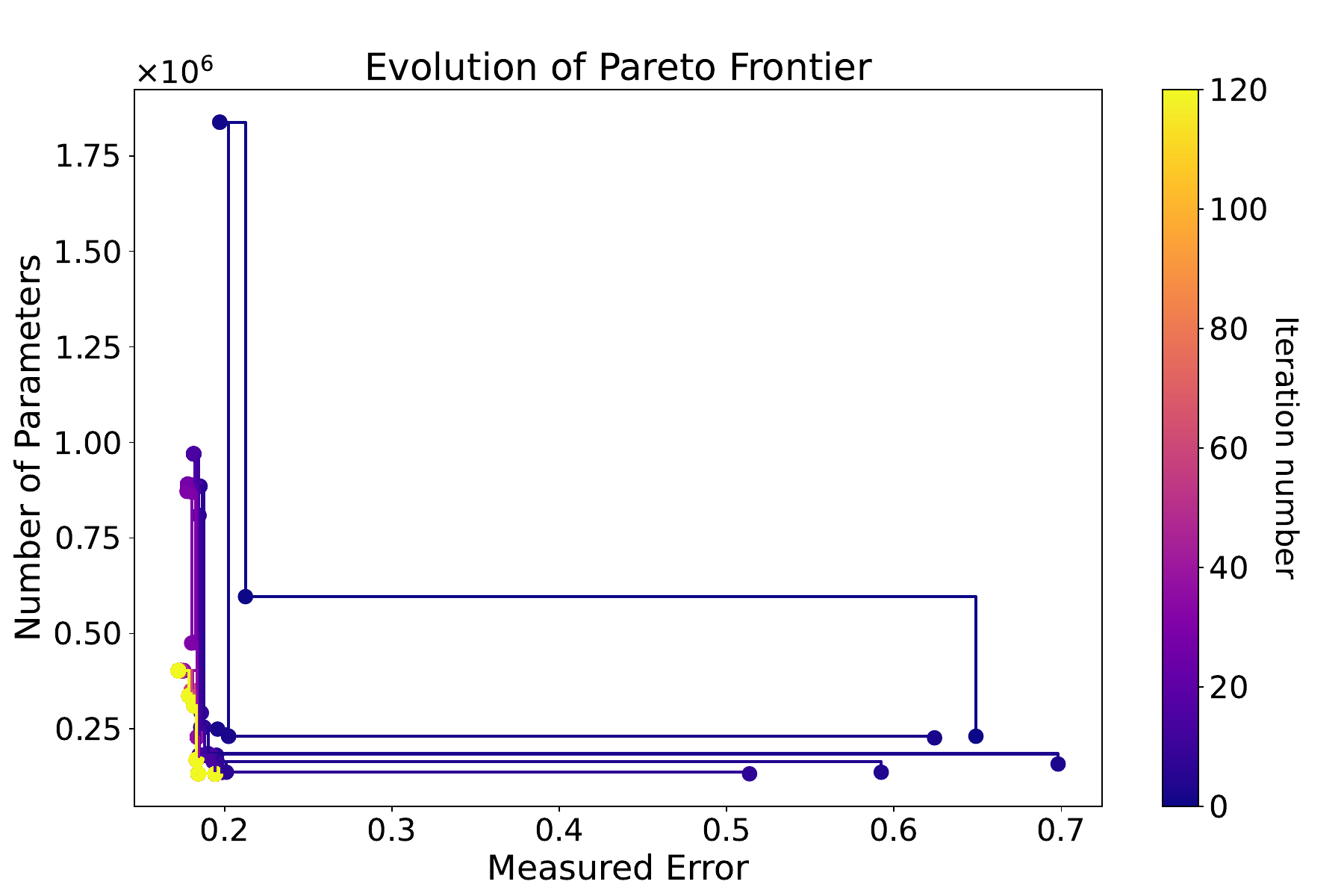}}
    \hfil
  \subfloat[PVPF Task 2: DQH-168]{\includegraphics[width=2.3in]{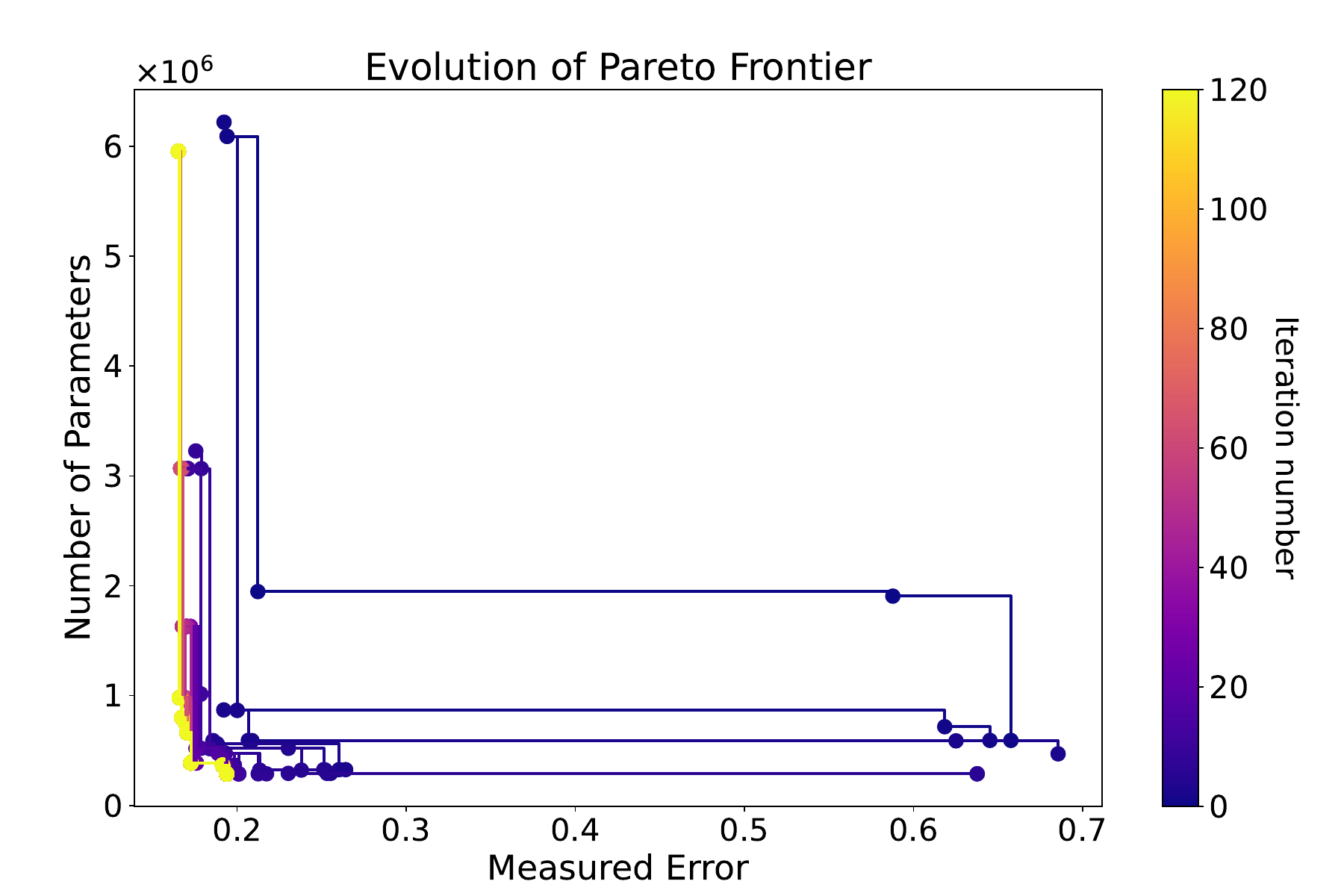}}
    \hfil
  \subfloat[PVPF Task 2: DQH-336]{\includegraphics[width=2.3in]{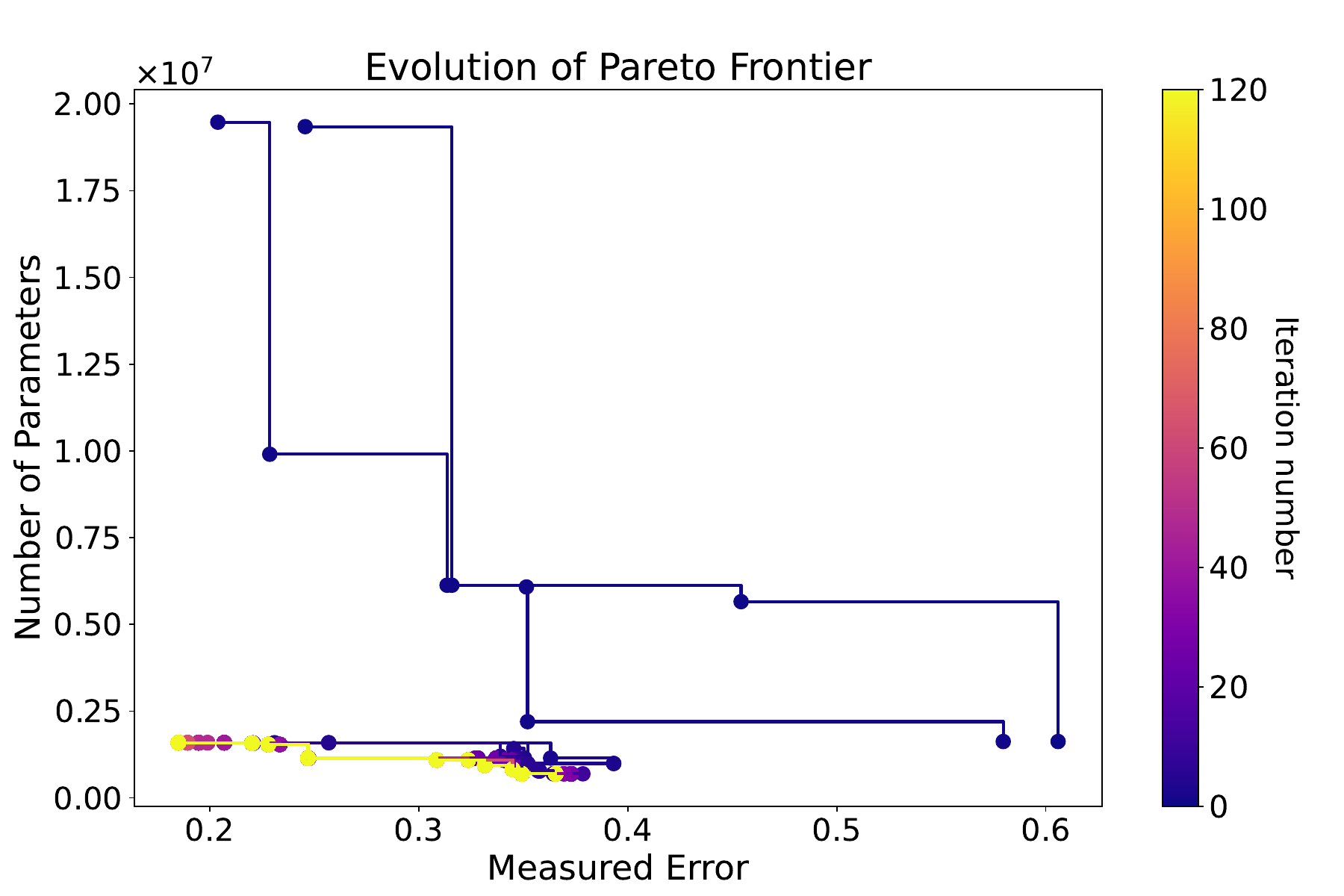}}

  \caption{The optimal architectures found during each iteration for PVPF Task 2. They constitute the Pareto Frontier. The yellow points on the figures indicate the measured error and model scale of architectures in the final Pareto Frontier.}
  \label{pareto2}
\end{figure*}



\begin{figure*}
  \centering
 \subfloat[FSM]{\includegraphics[width=1.6in]{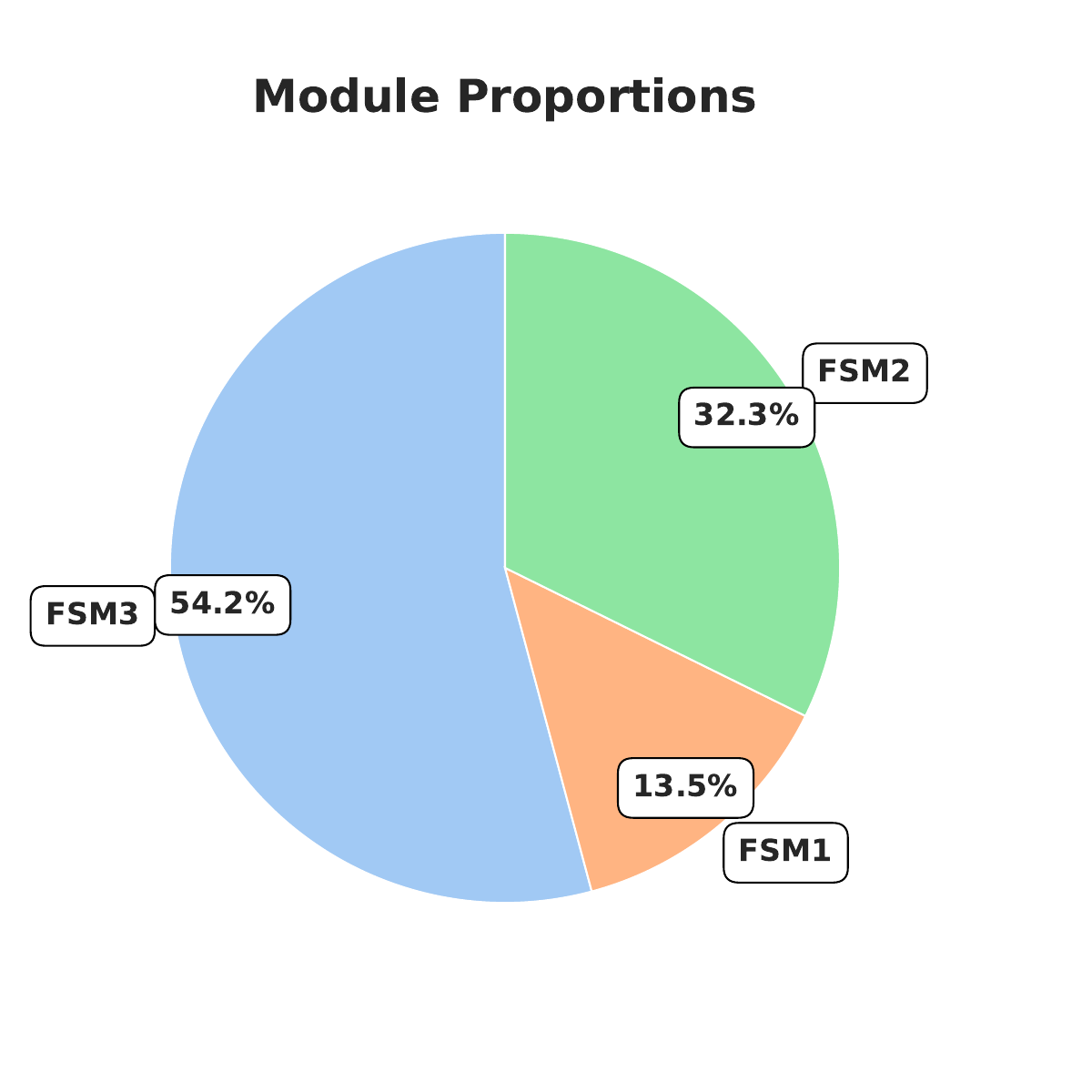}}
  \hfil
 \subfloat[FST]{\includegraphics[width=1.6in]{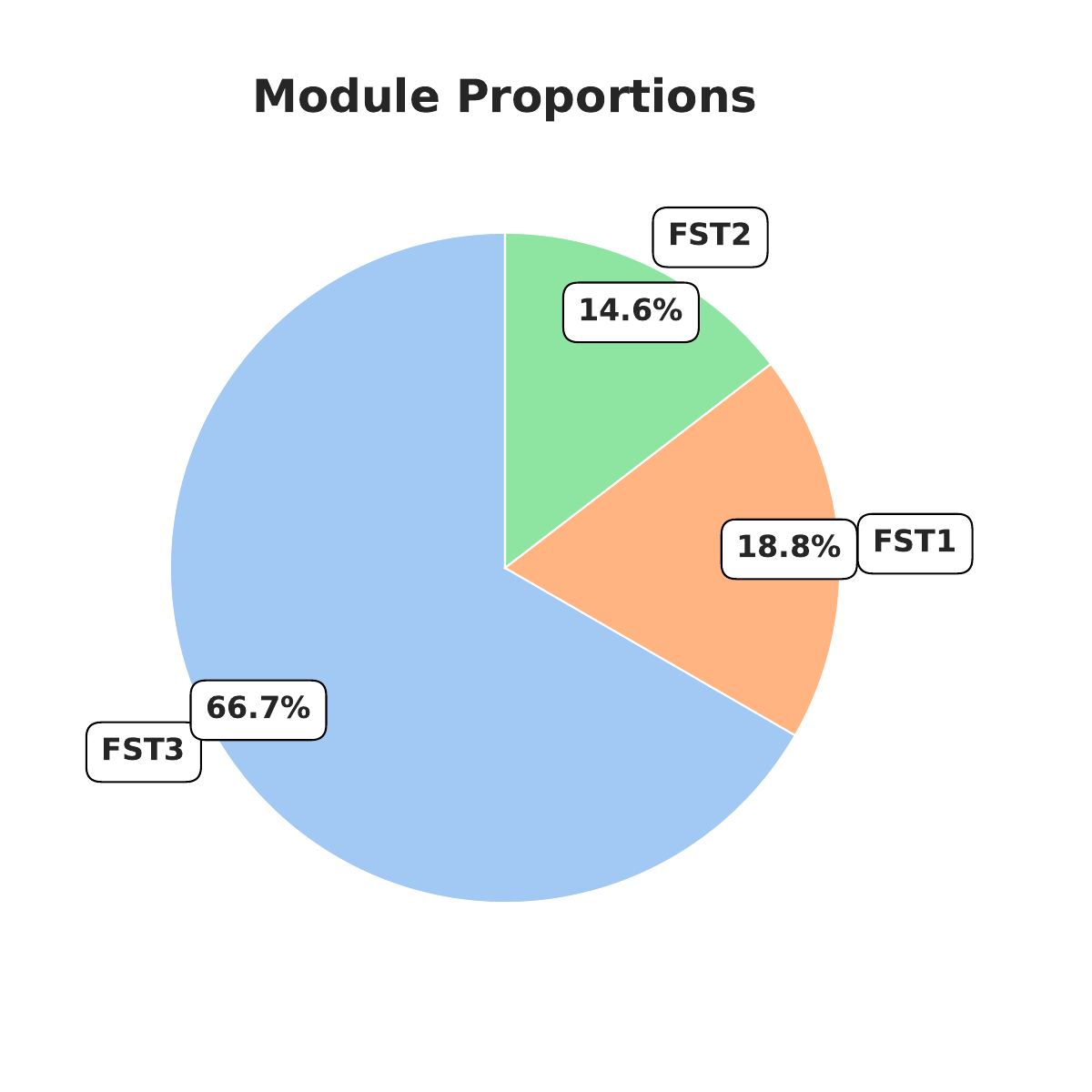}}
  \hfil
  \subfloat[DGM]{\includegraphics[width=1.6in]{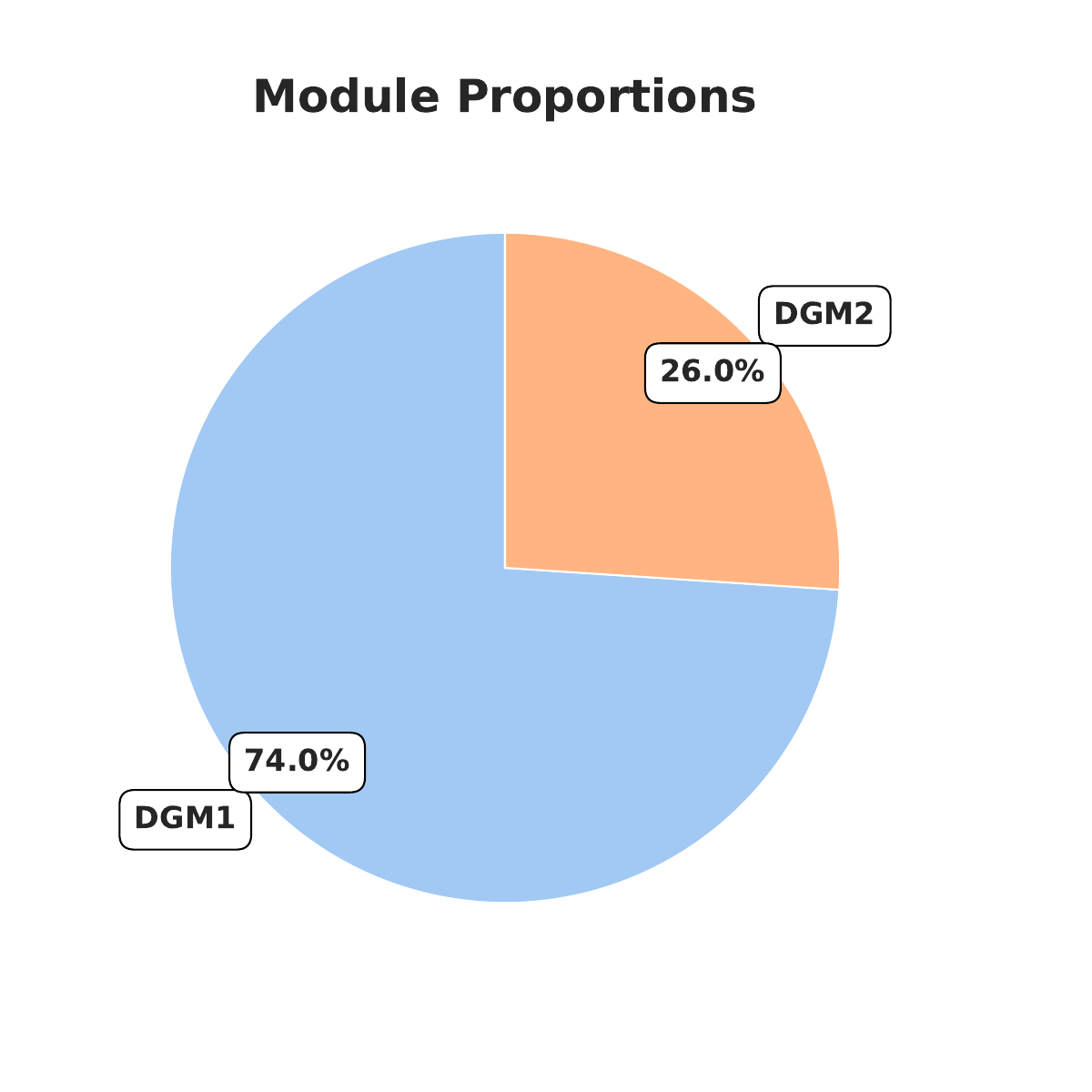}}
  \hfil
  \subfloat[SM]{\includegraphics[width=1.6in]{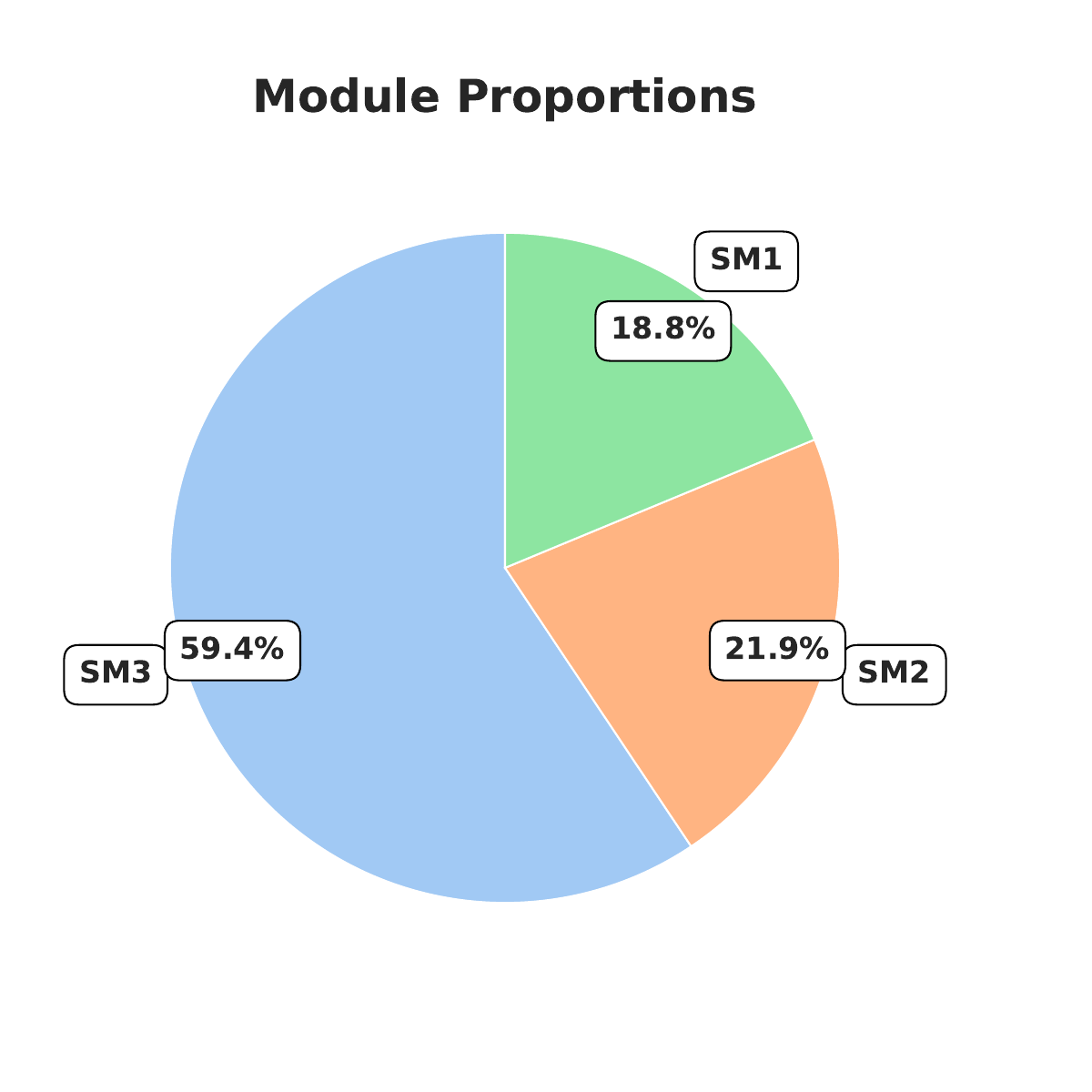}}
  \hfil
  \subfloat[FAM]{\includegraphics[width=1.6in]{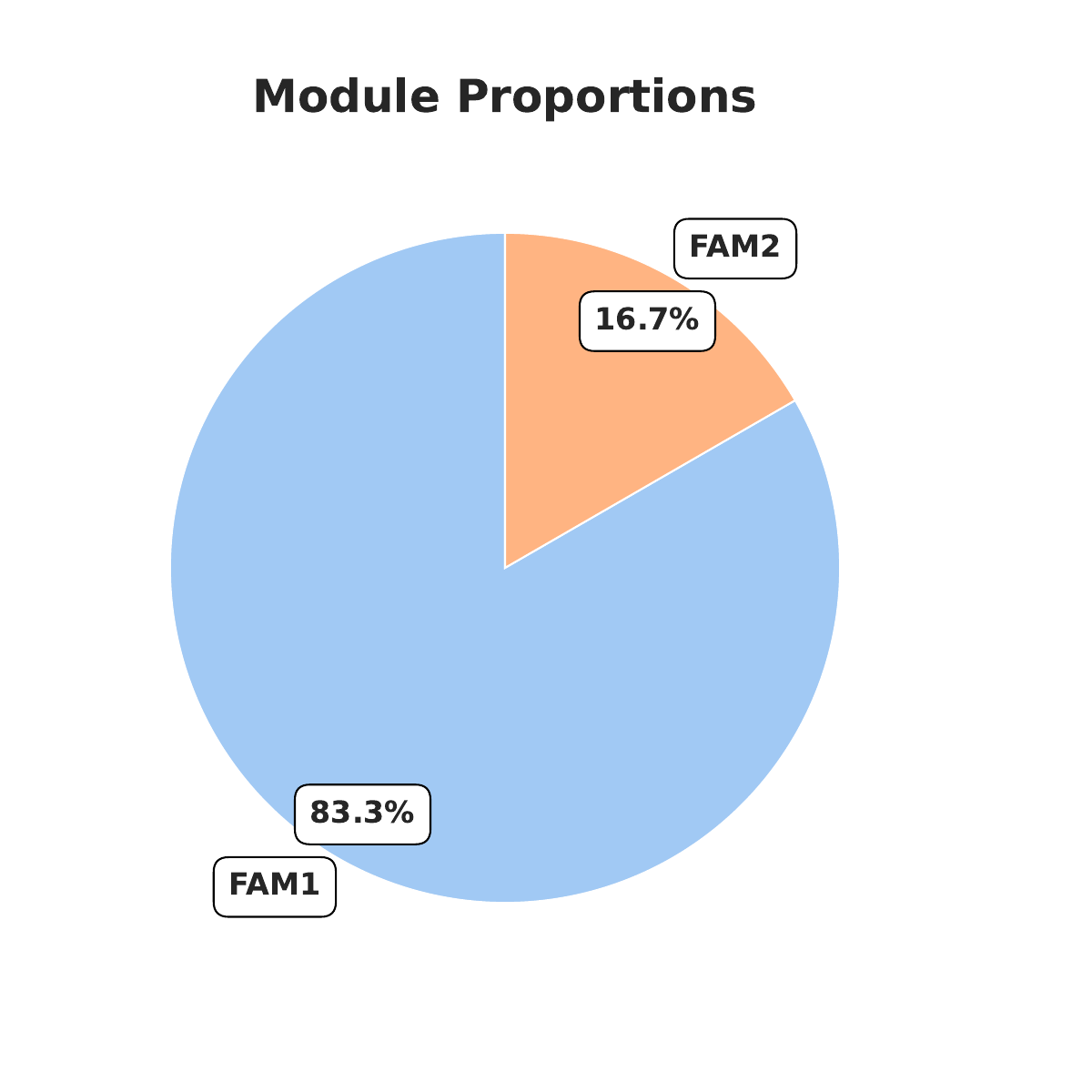}}
  \hfil
  \subfloat[FEM]{\includegraphics[width=1.6in]{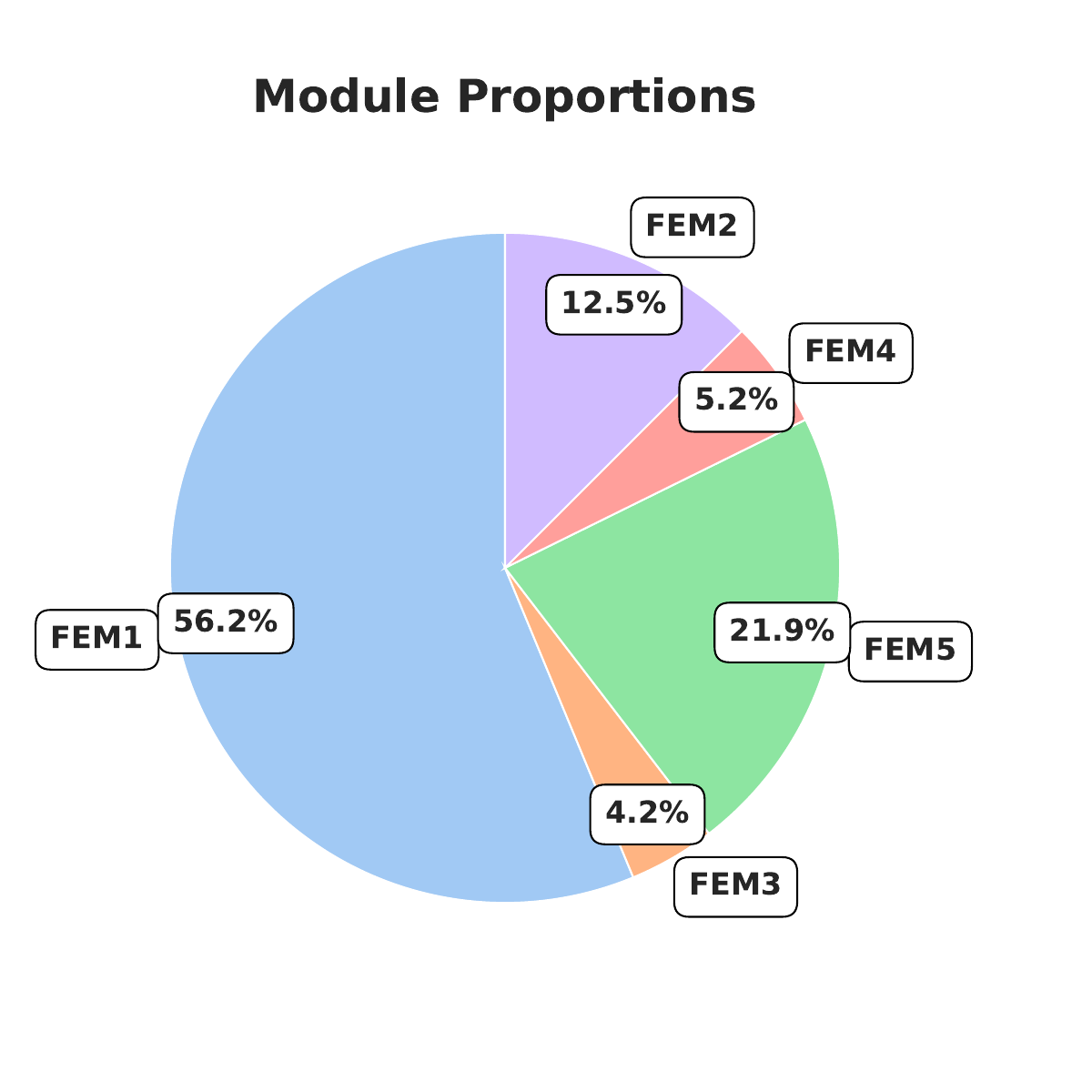}}
  \hfil
  \subfloat[CPS]{\includegraphics[width=1.6in]{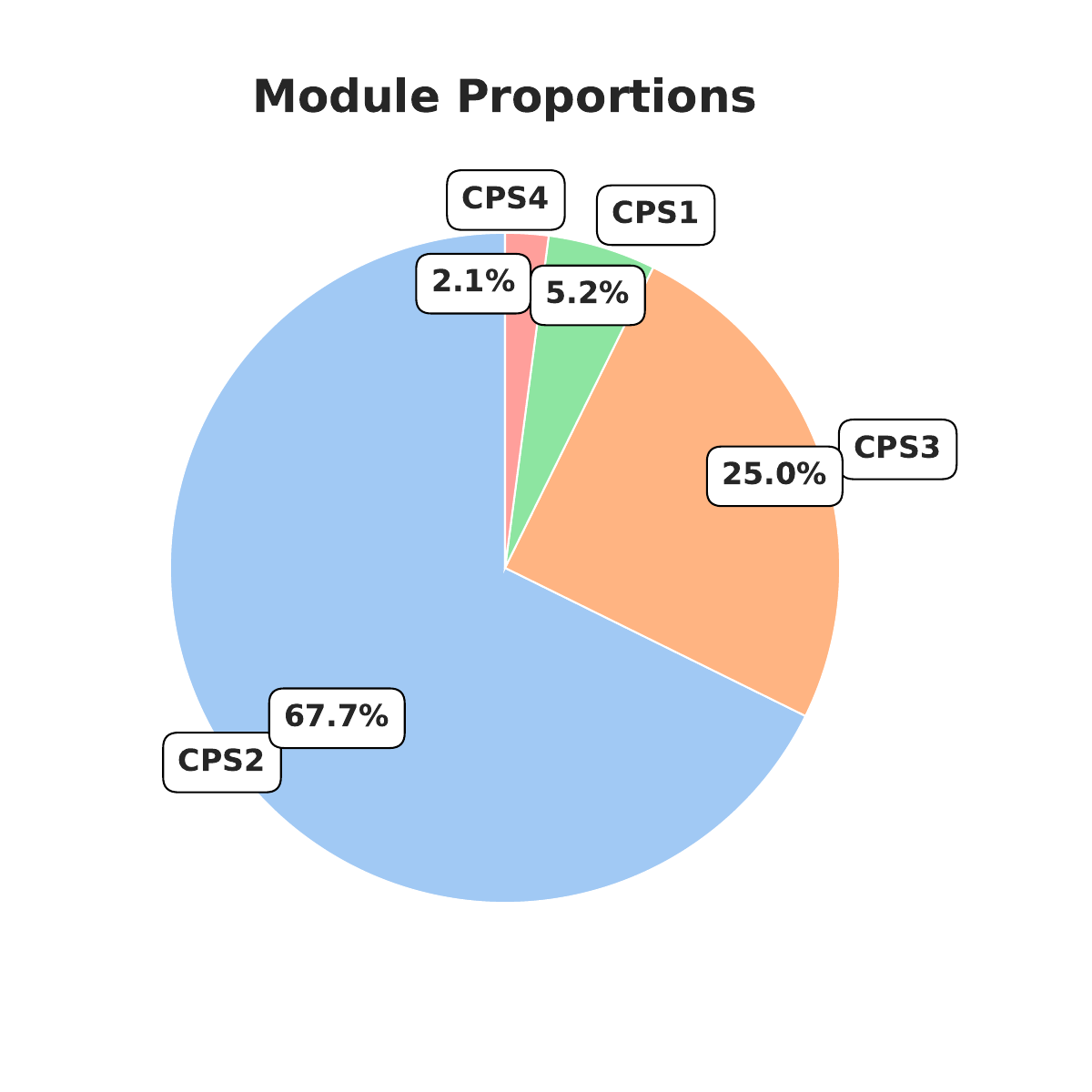}}
  \hfil
  \subfloat[LN]{\includegraphics[width=1.6in]{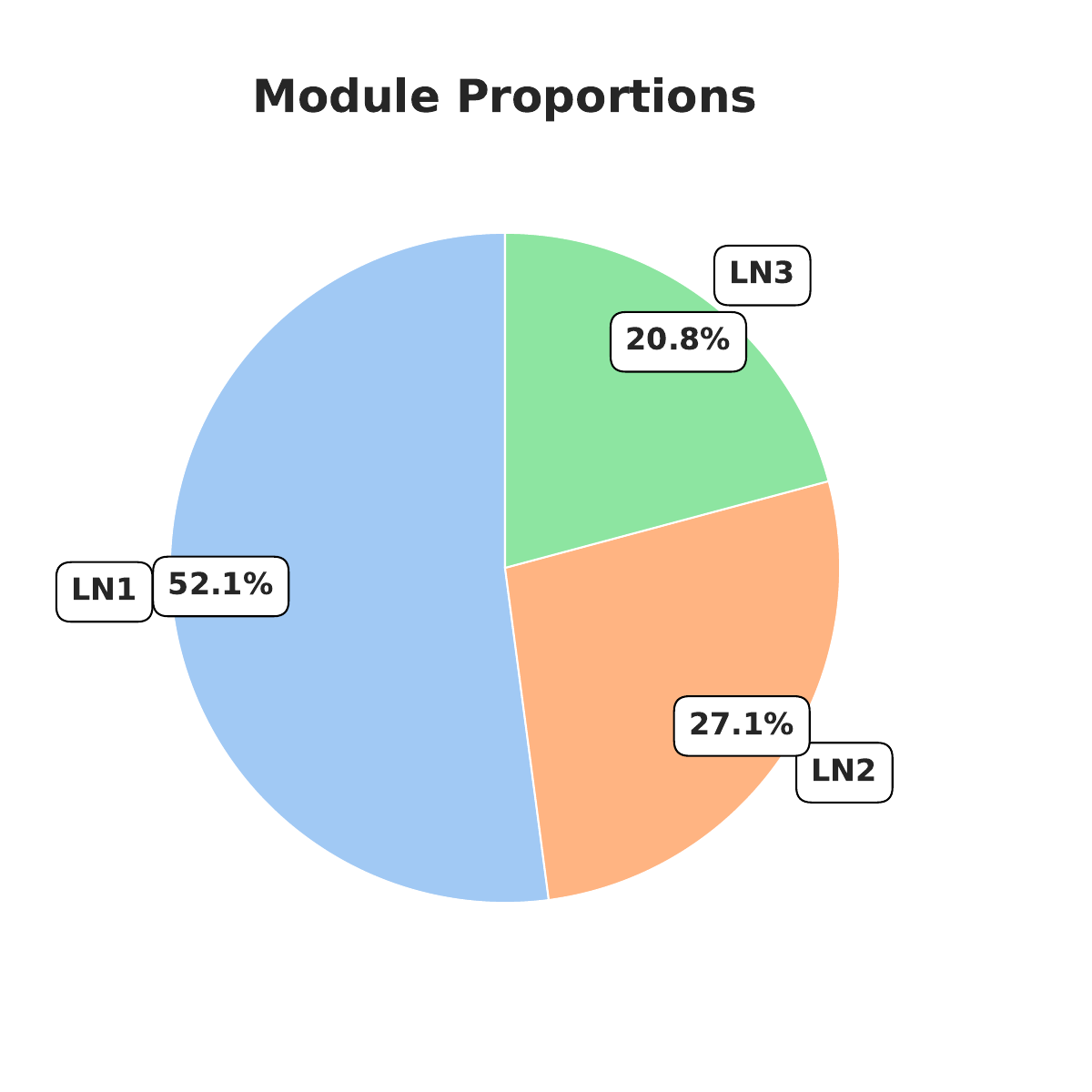}}
  \hfil
  \subfloat[HS]{\includegraphics[width=1.6in]{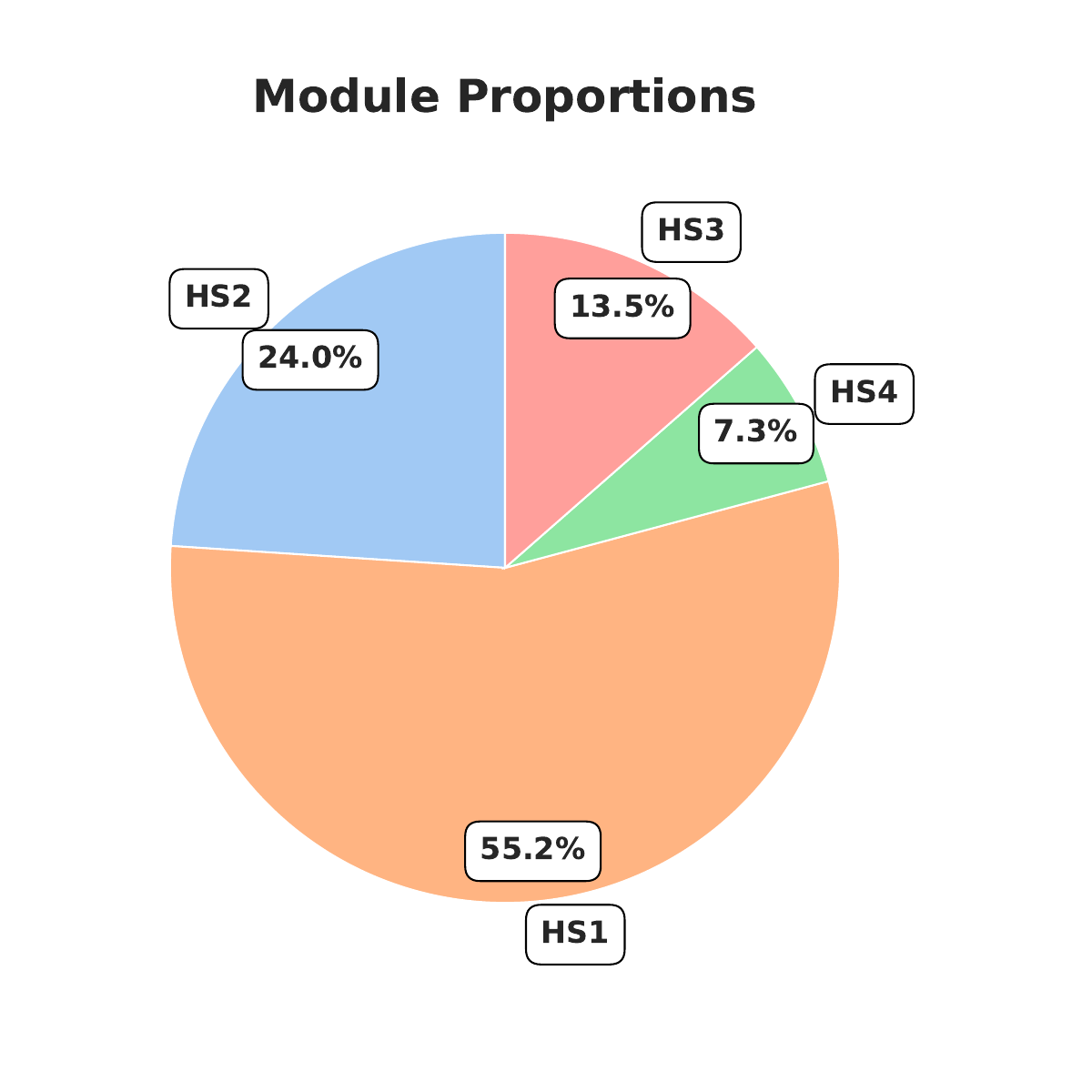}}
  \hfil
  \subfloat[LR]{\includegraphics[width=1.6in]{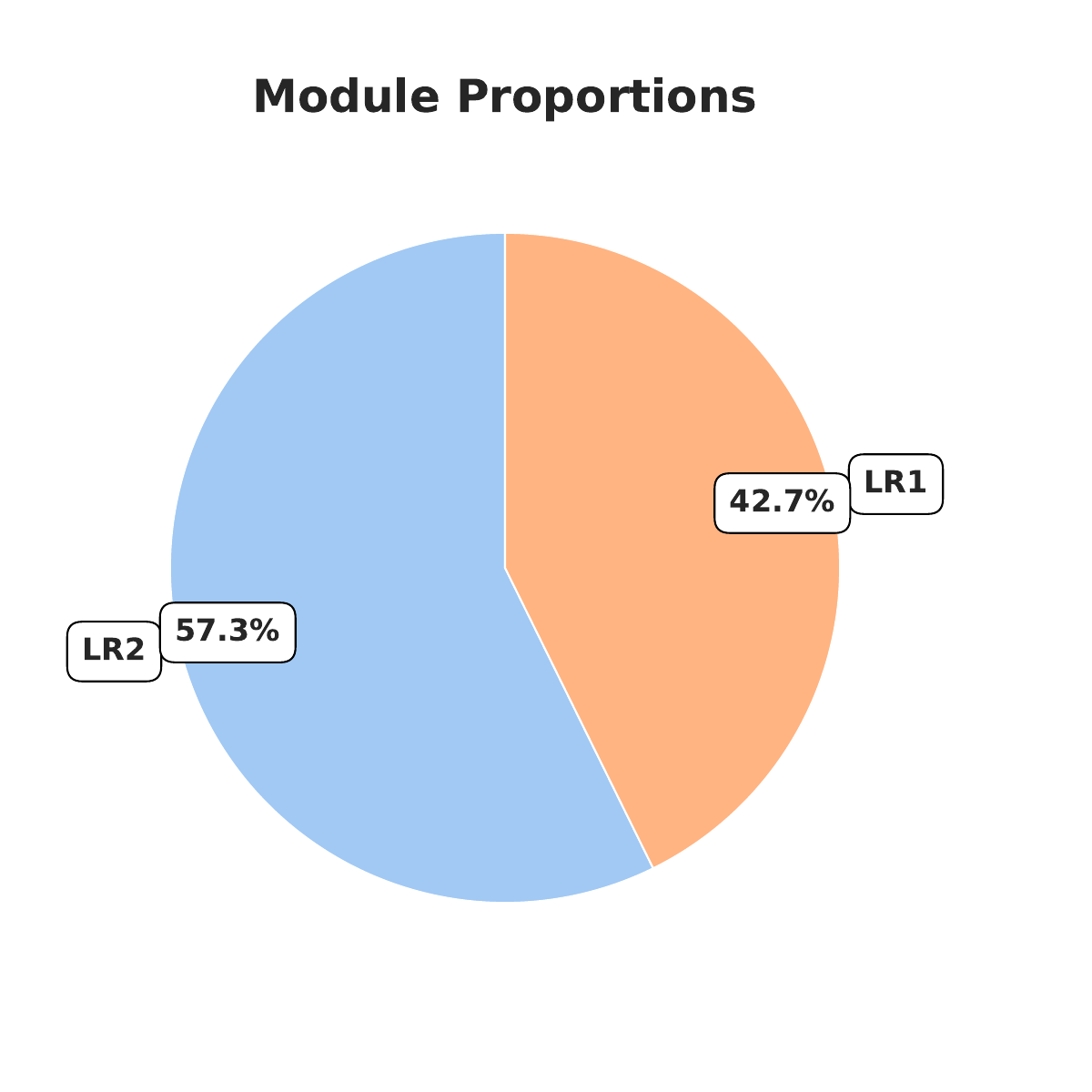}}
  \hfil
  \subfloat[OF]{\includegraphics[width=1.6in]{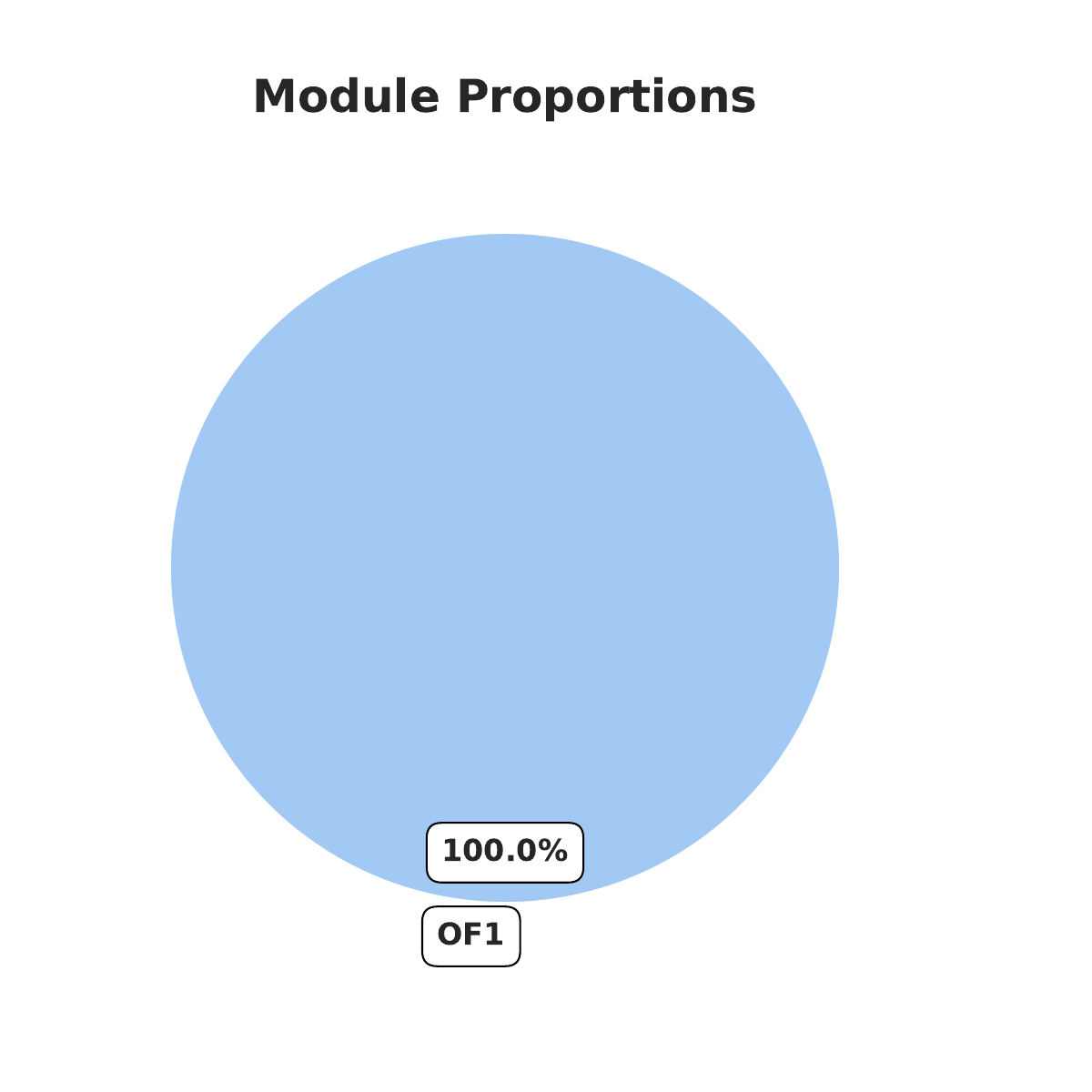}}
  \hfil
  \subfloat[BS]{\includegraphics[width=1.6in]{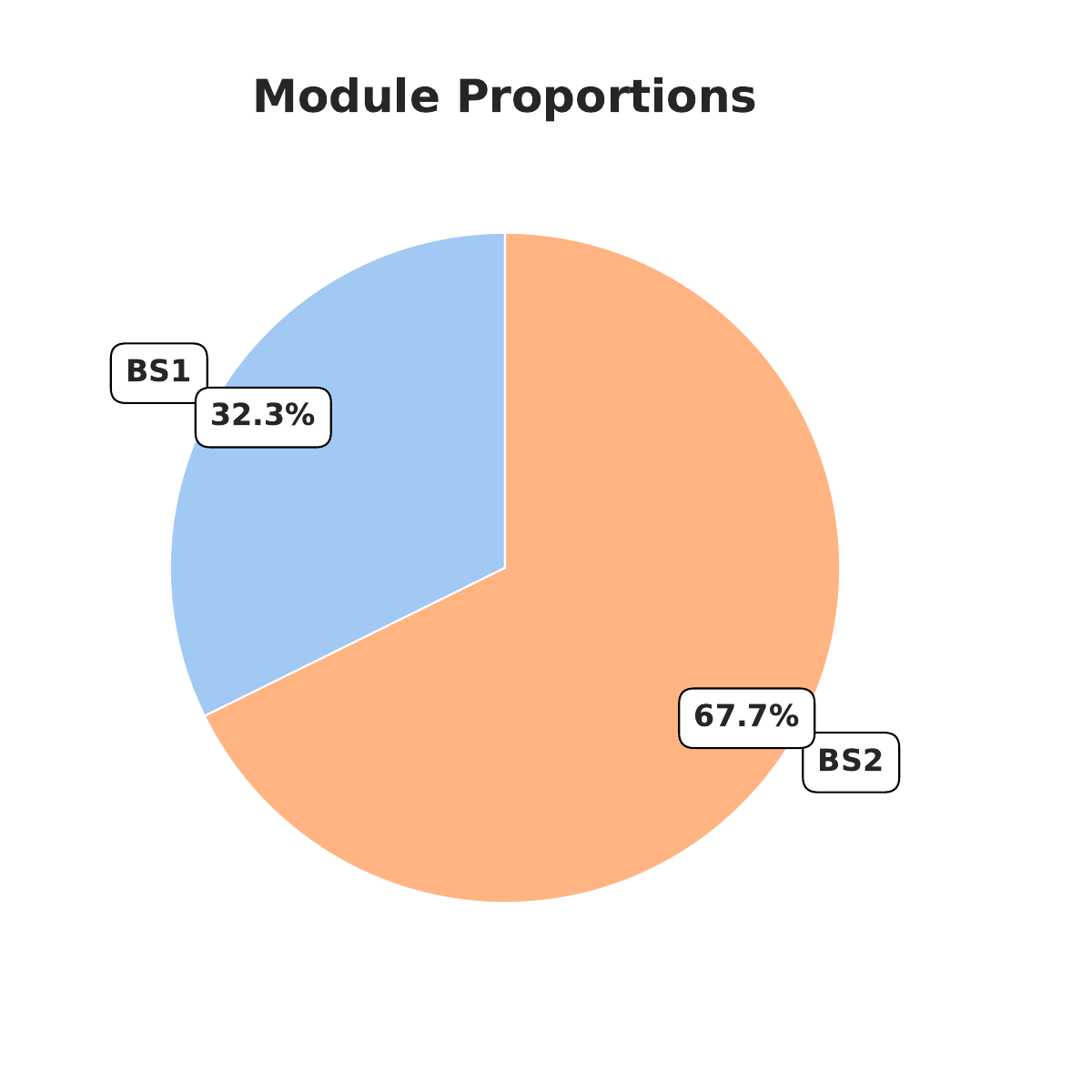}}
  \caption{Proportions of the 12 parameters in the AutoPV search space. The statistics encompass the final optimal set of the searched neural architectures across all 12 subtasks.}
  \label{modules}
\end{figure*}

